\documentclass[runningheads]{llncs}

\usepackage{eccv}

\usepackage{eccvabbrv}

\usepackage{graphicx}
\usepackage{booktabs}
\usepackage{multirow}
\usepackage{makecell}
\usepackage{xcolor}

\usepackage[pagebackref,breaklinks,colorlinks,citecolor=eccvblue]{hyperref}

\renewcommand{\footnotesize}{\scriptsize}

\begin{document}

\pagestyle{headings}
\mainmatter

\title{AVControl: Efficient Framework for Training Audio-Visual Controls}

\author{
Matan~Ben-Yosef$^*$ \and
Tavi~Halperin$^*$ \and
Naomi~Ken~Korem \and
Mohammad~Salama \and
Harel~Cain \and
Asaf~Joseph \and
Anthony~Chen \and
Urska~Jelercic \and
Ofir~Bibi
}
\authorrunning{M.~Ben-Yosef, T.~Halperin \textit{et al.}}
\institute{Lightricks}

\maketitle

\let\thefootnote\relax\footnotetext{
$^*$Equal contribution.\\[2pt]
Project page: \url{https://matanby.github.io/AVControl/}\\
Code: \url{https://github.com/Lightricks/LTX-2/tree/main/packages/ltx-trainer}\\
Video: \url{https://youtu.be/f8D2CwF6BIY}
}

\begin{abstract}
Controlling video and audio generation requires diverse modalities, from depth and pose to camera trajectories and audio transformations, yet existing approaches either train a single monolithic model for a fixed set of controls or introduce costly architectural changes for each new modality.
We introduce AVControl, a lightweight, extendable framework built on LTX-2~\cite{hacohen2026ltx2}, a joint audio-visual foundation model, where each control modality is trained as a separate LoRA on a \emph{parallel canvas} that provides the reference signal as additional tokens in the attention layers, requiring no architectural changes beyond the LoRA adapters themselves.
We show that simply extending image-based in-context methods to video fails for structural control, and that our parallel canvas approach resolves this.
On the VACE Benchmark~\cite{jiang2025vace}, we outperform all evaluated baselines on depth- and pose-guided generation, inpainting, and outpainting, and show competitive results on camera control and audio-visual benchmarks.
Our framework supports a diverse set of independently trained modalities: spatially-aligned controls such as depth, pose, and edges, camera trajectory with intrinsics, sparse motion control, video editing, and, to our knowledge, the first modular audio-visual controls for a joint generation model.
Our method is both compute- and data-efficient: each modality requires only a small dataset and converges within a few hundred to a few thousand training steps, a fraction of the budget of monolithic alternatives.
We publicly release our code and trained LoRA checkpoints.

\end{abstract}

\section{Introduction}
\begin{figure*}[t]
\centering
\newcommand{\timg}[1]{\includegraphics[width=\tcw]{figures/images/teaser_columns/#1.jpg}}%
\newcommand{\tcol}[2]{\begin{minipage}[t]{\tcw}\centering\timg{#1}\\[-5pt]{\tiny #2}\end{minipage}}%
\newlength{\tcw}\setlength{\tcw}{0.175\textwidth}%
%
\begin{minipage}[t]{0.555\textwidth}
\centering{\scriptsize\textbf{Spatial Controls}}\\[-8pt]
\rule{\linewidth}{0.3pt}\\[1pt]
\tcol{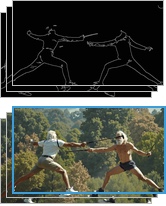}{Canny}\hfill\tcol{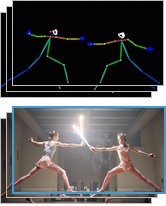}{Pose}\hfill\tcol{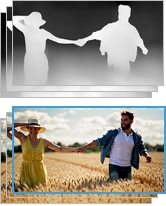}{Depth}
\end{minipage}%
\hspace{0.02\textwidth}%
\begin{minipage}[t]{0.36\textwidth}
\centering{\scriptsize\textbf{Camera Control}}\\[-8pt]
\rule{\linewidth}{0.3pt}\\[1pt]
\tcol{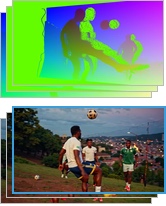}{From Image}\hfill\tcol{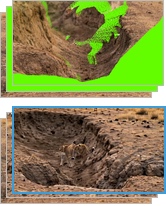}{From Video}
\end{minipage}%
\vspace{8pt}
%
\begin{minipage}[t]{\tcw}
\centering{\scriptsize\textbf{Motion}}\\[-8pt]
\rule{\linewidth}{0.3pt}\\[1pt]
\tcol{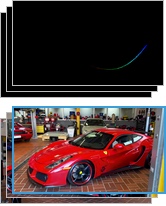}{Sparse Tracking}
\end{minipage}%
\hspace{0.02\textwidth}%
\begin{minipage}[t]{0.36\textwidth}
\centering{\scriptsize\textbf{Editing}}\\[-8pt]
\rule{\linewidth}{0.3pt}\\[1pt]
\tcol{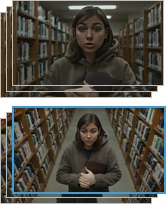}{Cut-on-Action}\hfill\tcol{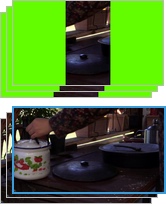}{Inpainting}
\end{minipage}%
\hspace{0.02\textwidth}%
\begin{minipage}[t]{\tcw}
\centering{\scriptsize\textbf{Audio-Visual}}\\[-8pt]
\rule{\linewidth}{0.3pt}\\[1pt]
\tcol{who_is_talking}{Who Is Talking}
\end{minipage}%
\hspace{0.02\textwidth}%
\begin{minipage}[t]{\tcw}
\centering{\scriptsize\textbf{Audio Only}}\\[-8pt]
\rule{\linewidth}{0.3pt}\\[1pt]
\tcol{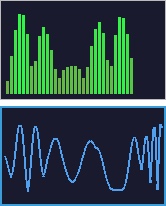}{Intensity$\to$Waveform}
\end{minipage}%
\caption{AVControl trains each control modality as a lightweight LoRA. Each column shows control input (top) and generated output (bottom), covering spatial controls, camera trajectory, motion, editing, and audio-visual generation.}
\label{fig:teaser}
\end{figure*}

Controlling the generation process of video and audio models is essential for practical creative applications.
However, the space of possible controls is vast: different modalities carry different types of information, and the same input (such as a mask) can have entirely different meanings depending on the context.
Rather than attempting to build a single monolithic system that handles all control types, we propose AVControl, a flexible and easily extendable framework that can be rapidly adapted to new modalities, whether standard controls like depth and pose or specialized ones such as rendering Blender previews for real-time game engines (Figure~\ref{fig:teaser}).
We build on LTX-2~\cite{hacohen2026ltx2}, a joint audio-visual DiT that natively generates synchronized video and audio, making it a natural backbone for multimodal control.

The range of controls extends well beyond spatially-aligned ControlNet-style~\cite{zhang2023controlnet} modalities like depth, pose, and canny edges.
We may wish to control camera motion from a single image, or re-render an existing video at a new trajectory while preserving scene dynamics.
When audio is also considered, the space grows further: adapting acoustics to a text-described environment, synchronizing video with a reference audio track, and more.

Our approach draws inspiration from In-Context LoRA (IC-LoRA)~\cite{huang2024iclora}, where a LoRA is trained on an image model to generate composite images, such as paired images side-by-side, with a learned relationship between the panels. At inference time, one half serves as the conditioning input while the other is generated via inpainting.
However, for structural controls such as depth, this approach fails to faithfully follow the conditioning signal (Figure~\ref{fig:canvas_vs_concat}). We hypothesize that the large spatial distance between semantically corresponding positions in the concatenated layout weakens their interaction in the attention layers.
We therefore adopted an approach inspired by Flux Kontext~\cite{batifol2025fluxkontext}, providing the reference on a parallel ``canvas,'' i.e., where additional tokens in the attention layers are processed alongside the generation target.
The challenge with this parallel layout is that the model must distinguish reference tokens from generation tokens. Flux Kontext~\cite{batifol2025fluxkontext} addresses this by introducing a new Rotary Position Embedding (RoPE)~\cite{su2021rope} dimension, which requires learning entirely new positional relationships from extensive compute and large-scale curated paired data. For video, the data cost is even more prohibitive, as temporally aligned multi-view video pairs are substantially harder to curate at scale than image pairs.
Our formulation avoids both costs. LTX-2~\cite{hacohen2026ltx2} assigns a unique per-token timestep, so the model inherently distinguishes clean reference tokens from noised generation tokens (see Section~\ref{sec:method}). The only trainable component is a minimal LoRA adapter~\cite{hu2022lora} on the frozen joint audio-visual backbone. Unlike methods that require new architectural components, this minimal formulation enables faithful video structural control where direct extensions of prior methods fail.
Moreover, because reference and target interact through self-attention, the reference influence can be continuously modulated at inference time globally or locally -- a capability unavailable to channel-concatenation methods.

Because each control modality is a lightweight, independently trained LoRA, this design enables easy extension to new controls without retraining existing ones. Unlike monolithic methods such as VACE~\cite{jiang2025vace}, which train all controls jointly, adding a new control, whether a neural renderer for Blender meshes (Section~\ref{sec:modalities}) or a speech-to-ambient audio transformation (Section~\ref{sec:av}), requires only a small dataset and a short training run. The total training budget across all 13 trained modalities is ${\sim}$55K steps, less than one third of VACE's 200K-step training run.

To accelerate inference, we further propose a small-to-large control grid that reduces the reference canvas resolution for sparse controls such as camera parameters.

\vspace{0.5em}
\noindent To summarize, our contributions are:
\begin{itemize}
    \item A compute- and data-efficient framework for training per-modality control LoRAs on a parallel canvas, enabling faithful video structural control and fine-grained inference-time strength modulation.
    \item A diverse set of independently trained control modalities, from spatially-aligned controls and camera trajectory to audio-visual applications, demonstrating the framework's flexibility.
    \item A small-to-large control grid training strategy that reduces the reference canvas resolution for spatially sparse modalities, lowering latency without sacrificing control fidelity.
\end{itemize}
\section{Related Work}
\subsection{Audio-Visual Foundation Models}
Building on latent diffusion models~\cite{rombach2022latentdiffusion}, recent foundation models have expanded from text-to-image to text-to-video~\cite{blattmann2023stablevideodiffusion,hacohen2025ltxv,yang2025cogvideox} and joint audio-visual generation~\cite{polyak2024moviegen}. A unified audio-visual backbone can share high-level semantics while learning cross-modal alignment, enabling cross-modal control: generating video from audio, audio from video, or editing one modality while preserving consistency with the other.

\subsection{LoRA and Reference-Guided Generation}
Low-Rank Adaptation (LoRA)~\cite{hu2022lora} injects trainable low-rank matrices into frozen layers, enabling parameter-efficient fine-tuning. Its flexibility has been leveraged for diverse use-cases, including identity preservation~\cite{ruiz2023dreambooth}, style transfer~\cite{sohn2023styledrop}, motion animation~\cite{guo2024animatediff}, and multi-LoRA fusion for joint spatial--temporal video control~\cite{zhang2025lionlora}.

Reference-guided generation introduces spatial inputs such as depth, pose, and masks to constrain generation beyond text. One dominant strategy is \emph{channel concatenation}~\cite{brooks2023instructpix2pix}, where the conditioning signal is concatenated along the channel dimension of the noisy latents. An alternative family~\cite{batifol2025fluxkontext,zhang2026omnitransfer} provides the reference as additional attention tokens, enabling richer interaction at the cost of a larger token budget.

\textbf{Controllable video generation.}
Early methods adapt image control to video via ControlNet extensions~\cite{zhao2023controlvideo,zhang2023controlvideo,chen2023controlavideo} or efficient transfer~\cite{cho2025ctrladapter,wang2024easycontrol,peng2024controlnext}. More recent work addresses motion editing~\cite{burgert2025motionv2v}, in-context LoRA for pose~\cite{he2025posegen}, text-driven editing~\cite{liu2023videop2p,liew2023magicedit}, camera and object motion control~\cite{wang2023motionctrl}, and sparse trajectory control~\cite{wang2025ati}.

\textbf{Unified frameworks.}
UNIC~\cite{ye2025unic} represents multimodal conditions as a single token sequence with task-aware RoPE.
Phantom~\cite{liu2025phantom} and OminiControl2~\cite{tan2025ominicontrol2} address subject-consistent and efficient multi-conditional generation, respectively.
OmniTransfer~\cite{zhang2026omnitransfer} unifies spatio-temporal video transfer via task-aware RoPE biases and reference-decoupled causal learning.
VACE~\cite{jiang2025vace} unifies diverse video tasks into a single model with shared condition units but remains limited to its training-time control set.

\textbf{Camera trajectory control.}
ReCamMaster~\cite{bai2025recammaster} re-renders videos at new trajectories via frame-dimension concatenation, controlling only camera extrinsics. BulletTime~\cite{wang2025bullettime} decouples time from camera pose via 4D-RoPE, requiring 40K iterations at batch size 64. VerseCrafter~\cite{zheng2026versecrafter} uses 4D geometric control via a GeoAdapter trained for 380 GPU hours. All introduce new architectural components; our camera LoRAs require only 3{,}000--10{,}000 steps and no backbone modifications.

\textbf{Audio-visual control.}
AV-Link~\cite{hajiali2025avlink} links frozen diffusion models for cross-modal generation but lacks structural controls. EchoMotion~\cite{yang2025echomotion} jointly models video and human motion. Audio ControlNet~\cite{zhu2026audiocontrolnet} provides fine-grained audio control without video generation. Seedance 1.5 Pro~\cite{chen2025seedance} is a joint audio-visual model with lip-sync but no modular control framework.
For video-to-audio intensity control, ReWaS~\cite{jeong2024rewas} and CAFA~\cite{benita2025cafa} train dedicated adapters on unimodal backbones using ${\sim}$160--200K samples; our framework trains a single LoRA on the joint model with ${\sim}$8K samples.

\textbf{Audio-driven talking video.}
MultiTalk~\cite{kong2025multitalk} generates multi-person conversational video by adding audio cross-attention layers and Label RoPE binding to a DiT backbone. Our who-is-talking modality addresses a related problem as a single LoRA on the unmodified joint audio-visual backbone, using only an abstract bounding-box activity signal.

\textbf{Concurrent work.}
VideoCanvas~\cite{cai2025videocanvas} uses in-context conditioning for unified video completion, including inpainting, extension, and interpolation, via Temporal RoPE Interpolation on a frozen backbone. Their approach handles spatiotemporal completion but does not address structural controls such as depth and pose, camera trajectory, or audio-visual modalities.
LoRA-Edit~\cite{gao2025loraedit} uses mask-aware LoRA fine-tuning for first-frame-guided video inpainting but is limited to editing and does not support structural controls or audio.
CtrlVDiff~\cite{xi2025ctrlvdiff} trains a unified diffusion model with multiple graphics-based modalities including depth, normals, albedo, and segmentation, but uses a fixed set of controls determined at training time and does not extend to camera trajectory or audio-visual modalities.

\textbf{Our approach.}
In contrast to monolithic models such as VACE or unified token approaches like UNIC, we train each control modality as a separate LoRA, with no new layers or input projections. Unlike Flux Kontext~\cite{batifol2025fluxkontext}, which introduces RoPE~\cite{su2021rope} offsets, OmniTransfer~\cite{zhang2026omnitransfer}, which uses task-aware RoPE for video, and VideoCanvas~\cite{cai2025videocanvas}, which uses Temporal RoPE Interpolation, we require no positional encoding changes, relying instead on LTX-2's per-token timestep to distinguish reference from generation tokens.
\section{Method}
\label{sec:method}
\begin{figure}[t]
\centering
\includegraphics[width=\linewidth]{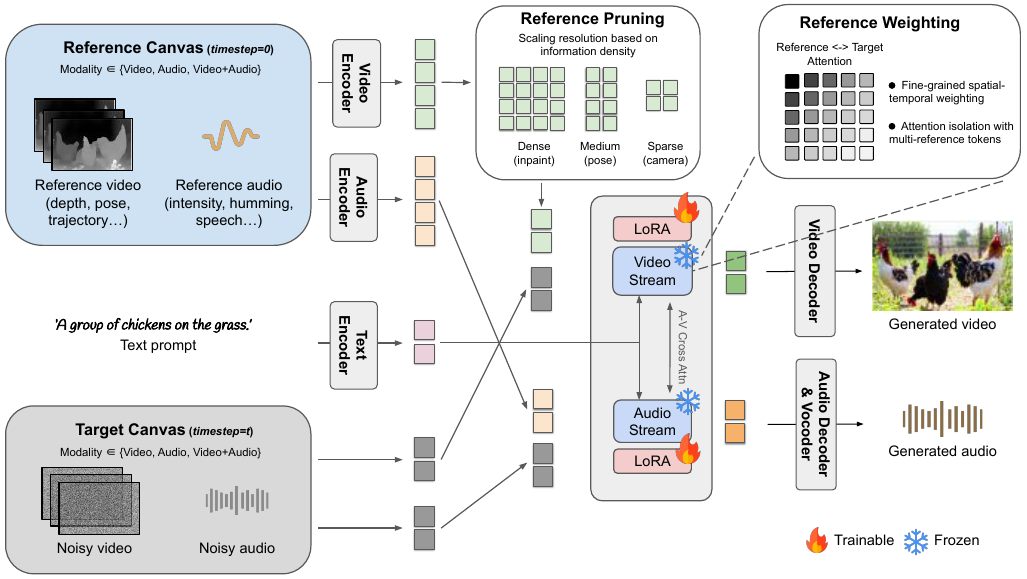}
\caption{Overview of AVControl. The reference signal is placed on a parallel canvas as additional tokens in self-attention. A LoRA adapter is the only trainable component; the backbone remains frozen.}
\label{fig:method_overview}
\end{figure}

An overview of AVControl is shown in Figure~\ref{fig:method_overview}. The reference control signal is placed on a parallel canvas alongside the generation target, and a lightweight LoRA adapter is the only trainable component. We describe each design decision below.

\subsection{Parallel Canvas Conditioning}
A common strategy for incorporating reference signals into diffusion models is \emph{channel concatenation}, where the reference is concatenated along the channel dimension of the noisy latents and fed into the diffusion model. This incurs negligible latency overhead but requires new input-projection weights.

We instead encode the reference signal through the same VAE as the generation target, producing a set of latent patch tokens. These reference tokens are concatenated along the sequence dimension with the noisy target tokens and processed jointly through the transformer's self-attention layers. Reference tokens are assigned a clean timestep ($t{=}0$) while generation tokens carry the current noise level, allowing the model to inherently distinguish the two without positional encoding changes. Training uses the standard diffusion denoising objective, with the loss computed only on the generation tokens; reference tokens serve as clean conditioning context. A lightweight LoRA adapter on the frozen transformer is the only trainable component, applied by default to all attention projection matrices and feed-forward layers, with the exact set of target modules optimized per modality (see supplementary Table~\ref{tab:training_details}). While this approach increases the token count, it provides three important advantages:

\begin{enumerate}
    \item \textbf{Training efficiency.} Some modalities converge in as few as a few hundred steps, with most spatially-aligned controls requiring only a few thousand. This is because we leverage the pre-trained self-attention layers for injecting the control.

    \item \textbf{Fine-grained reference weighting.} Because reference and target interact through self-attention, we can directly scale the attention weights between target queries and reference keys. A global strength parameter uniformly scales all target-to-reference attention, providing a continuous trade-off between structural fidelity and generative freedom. Local modulation varies this scaling per token, enabling spatial or temporal fading of the reference influence (see Fig.~\ref{fig:strength_modulation} in the supplementary). Such modulation is impossible when the reference is fused at the input channel level.

    \item \textbf{Support for misaligned references.} Channel concatenation assumes pixel-level spatial alignment. Our formulation imposes no such constraint: for example, our \emph{cut-on-action} control, which re-renders a scene from a substantially different camera angle, is similar to camera trajectory control but with potentially large viewpoint changes and a different starting frame. The reference and target videos are temporally aligned but spatially different, and the parallel canvas learns this correspondence despite the lack of pixel-level alignment.
\end{enumerate}

\begin{figure}[t]
\centering
\includegraphics[width=0.32\linewidth]{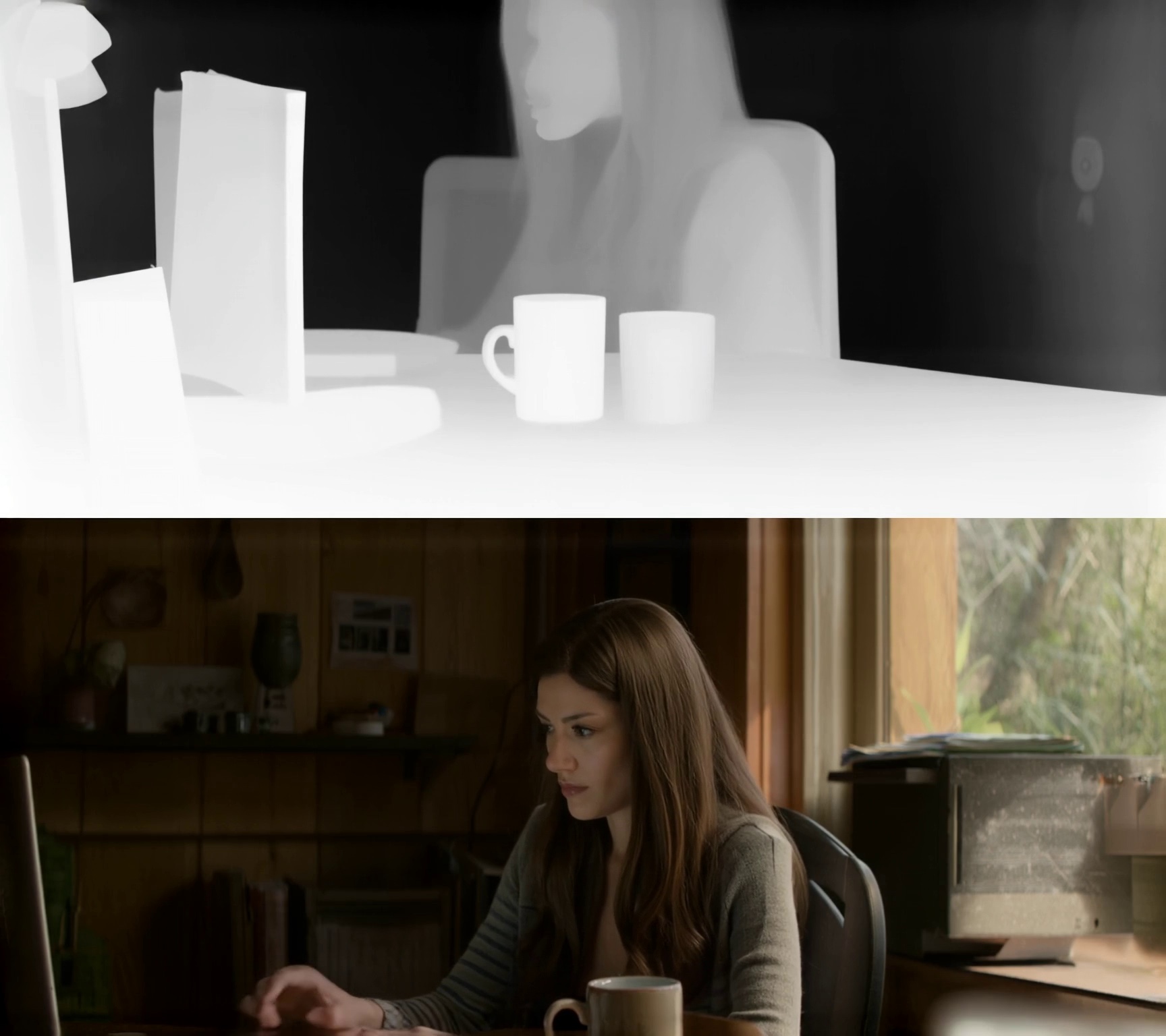}\hfill
\includegraphics[width=0.32\linewidth]{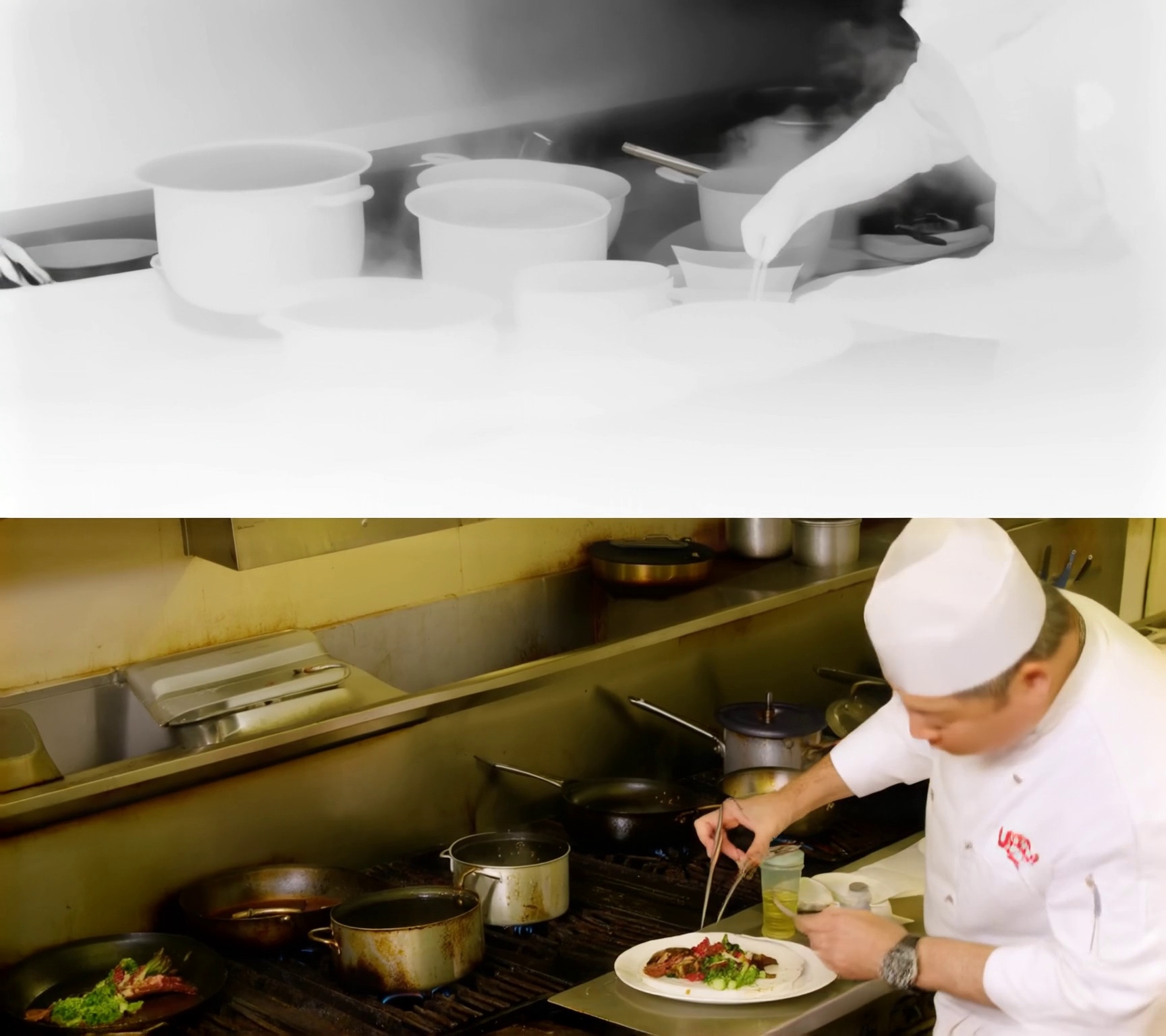}\hfill
\includegraphics[width=0.32\linewidth]{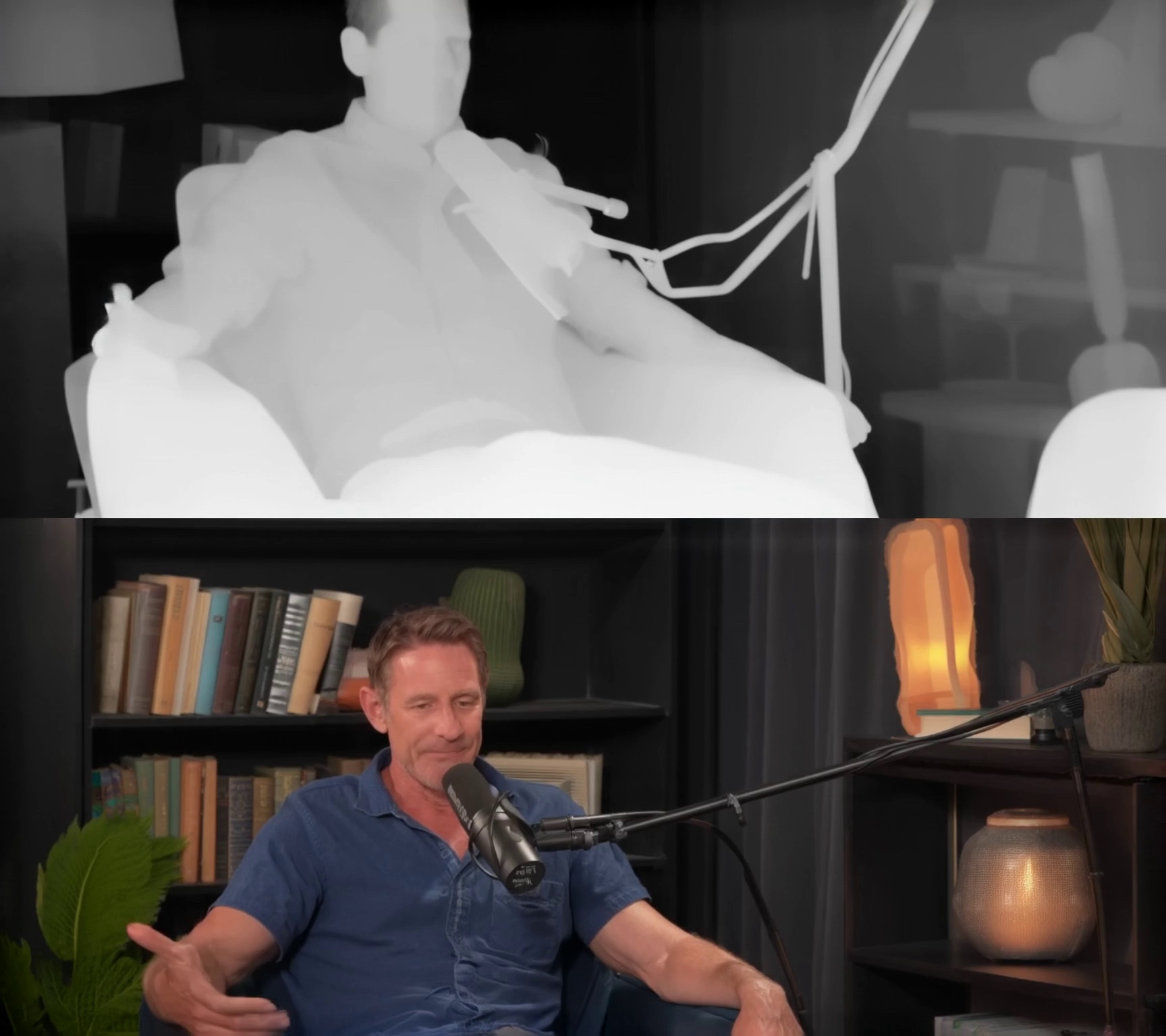}
\caption{Spatial concatenation for depth-guided generation. Each panel shows the input depth map (top) and the output from a concatenation-based LoRA (bottom). The model captures general scene semantics but fails to faithfully follow the spatial structure of the depth signal, motivating our adoption of the parallel canvas approach.}
\label{fig:canvas_vs_concat}
\end{figure}

\noindent We illustrate the failure of spatial concatenation in Figure~\ref{fig:canvas_vs_concat}: a concatenation-based LoRA trained for depth-guided generation captures scene semantics but does not faithfully follow the spatial structure of the depth signal, motivating our use of a parallel canvas approach.

\subsection{Per-Modality LoRA Training}
Our framework is built on top of a joint audio-visual model, yet each LoRA can be trained on a \emph{single} modality (either video or audio) or on joint audio-visual pairs. A video-only LoRA (e.g., depth-to-video) controls the video stream while the base model freely generates synchronized audio. An audio-only LoRA (e.g., speech-to-ambient) controls the audio stream while the base model generates accompanying video. This single-modality training keeps individual runs small and focused while the joint foundation model provides cross-modal generation at no additional training cost. We can apply a video LoRA and an audio LoRA at the same generation.

\subsection{Combining Conditions}
The framework conditions on a single reference signal. To combine multiple control signals, we merge them onto one canvas by compositing; for instance, masked depth overlaid with pose for neural rendering from Blender, keeping geometry aligned while allowing the model freedom on character motion.

\subsection{Small-to-Large Control Grid}
Not all control modalities carry the same amount of information. Dense, spatially-aligned signals such as depth maps require a relatively high-resolution reference canvas, while sparser controls like camera parameters can be expressed with far fewer tokens. We leverage this by scaling the reference canvas resolution according to the information density of each modality. This \emph{small-to-large control grid} reduces the number of additional attention tokens, and consequently the inference latency and memory overhead, for modalities that do not require pixel-level reference detail; see the supplementary for details.

\noindent We next validate these design choices on standard benchmarks and demonstrate the framework across a diverse set of control modalities.

\section{Experiments}
We evaluate AVControl on standard benchmarks, demonstrate its extendability across diverse modalities, and analyze training efficiency.

\subsection{Experimental Setup}
\label{sec:setup}
We train all LoRAs on top of LTX-2~\cite{hacohen2026ltx2}, a frozen joint audio-visual foundation model. Each per-modality LoRA is trained independently on a single H100 GPU; full training details are provided in the supplementary (Table~\ref{tab:training_details}).

We use fixed generation parameters across all evaluations: constant seed~42, CFG~1.0, and LoRA strength~1.0. The low guidance scale reflects LTX-2's distilled inference mode; the control signal provided via the parallel canvas supplies sufficient structural context, making high guidance scales unnecessary.

For quantitative evaluation, we adopt the VACE Benchmark~\cite{jiang2025vace} (20~samples each for depth, pose, inpainting, and outpainting). We use the exact same published input videos and control signals as all baselines, ensuring a fair comparison against reported numbers. We report six VBench~\cite{huang2024vbench} metrics: Aesthetic Quality (AQ), Background Consistency (BC), Dynamic Degree (DD), Imaging Quality (IQ), Motion Smoothness (MS), and Subject Consistency (SC).

\subsection{Quantitative Evaluation}
\label{sec:quant}

\begin{table*}[t]
\centering
\caption{VBench evaluation on the VACE Benchmark. All methods use the same inputs. Baseline numbers from~\protect\cite{jiang2025vace}. \textbf{Bold}: best per metric.}
\label{tab:vbench_comparison}
\setlength{\tabcolsep}{4pt}
\begin{tabular}{llccccccc}
\toprule
Task & Method & AQ & BC & DD & IQ & MS & SC & Avg. \\
\midrule
\multirow{5}{*}{Depth} 
& Control-A-Video~\cite{chen2023controlavideo} & 50.6 & 91.7 & \textbf{70.0} & 67.8 & 97.6 & 88.1 & 77.6 \\
& VideoComposer~\cite{wang2023videocomposer} & 50.0 & 94.2 & \textbf{70.0} & 59.4 & 96.2 & 89.8 & 76.6 \\
& ControlVideo~\cite{zhao2023controlvideo} & \textbf{63.3} & 95.0 & 10.0 & 65.1 & 96.5 & 92.3 & 70.4 \\
& VACE & 56.7 & \textbf{96.1} & 60.0 & 66.4 & 98.8 & \textbf{94.1} & 78.7 \\
& Ours & 62.9 & 95.1 & 68.4 & \textbf{70.4} & \textbf{99.0} & 94.1 & \textbf{81.6} \\
\midrule
\multirow{5}{*}{Pose} 
& Text2Video-Zero~\cite{khachatryan2023text2videozero} & 57.6 & 87.7 & \textbf{100.0} & \textbf{70.7} & 79.7 & 84.8 & 80.1 \\
& ControlVideo~\cite{zhao2023controlvideo} & \textbf{65.4} & 94.6 & 25.0 & 65.3 & 97.3 & 92.8 & 73.4 \\
& Follow-Your-Pose~\cite{ma2024followyourpose} & 48.8 & 86.8 & \textbf{100.0} & 67.4 & 90.1 & 80.2 & 78.9 \\
& VACE & 60.2 & \textbf{94.9} & 75.0 & 64.7 & 98.6 & \textbf{94.8} & 81.4 \\
& Ours & 63.6 & 93.1 & 84.2 & 68.5 & \textbf{98.9} & 94.0 & \textbf{83.7} \\
\midrule
\multirow{3}{*}{Inpainting} 
& ProPainter~\cite{zhou2023propainter} & 44.7 & 95.6 & 50.0 & 61.6 & 99.0 & 93.0 & 74.0 \\
& VACE & 51.3 & \textbf{96.3} & 50.0 & 60.4 & 99.1 & 94.6 & 75.3 \\
& Ours & \textbf{59.7} & 96.3 & \textbf{55.0} & \textbf{68.8} & \textbf{99.3} & \textbf{95.4} & \textbf{79.1} \\
\midrule
\multirow{4}{*}{Outpainting} 
& Follow-Your-Canvas~\cite{chen2024followyourcanvas} & 53.3 & 96.0 & 5.0 & \textbf{69.5} & 98.1 & \textbf{95.4} & 69.5 \\
& M3DDM~\cite{fan2023m3ddm} & 53.3 & 95.9 & 30.0 & 65.1 & 99.2 & 93.7 & 72.9 \\
& VACE & \textbf{57.0} & 96.6 & 30.0 & 69.5 & 99.2 & 94.5 & 74.5 \\
& Ours & 56.1 & \textbf{96.7} & \textbf{45.0} & 68.3 & \textbf{99.4} & 95.4 & \textbf{76.8} \\
\bottomrule
\end{tabular}
\vspace{1mm}
\small

AQ: Aesthetic Quality, BC: Background Consistency, DD: Dynamic Degree, IQ: Imaging Quality, MS: Motion Smoothness, SC: Subject Consistency.
\end{table*}

Table~\ref{tab:vbench_comparison} reports VBench metrics on the VACE Benchmark. Our method achieves the highest average score on all four tasks.

\paragraph{Depth and pose.} Our method outperforms VACE by 2.9 points on depth and 2.3 on pose, while maintaining high dynamic degree (68.4 depth, 84.2 pose) and avoiding the over-constraining failure mode of methods like ControlVideo (DD of 10--25).

\paragraph{Inpainting and outpainting.} We use the same inpainting LoRA for both tasks. Our method outperforms VACE by 3.8~points on inpainting and 2.3~points on outpainting, with gains driven by substantially higher aesthetic quality (+8.4) and imaging quality (+8.4) on inpainting.

\begin{figure*}[t]
\centering
\setlength{\tabcolsep}{0.5pt}
\renewcommand{\arraystretch}{0.3}
\newcommand{\frameimg}[1]{\includegraphics[width=0.08\linewidth]{figures/images/qual_comp/#1}}
\small
\begin{tabular}{@{} *{4}{c} @{\hskip 4pt} *{4}{c} @{\hskip 4pt} *{4}{c} @{}}
\multicolumn{4}{c}{Control Input}
& \multicolumn{4}{c}{\textcolor{blue}{Ours}}
& \multicolumn{4}{c}{\textcolor{red}{VACE}} \\[2pt]
\frameimg{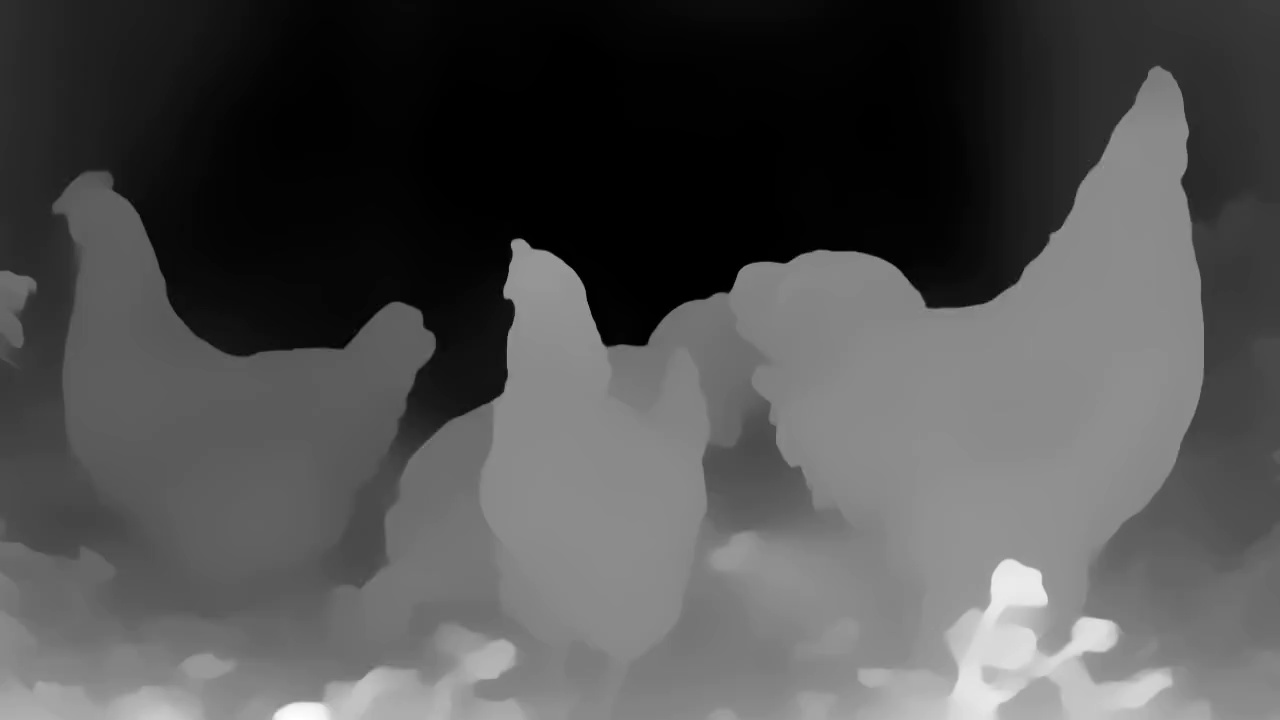} & \frameimg{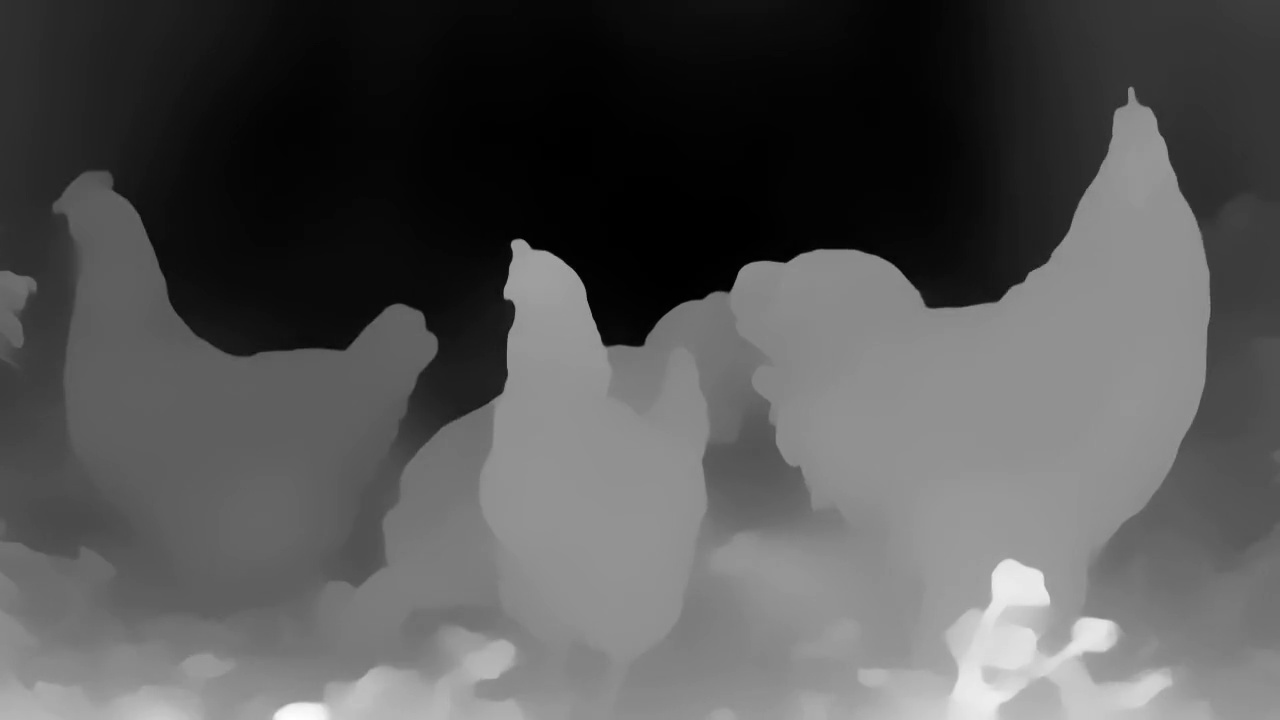} & \frameimg{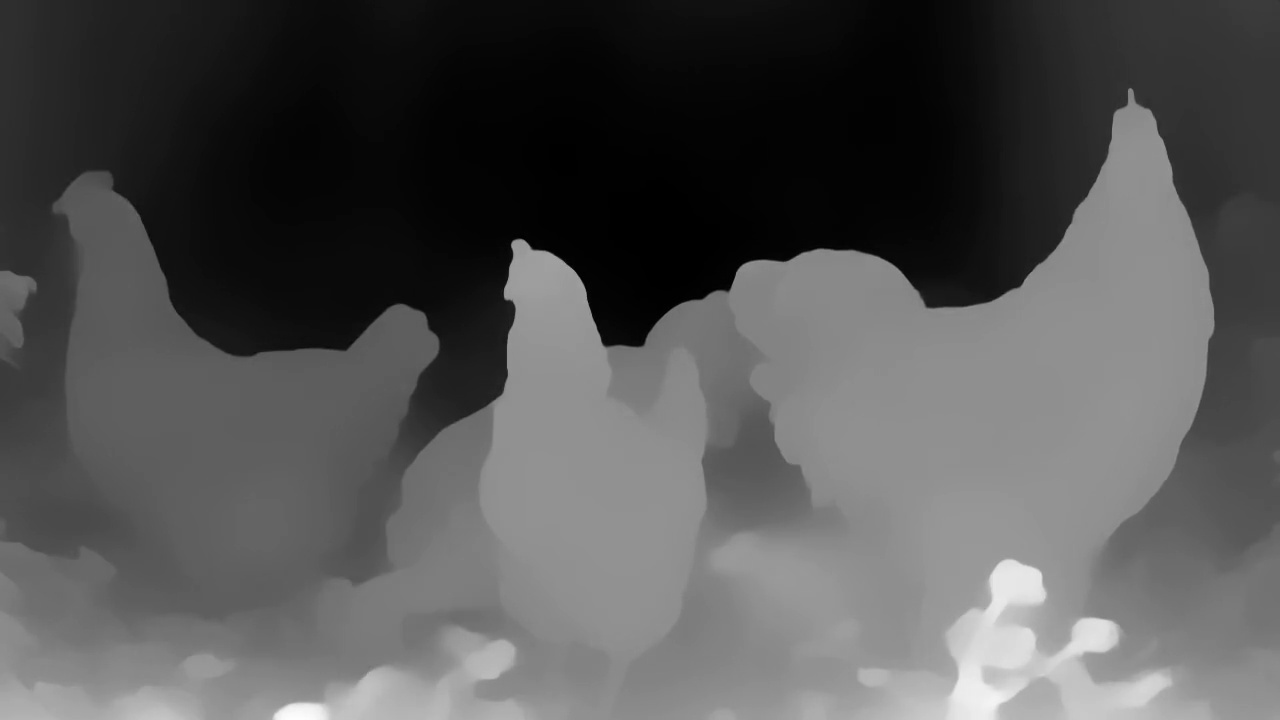} & \frameimg{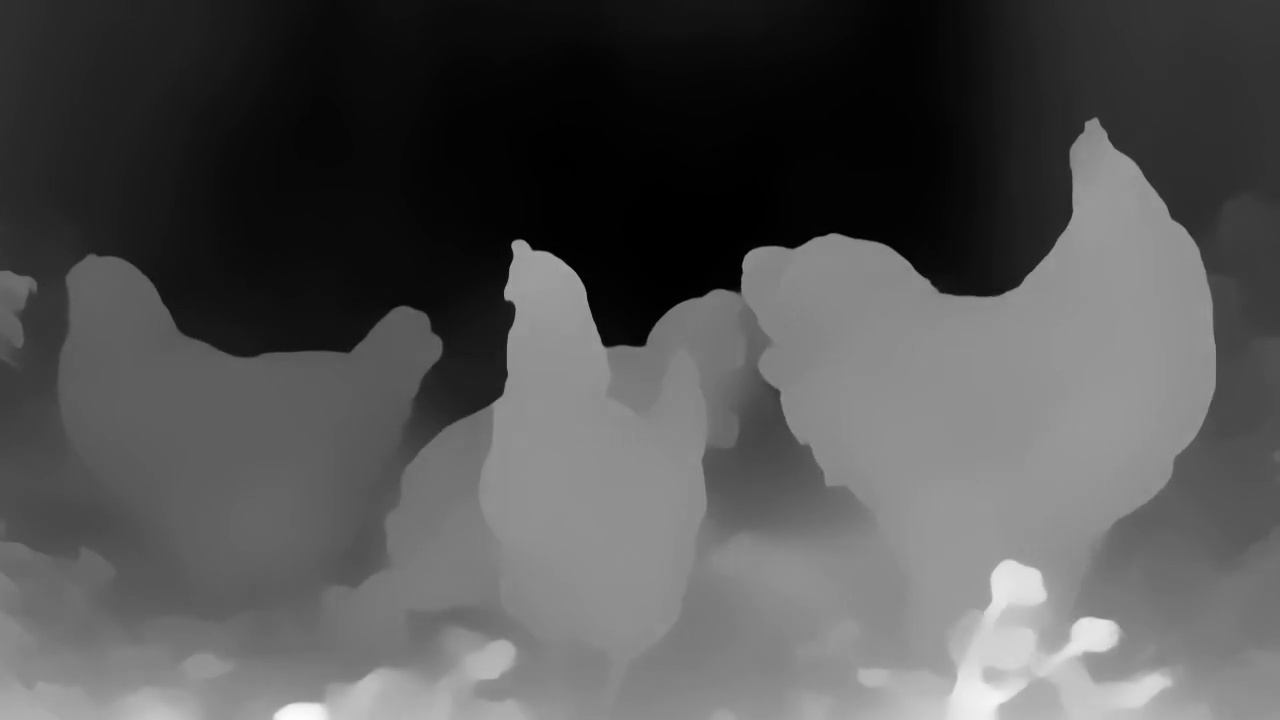} & 
\frameimg{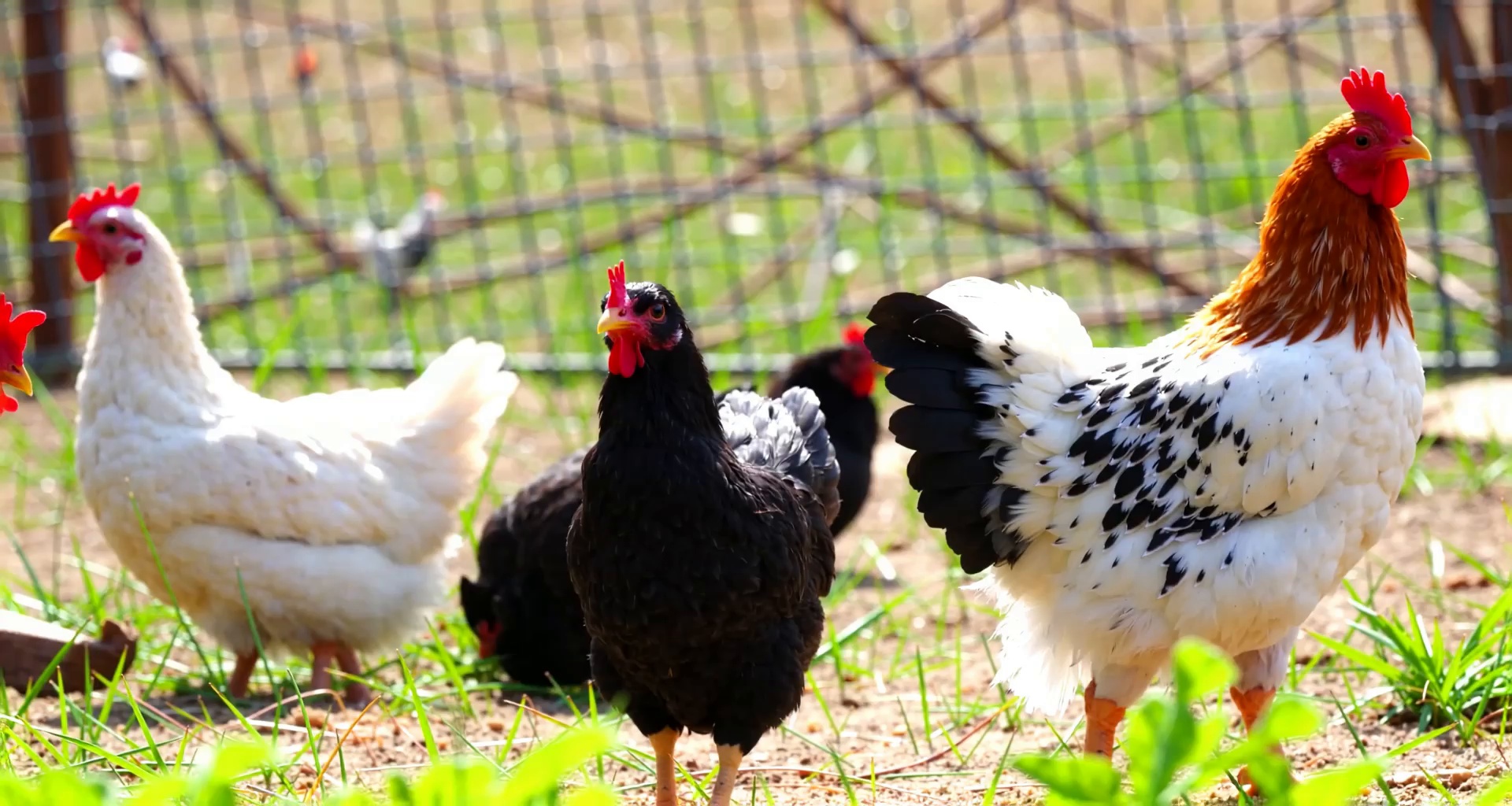} & \frameimg{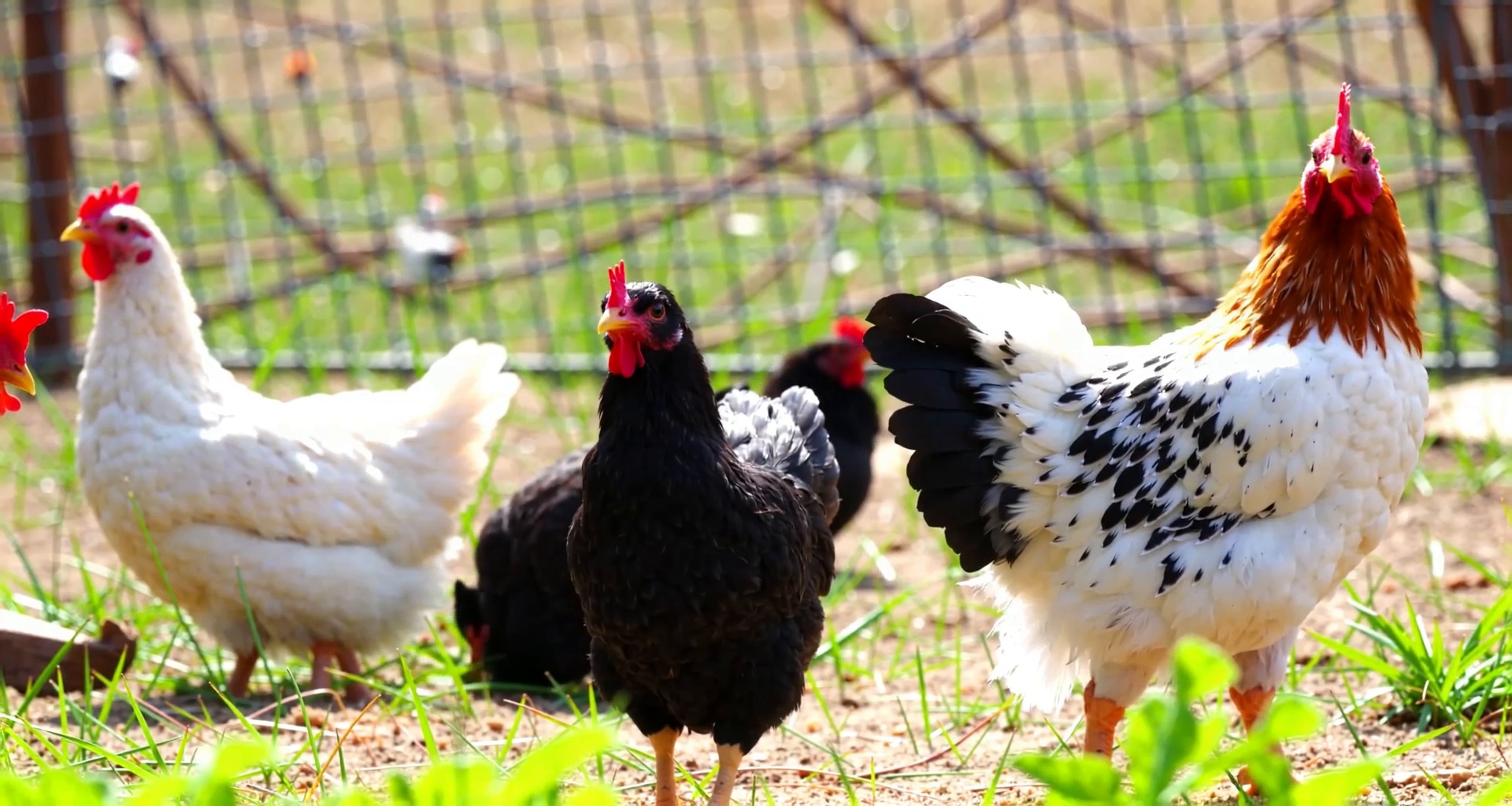} & \frameimg{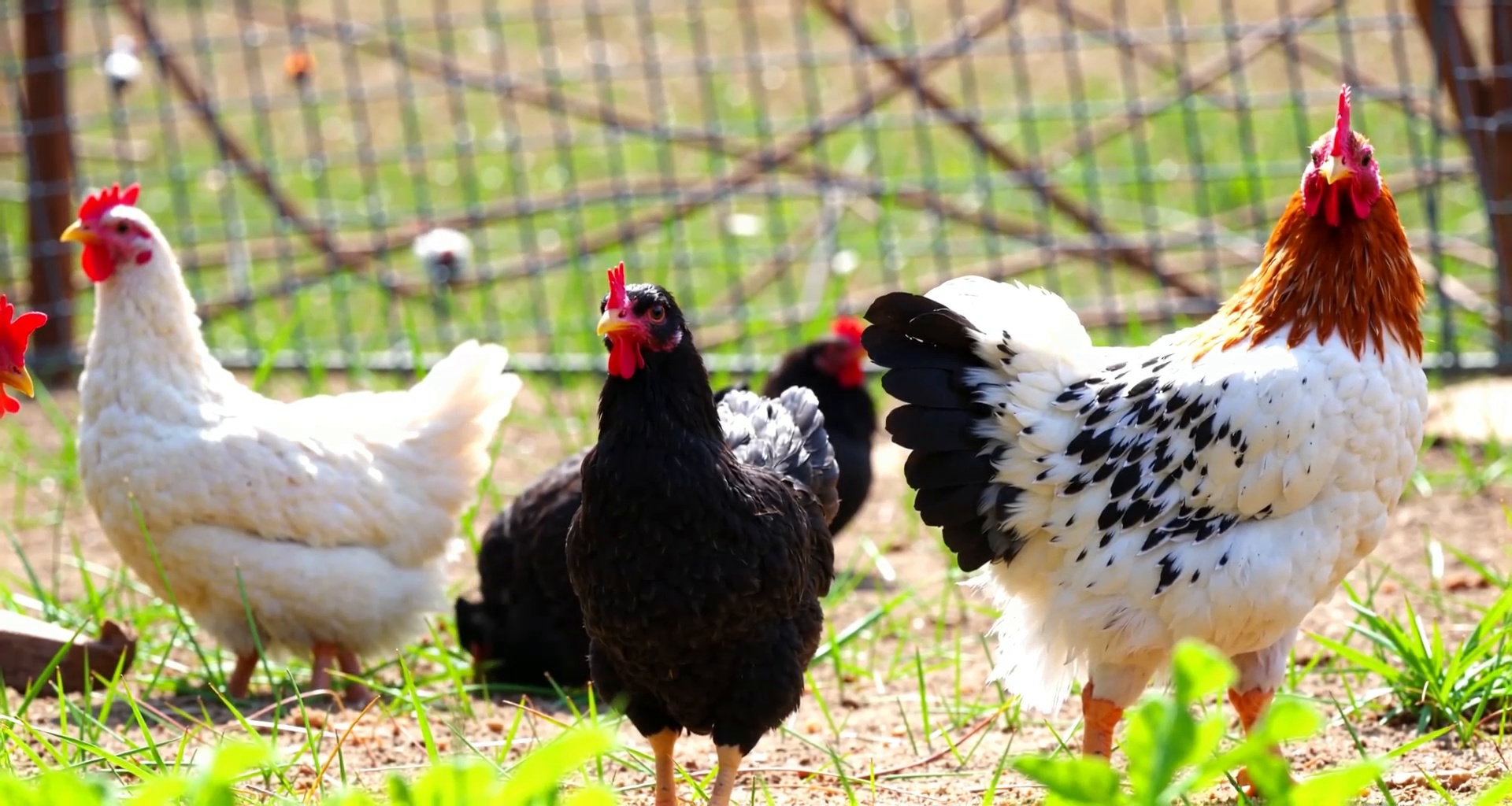} & \frameimg{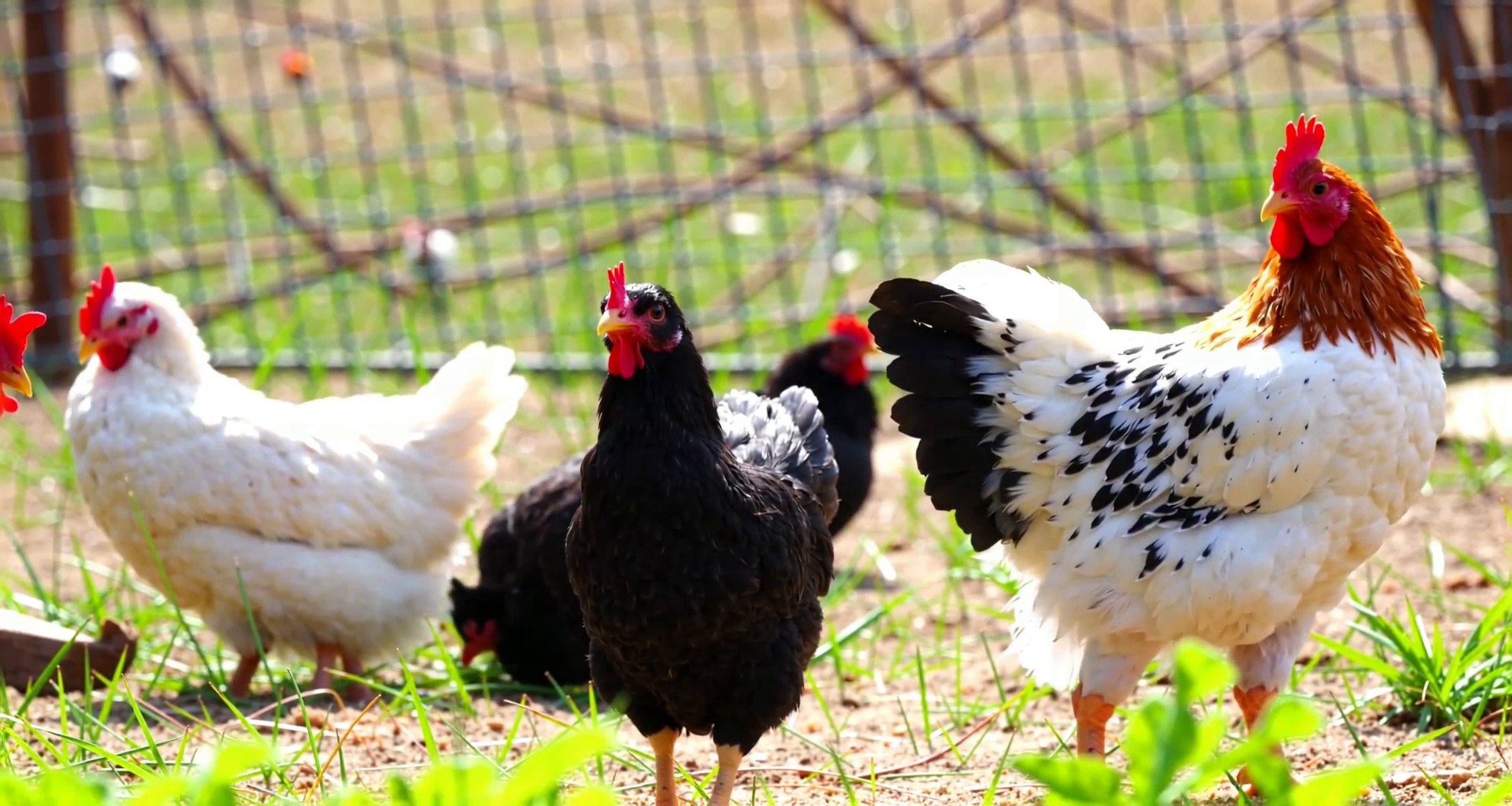} & 
\frameimg{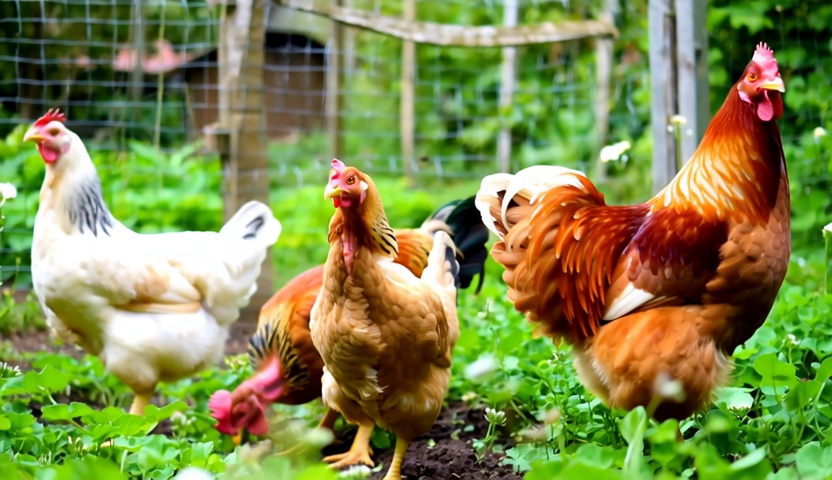} & \frameimg{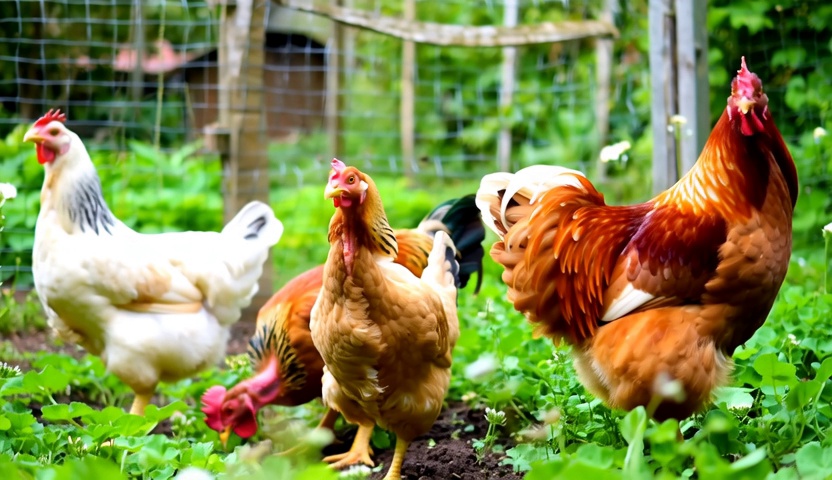} & \frameimg{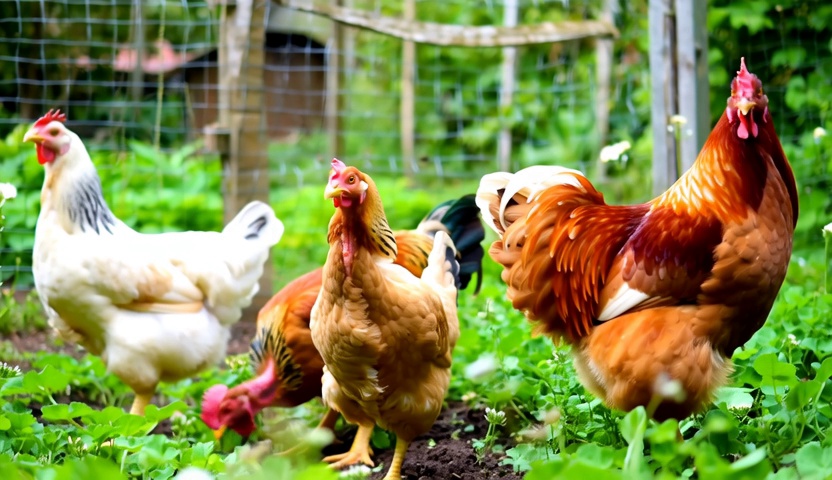} & \frameimg{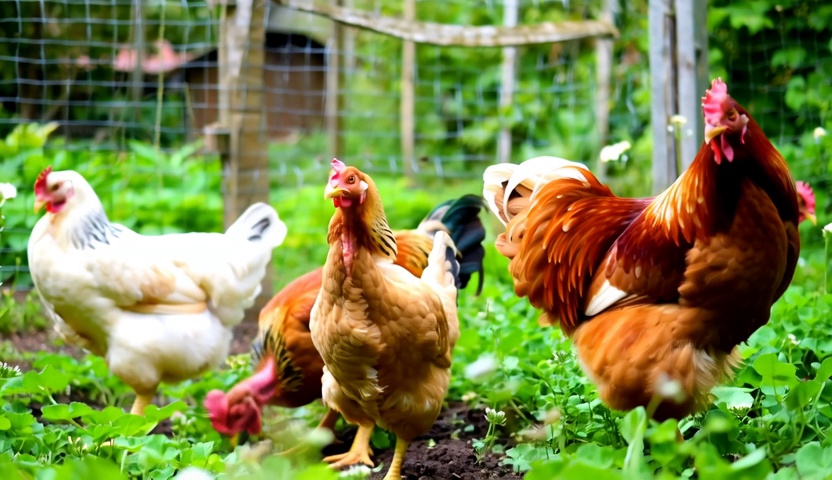} \\ 
\frameimg{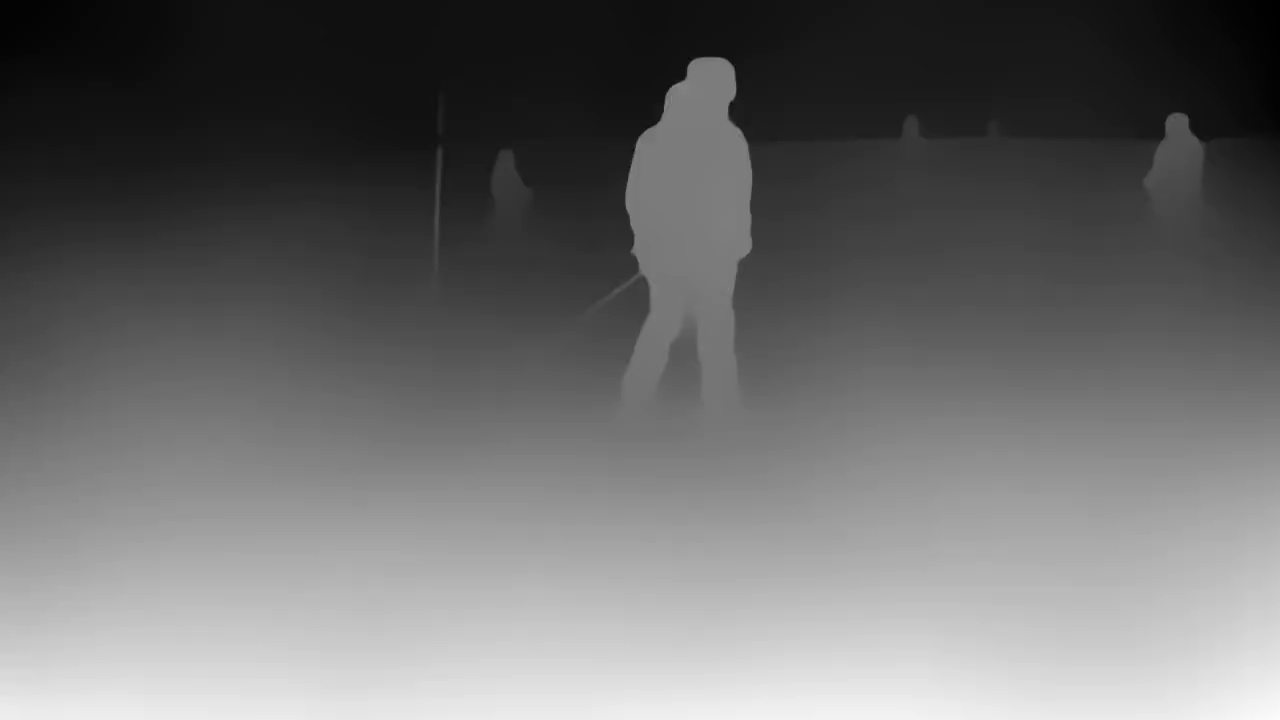} & \frameimg{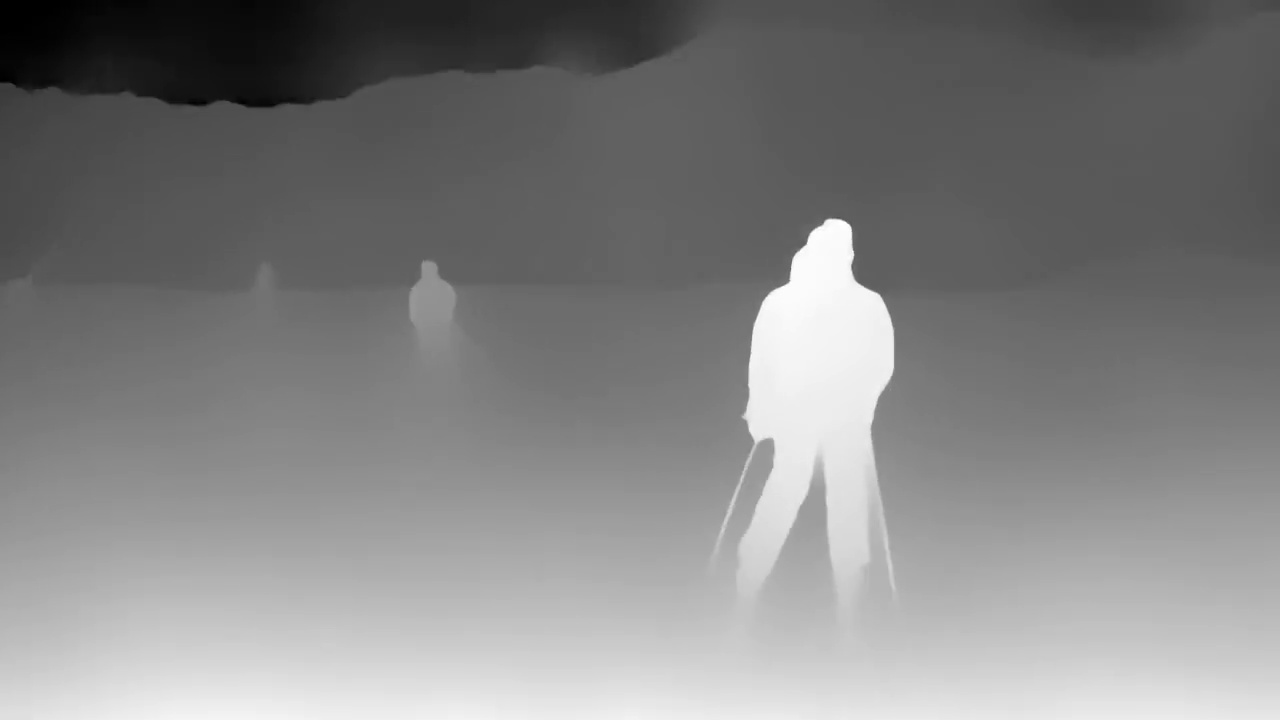} & \frameimg{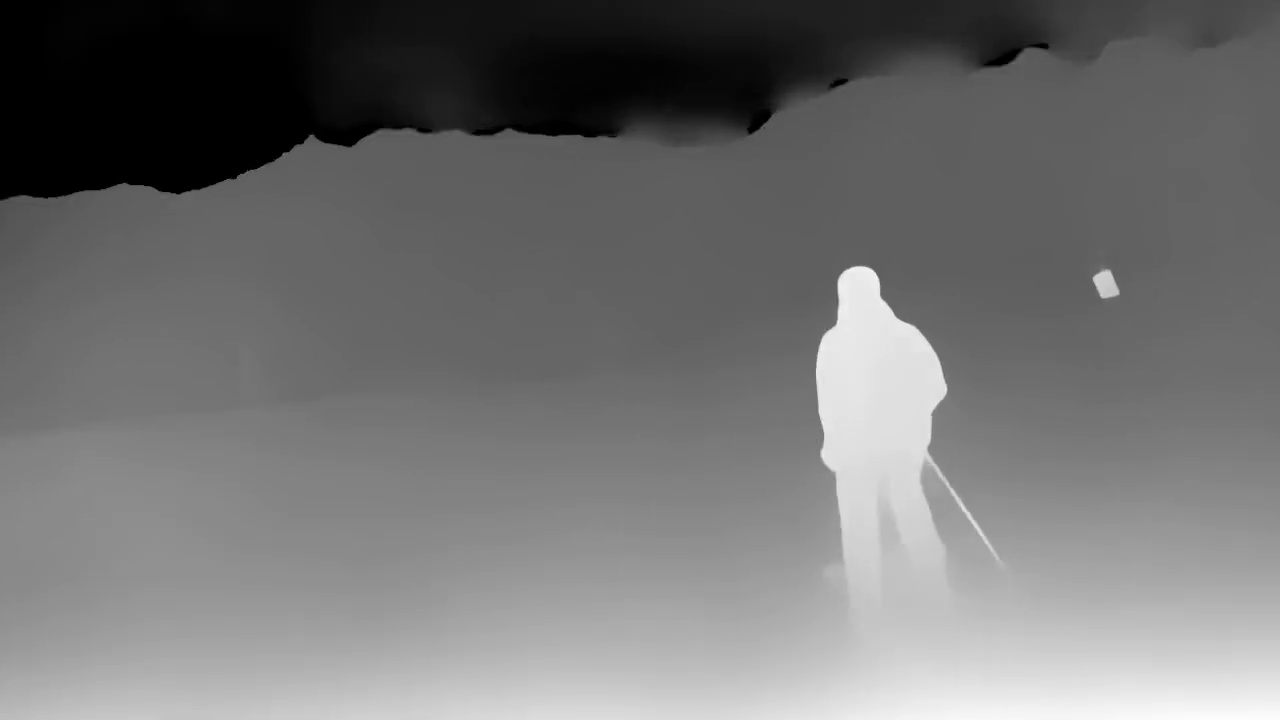} & \frameimg{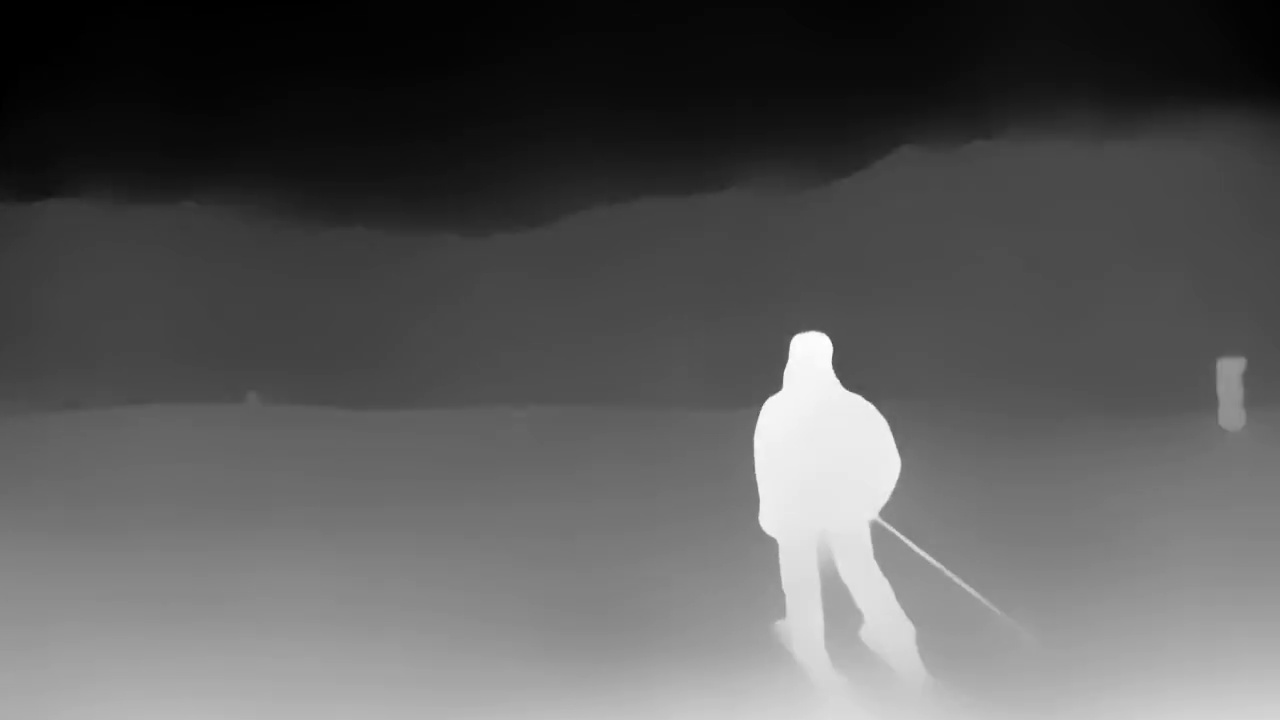} & 
\frameimg{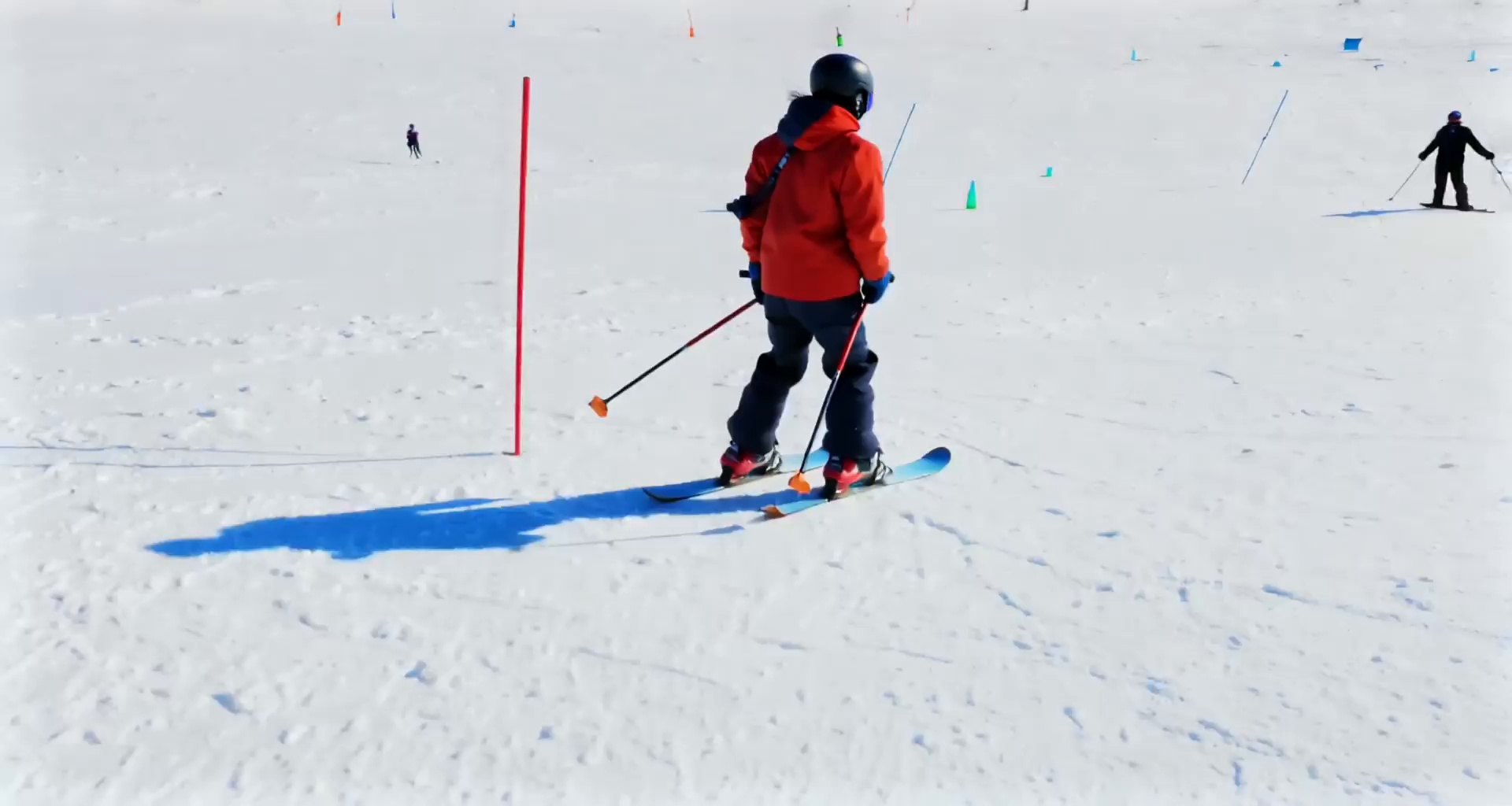} & \frameimg{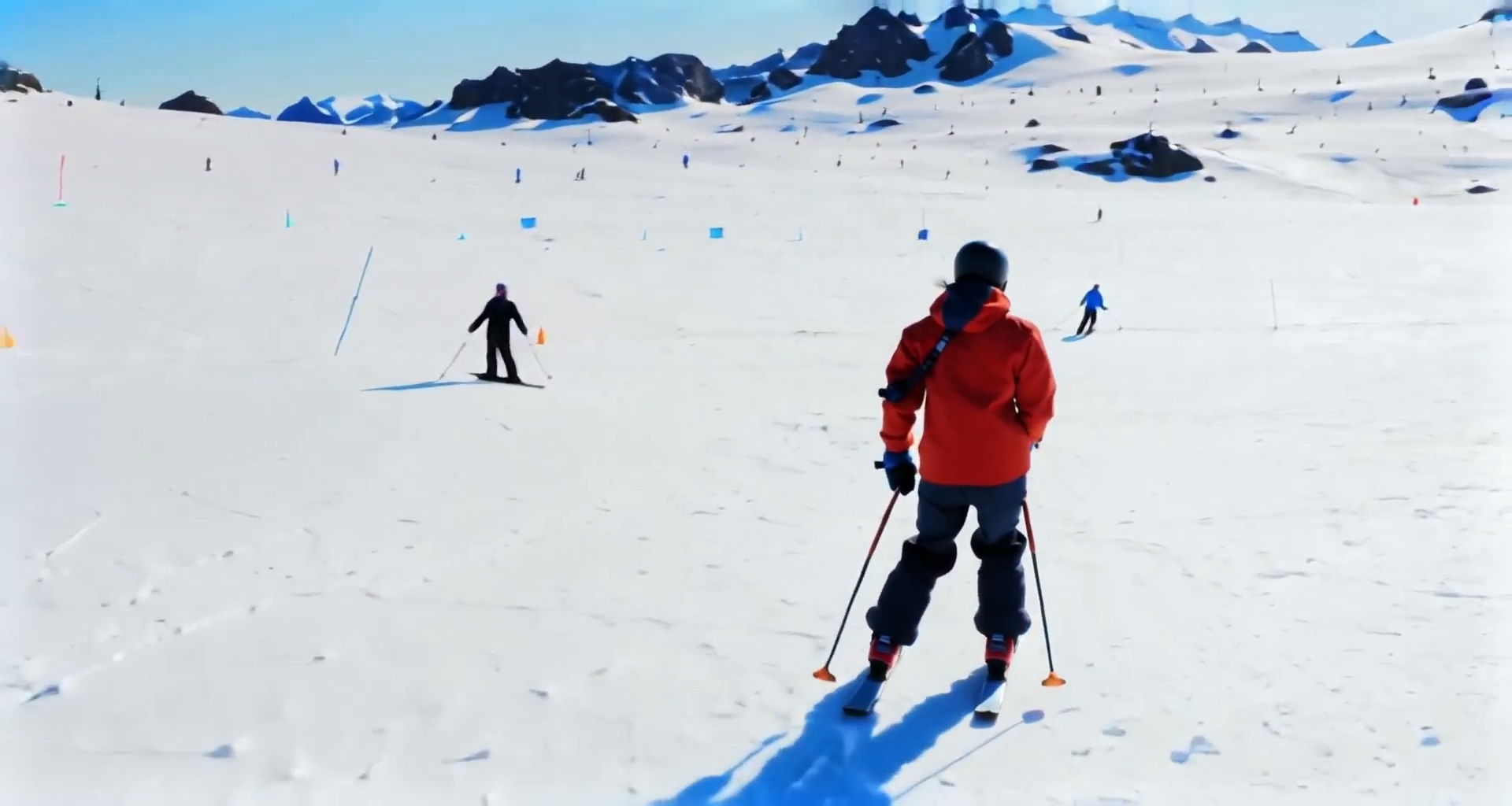} & \frameimg{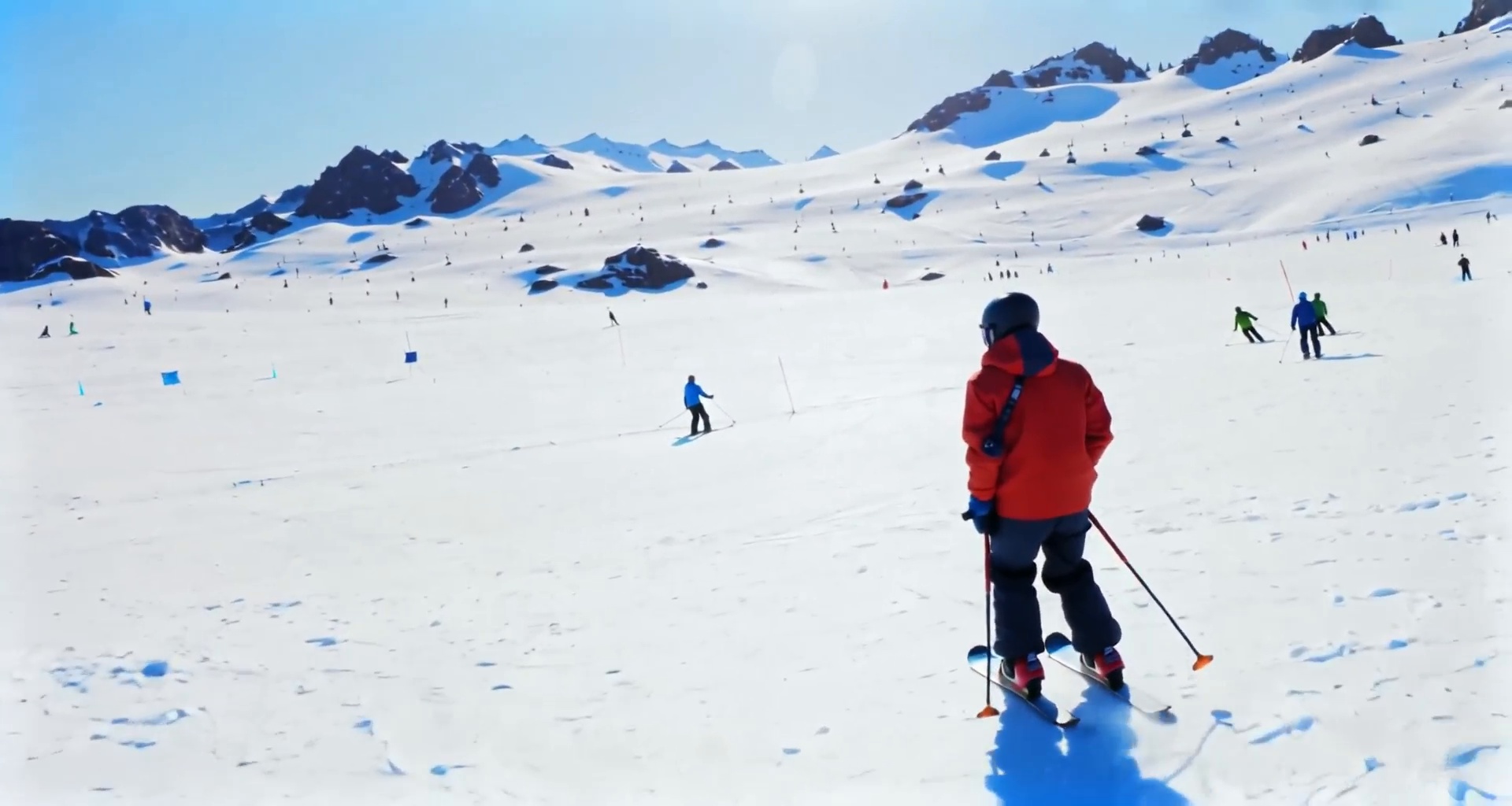} & \frameimg{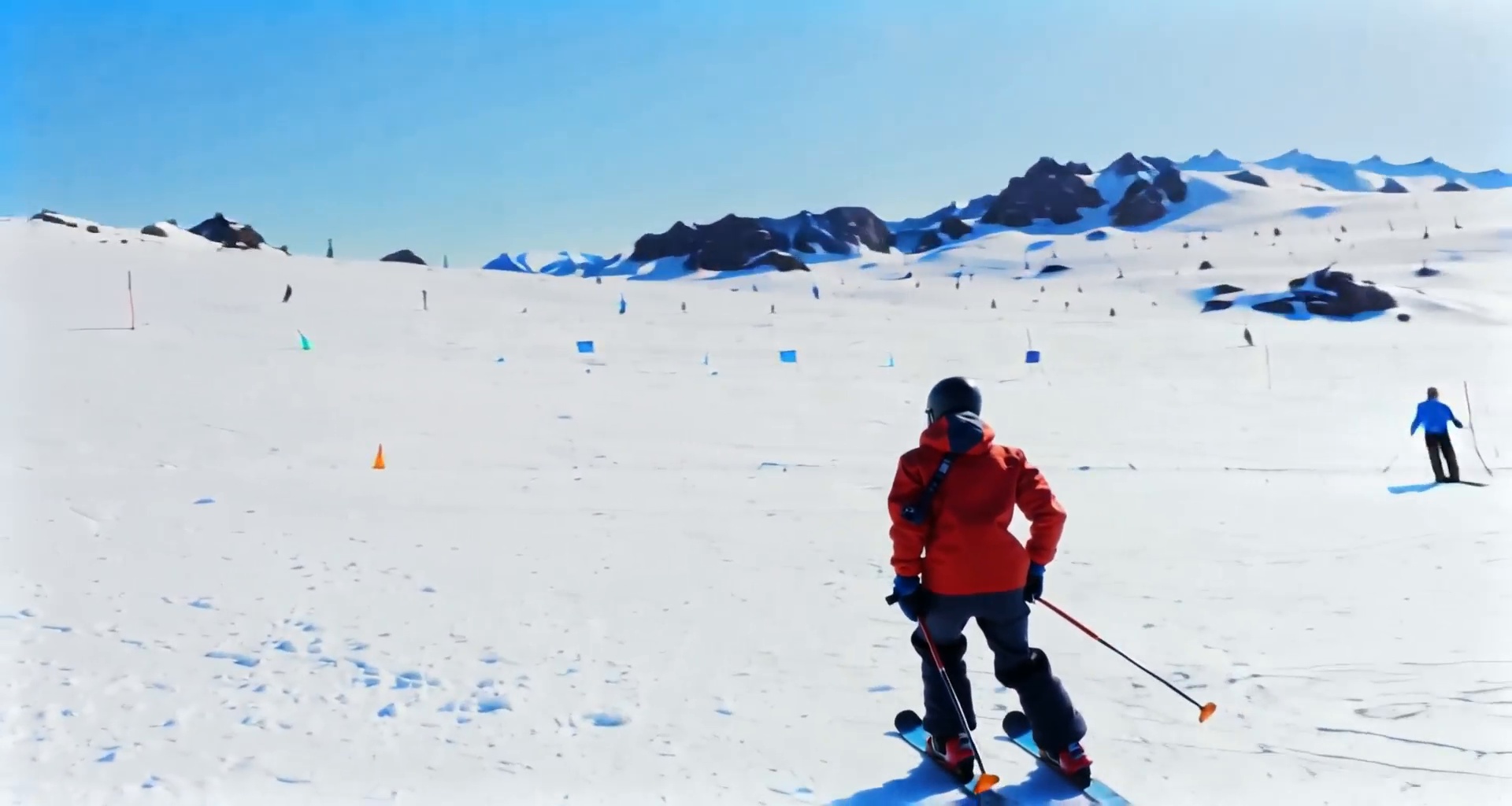} & 
\frameimg{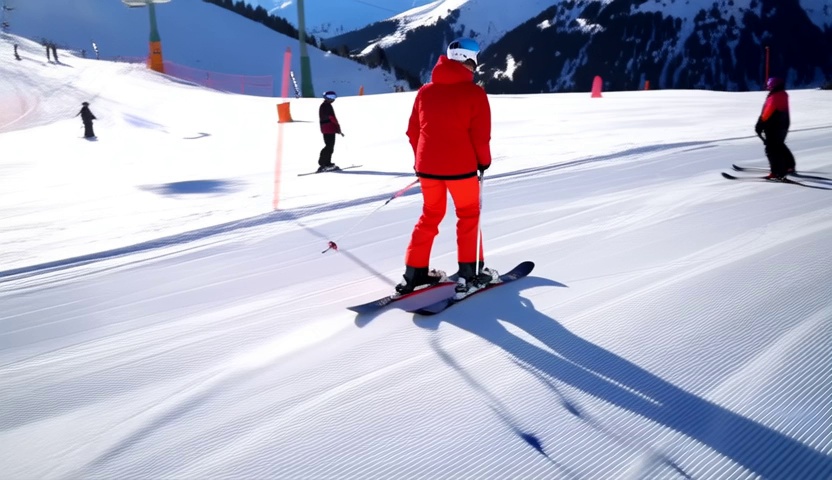} & \frameimg{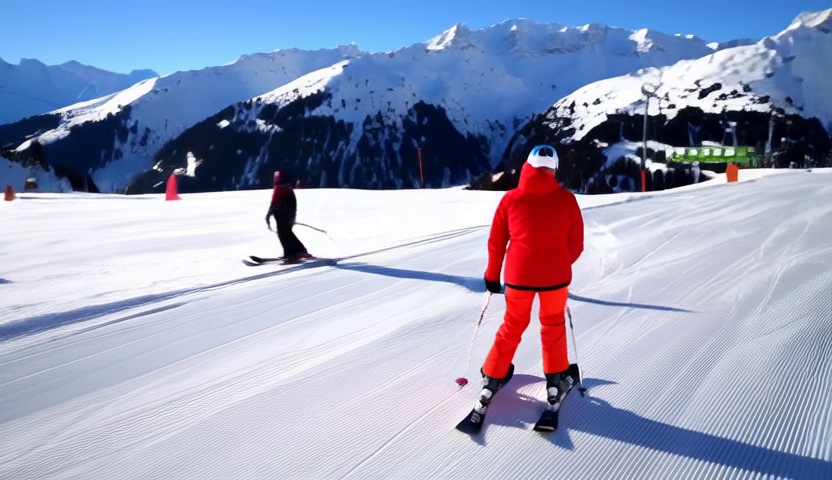} & \frameimg{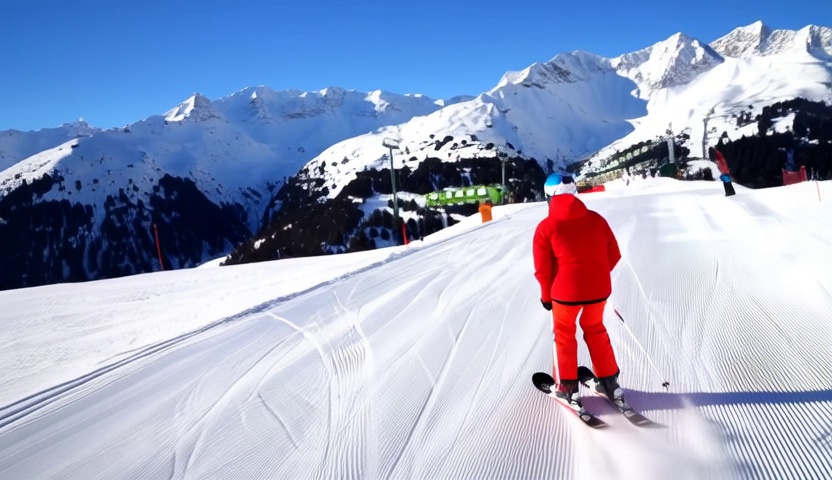} & \frameimg{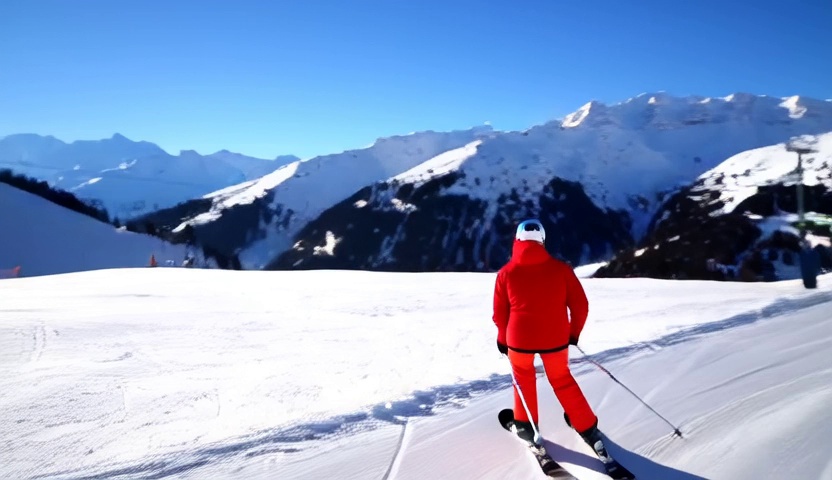} \\ 
\frameimg{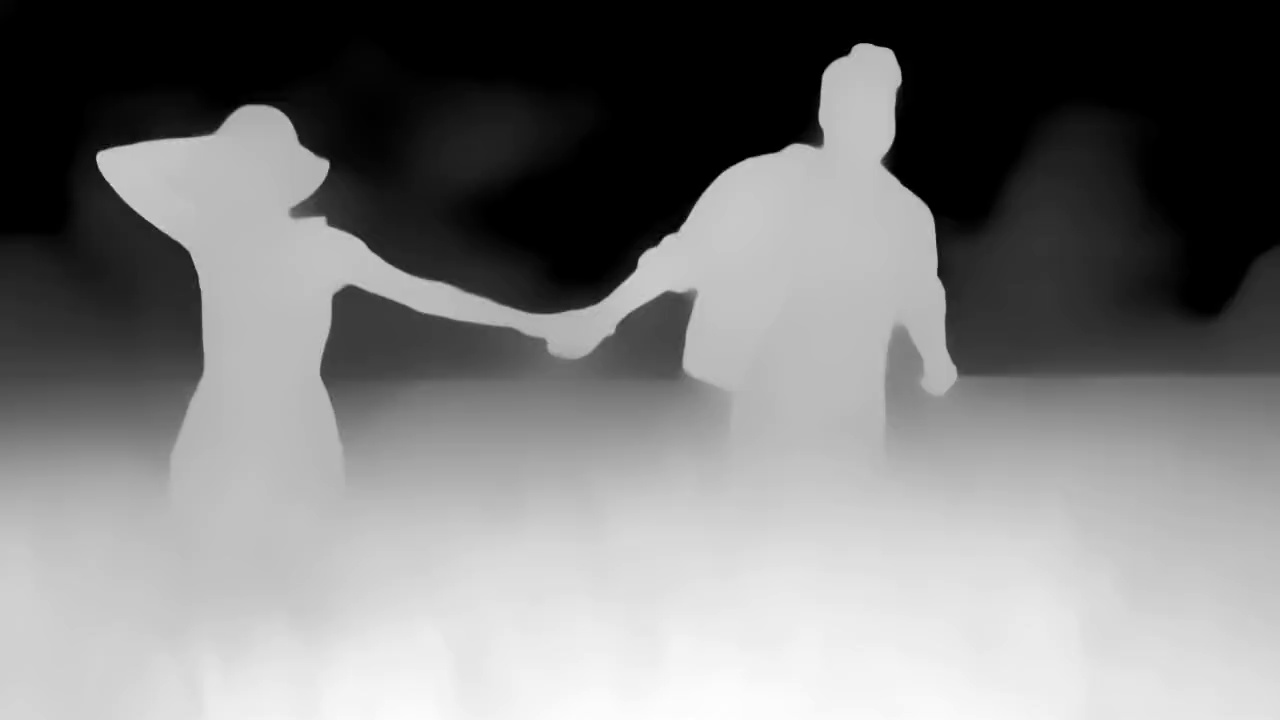} & \frameimg{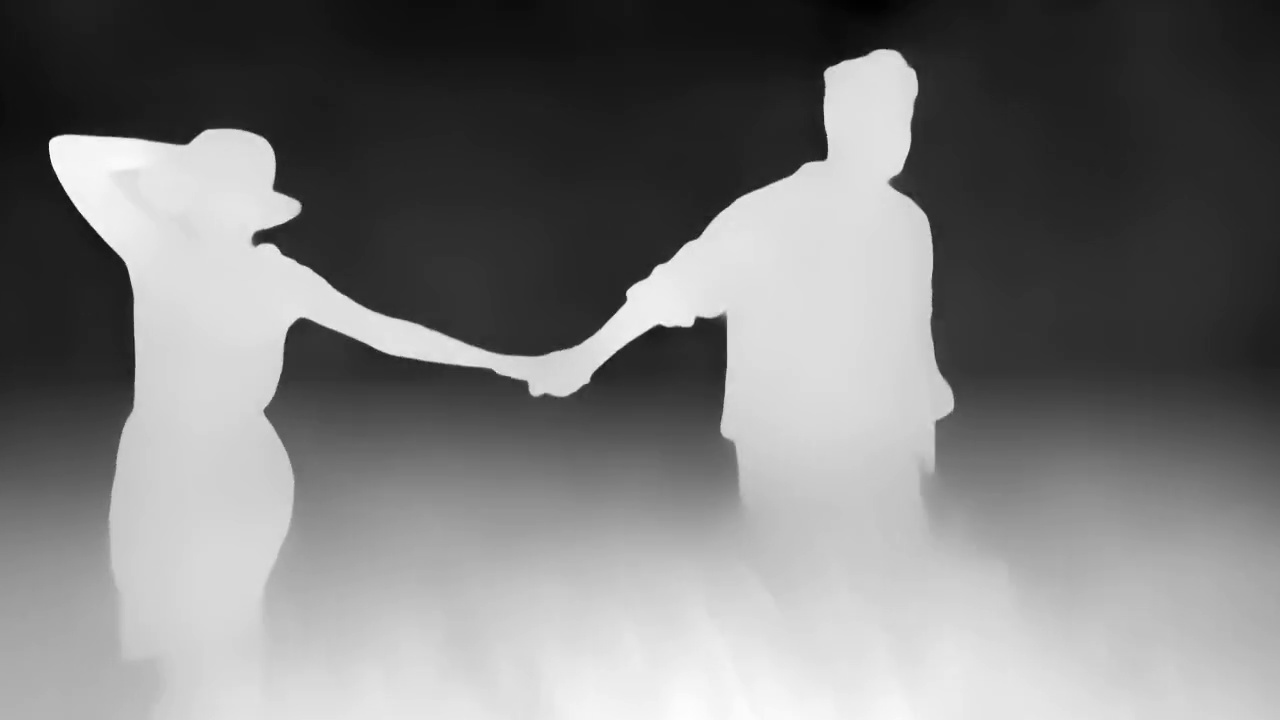} & \frameimg{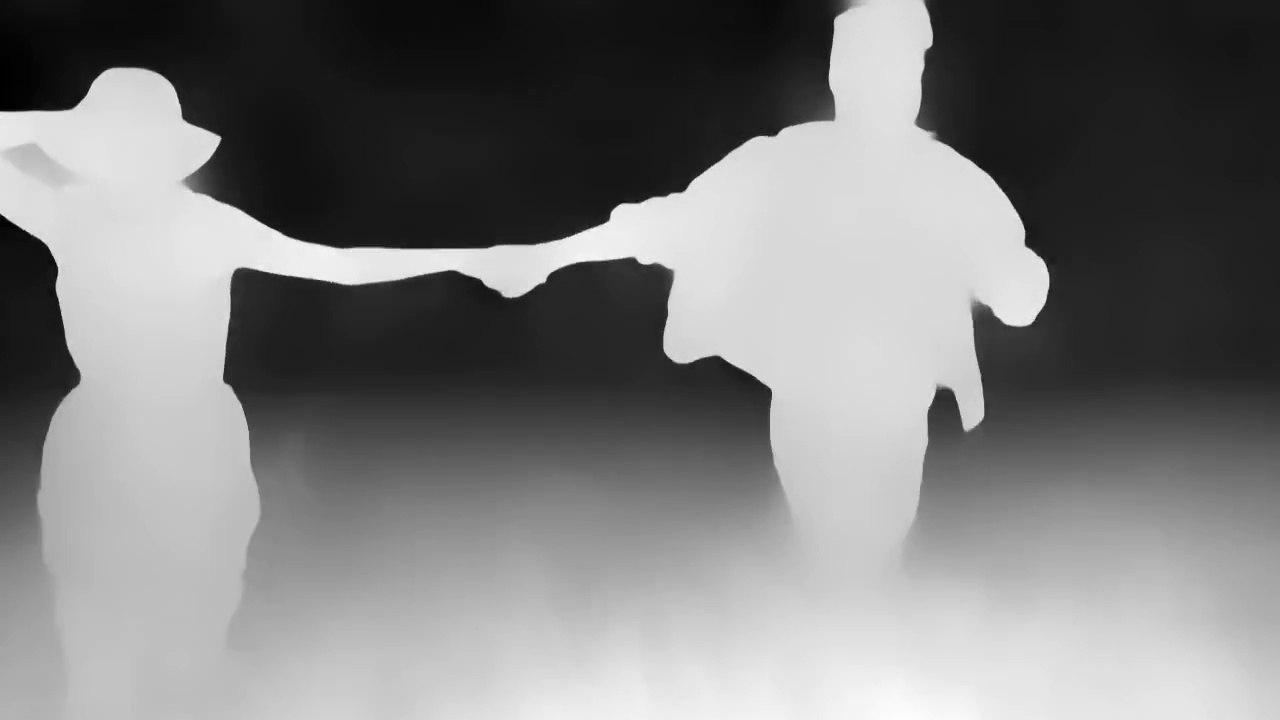} & \frameimg{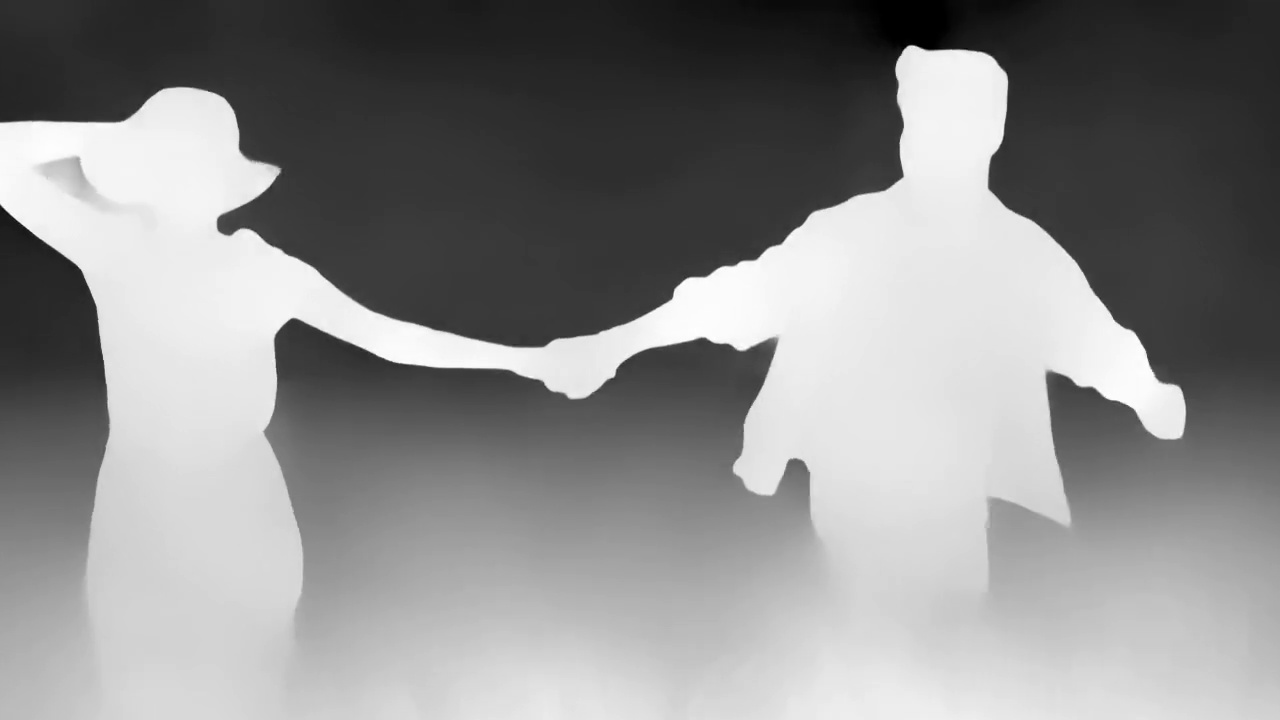} & 
\frameimg{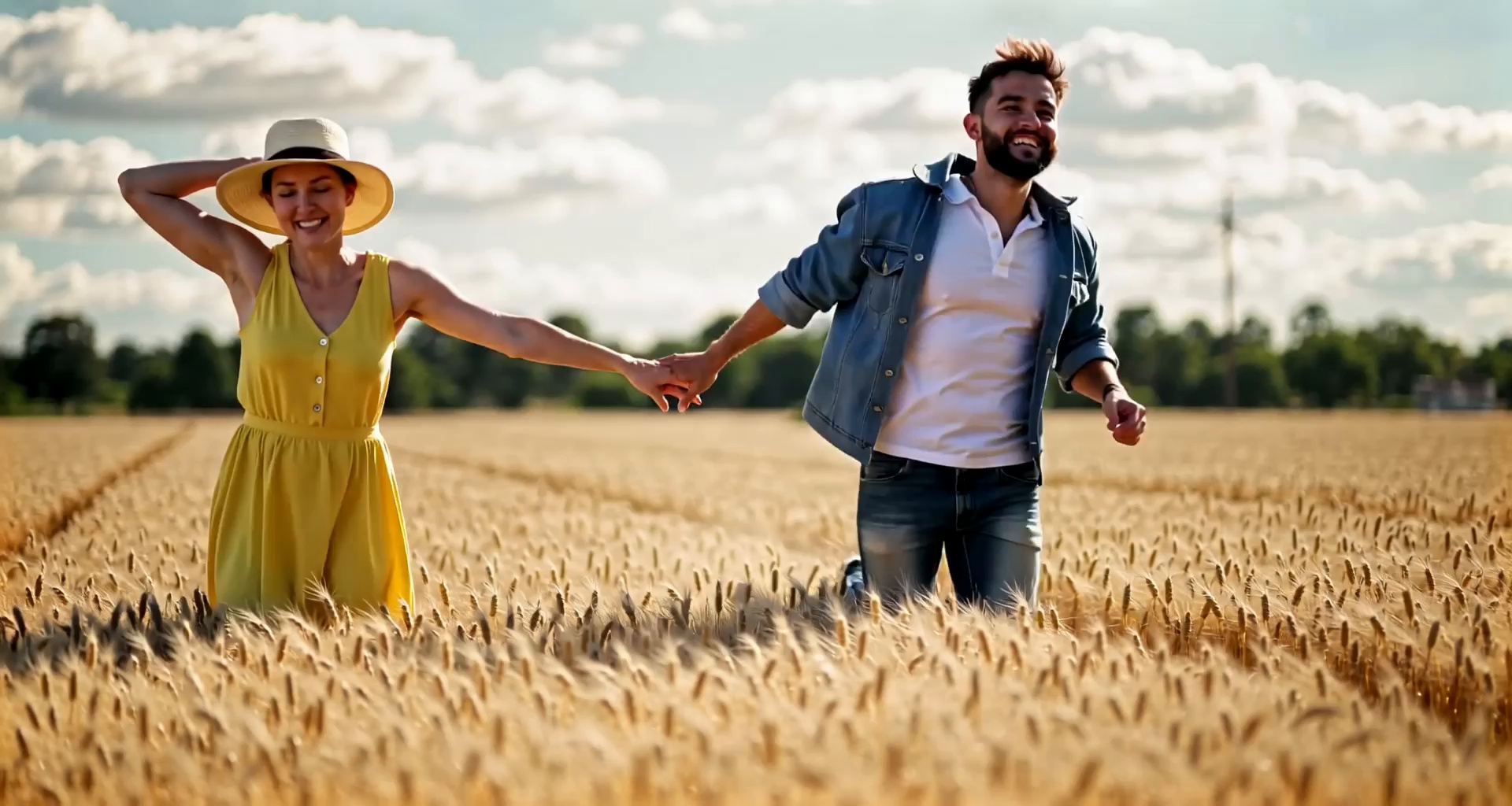} & \frameimg{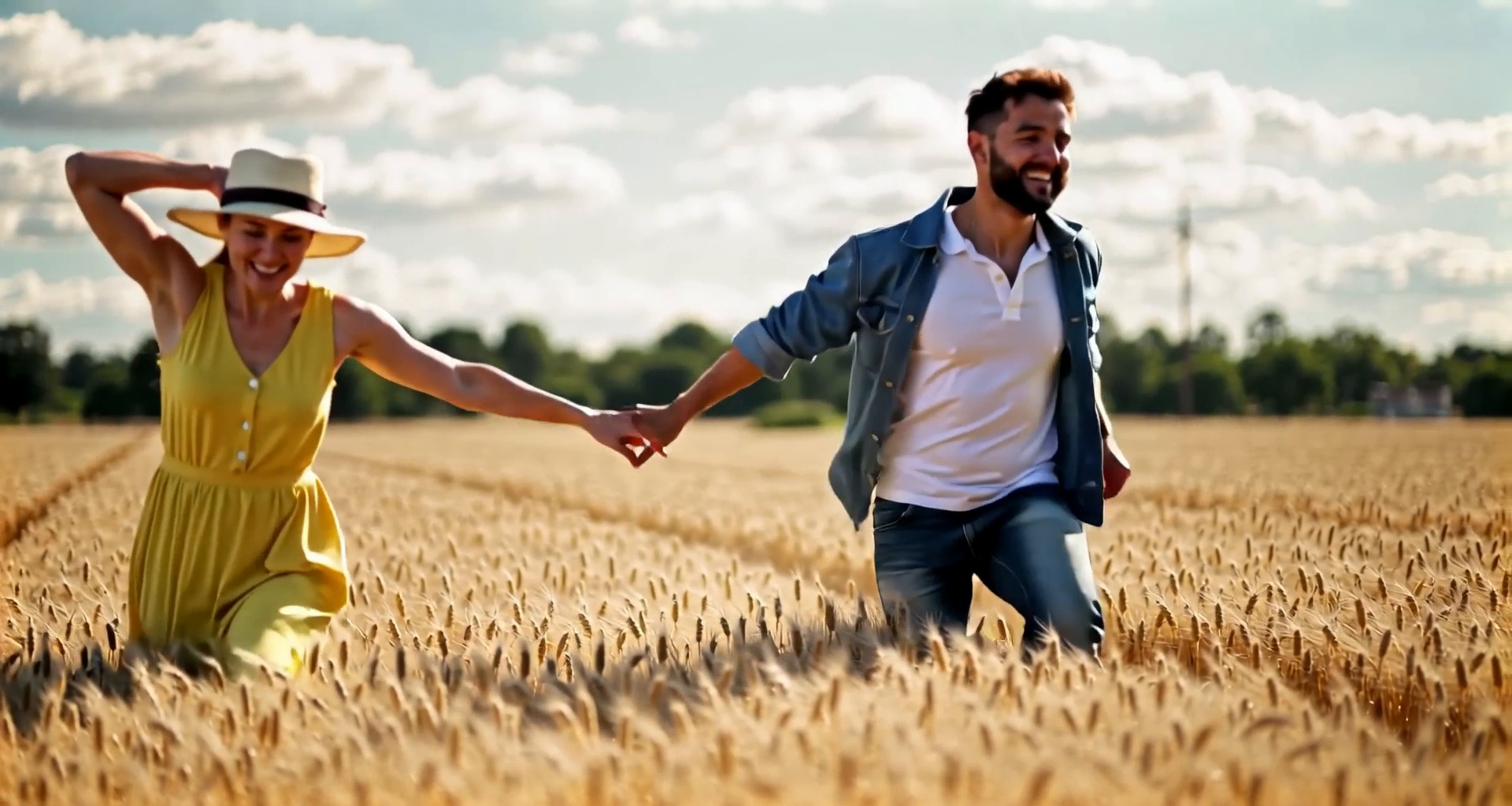} & \frameimg{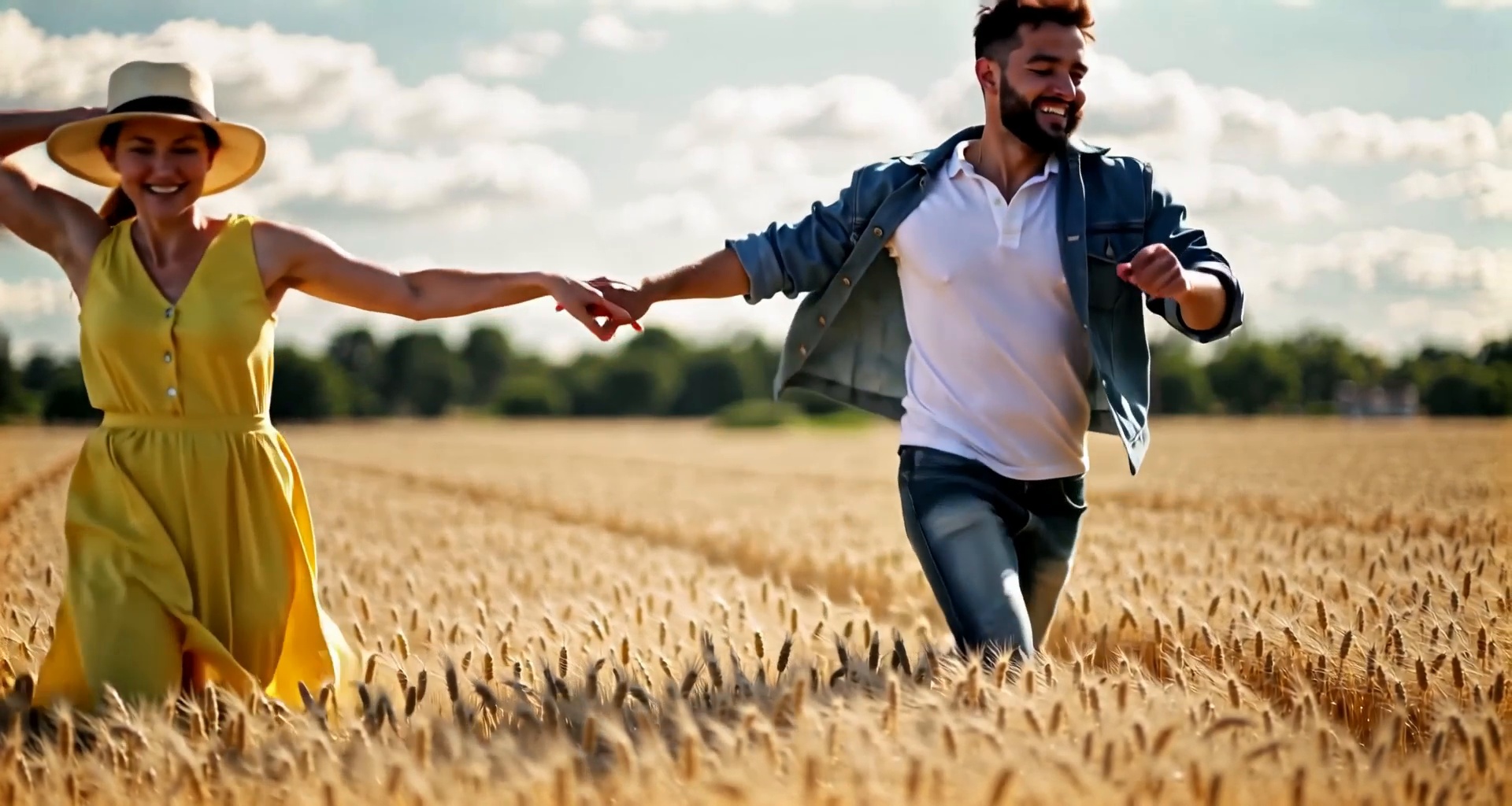} & \frameimg{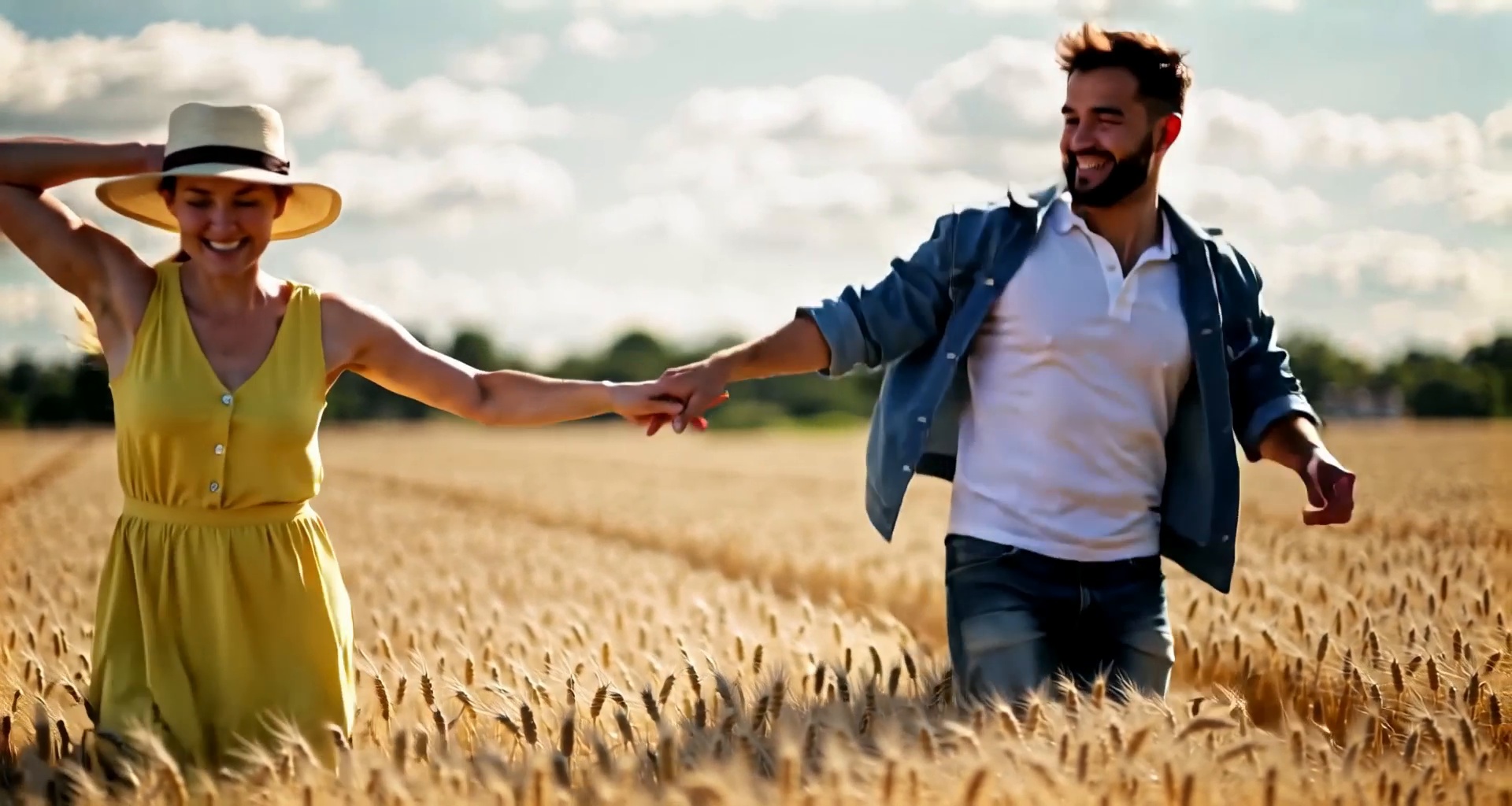} & 
\frameimg{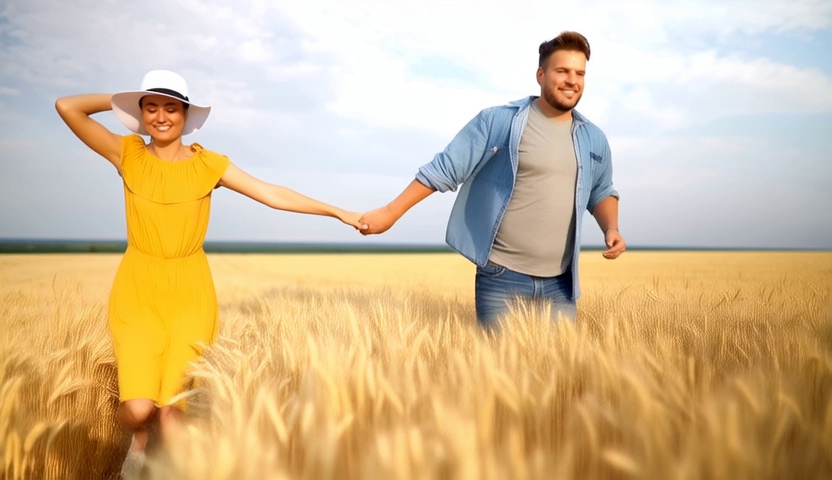} & \frameimg{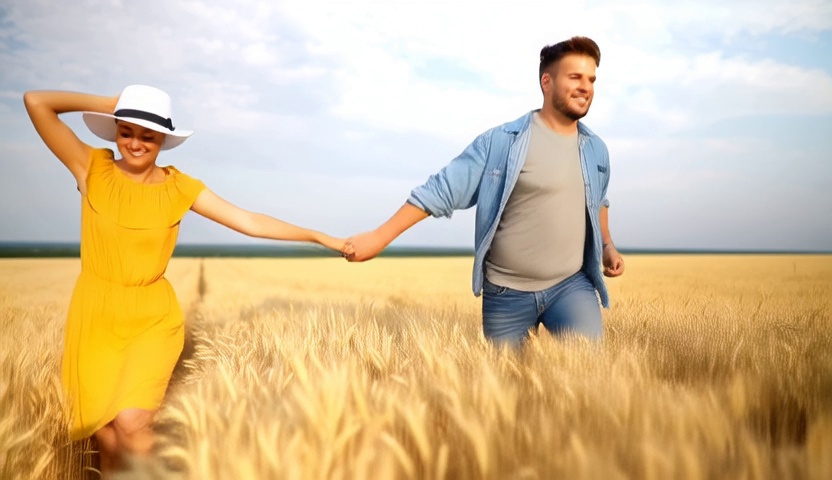} & \frameimg{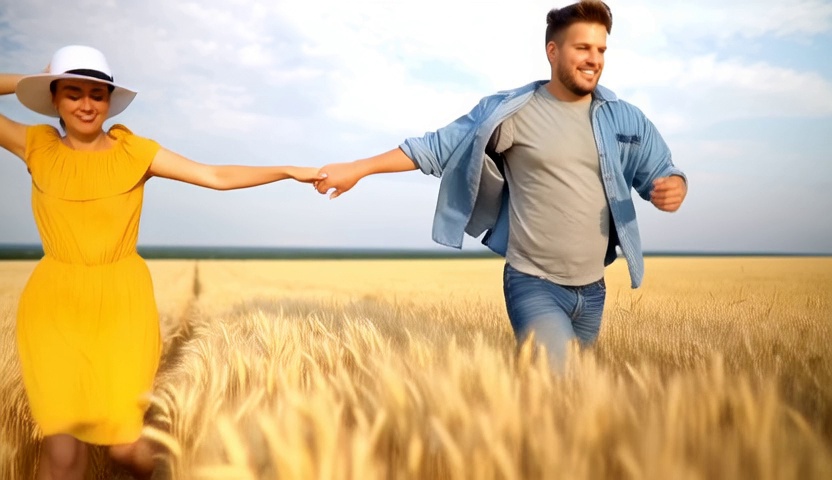} & \frameimg{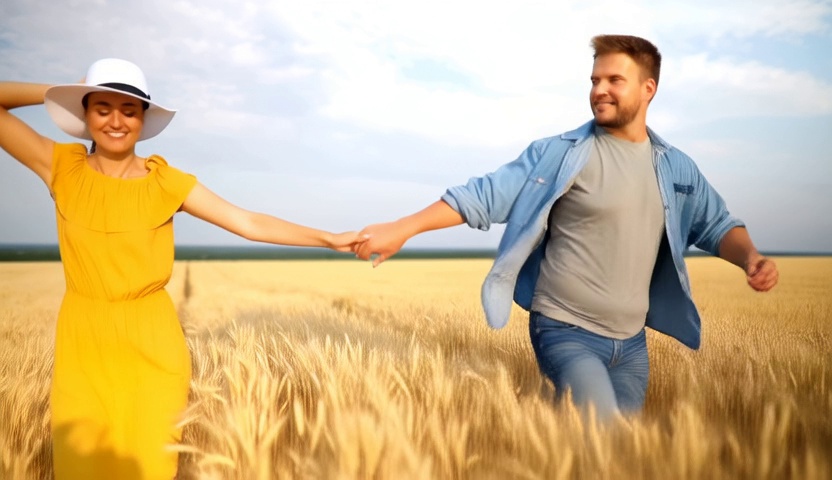} \\[4pt] 
\frameimg{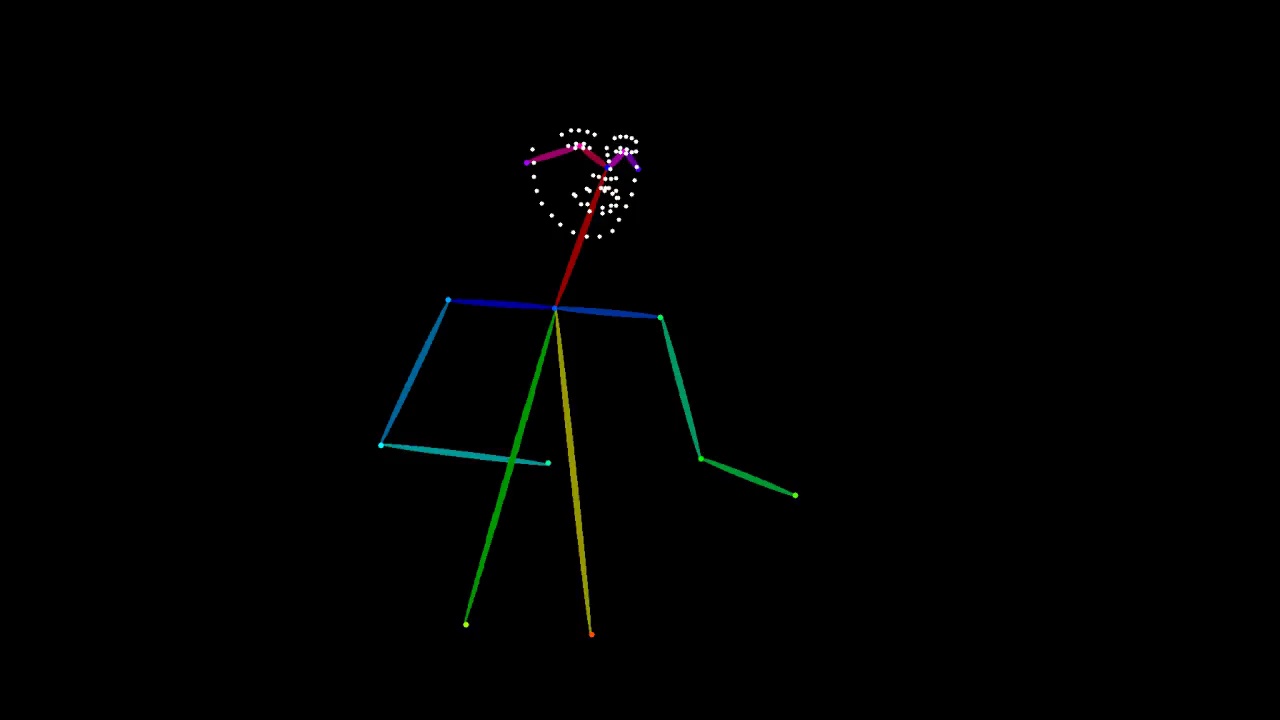} & \frameimg{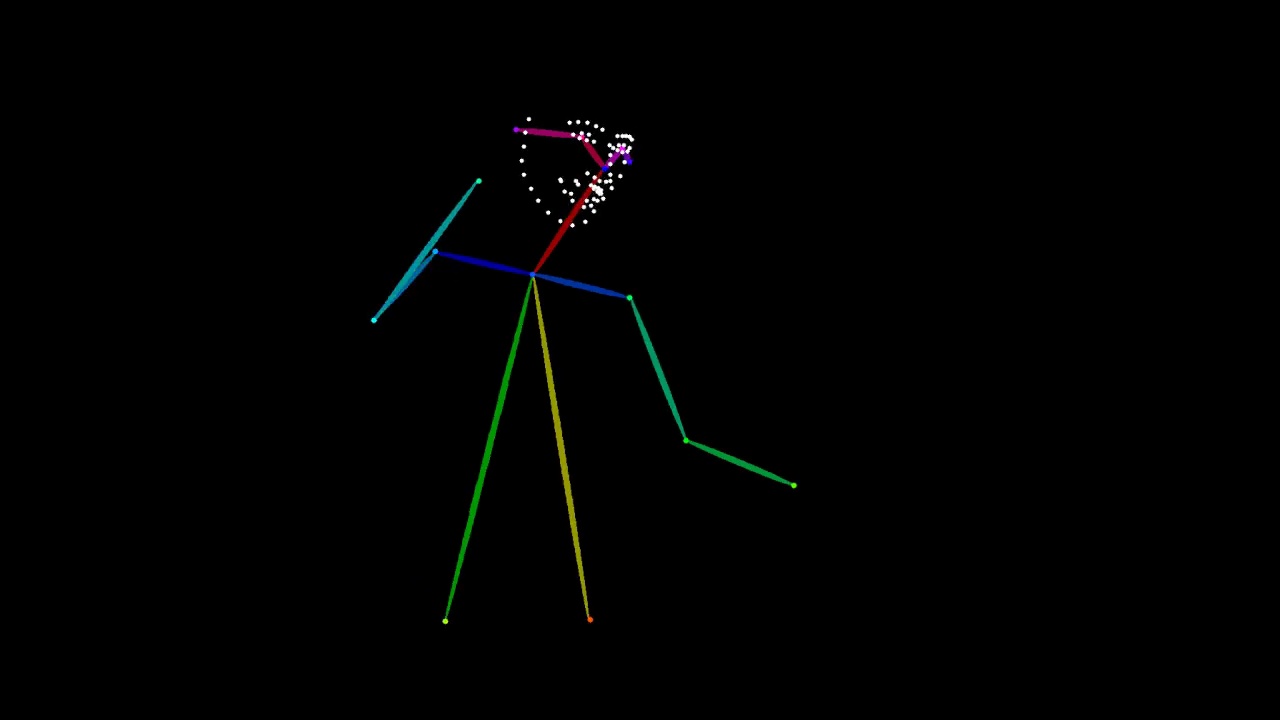} & \frameimg{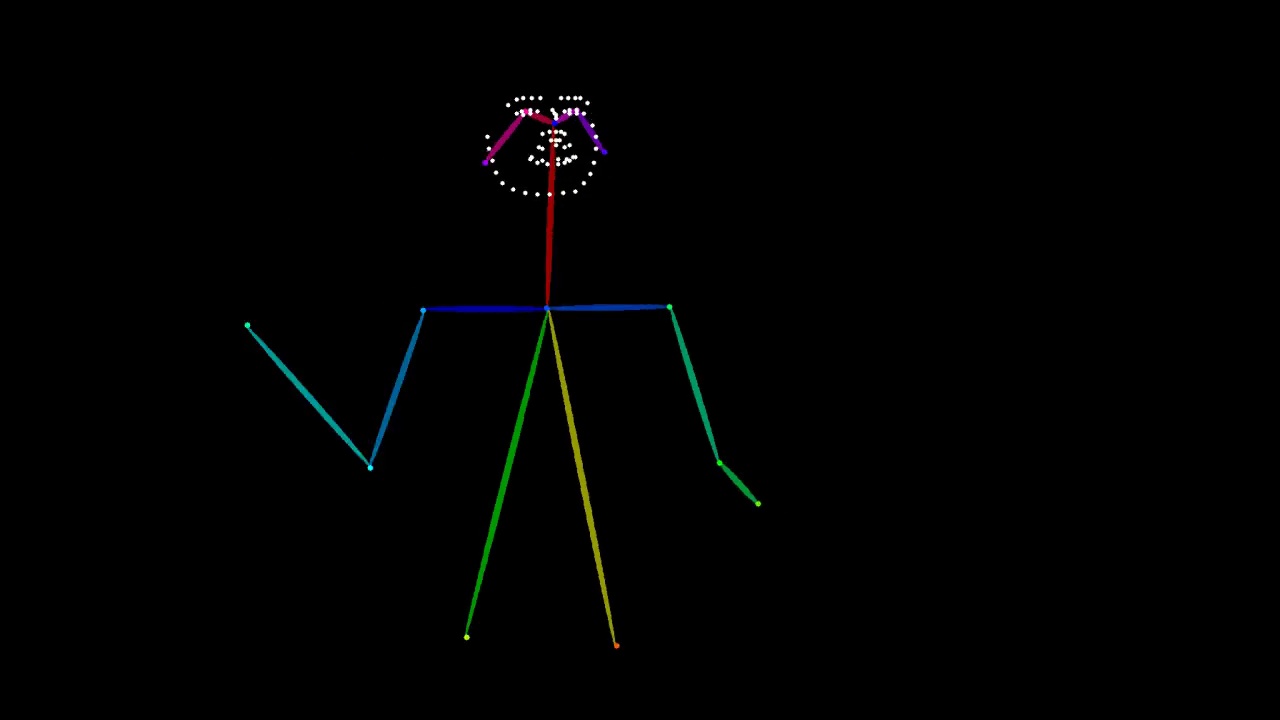} & \frameimg{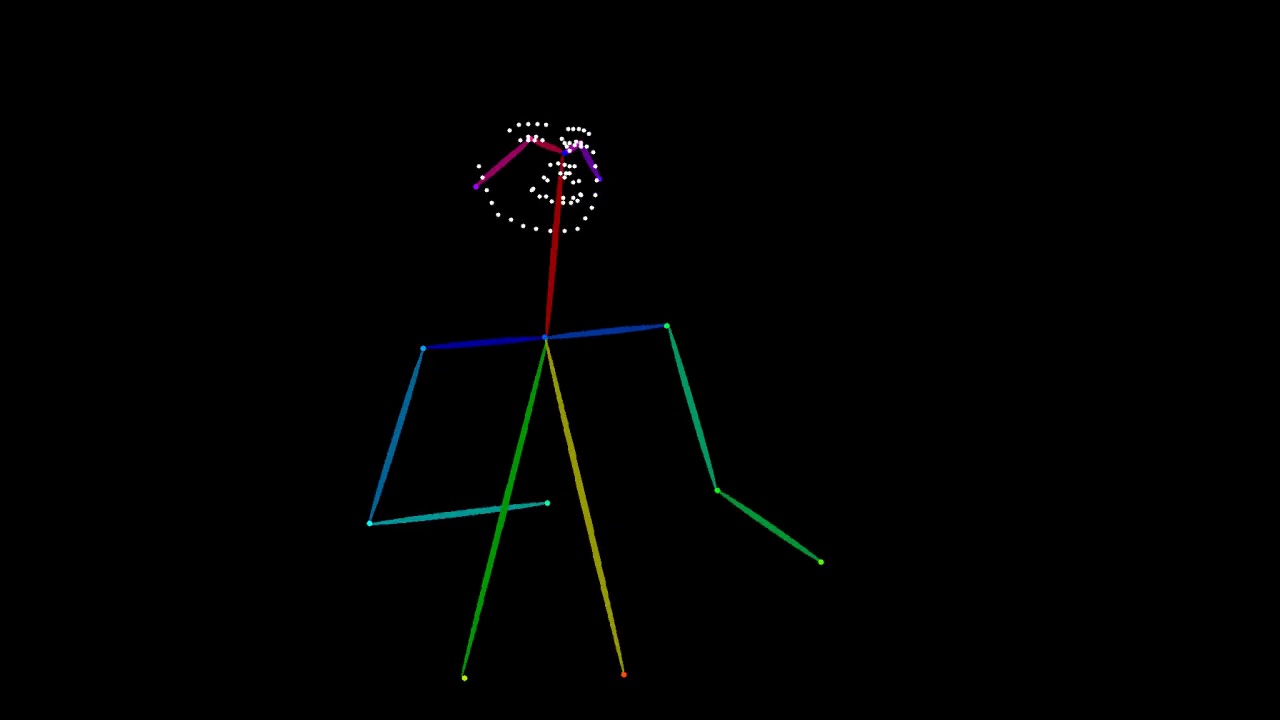} & 
\frameimg{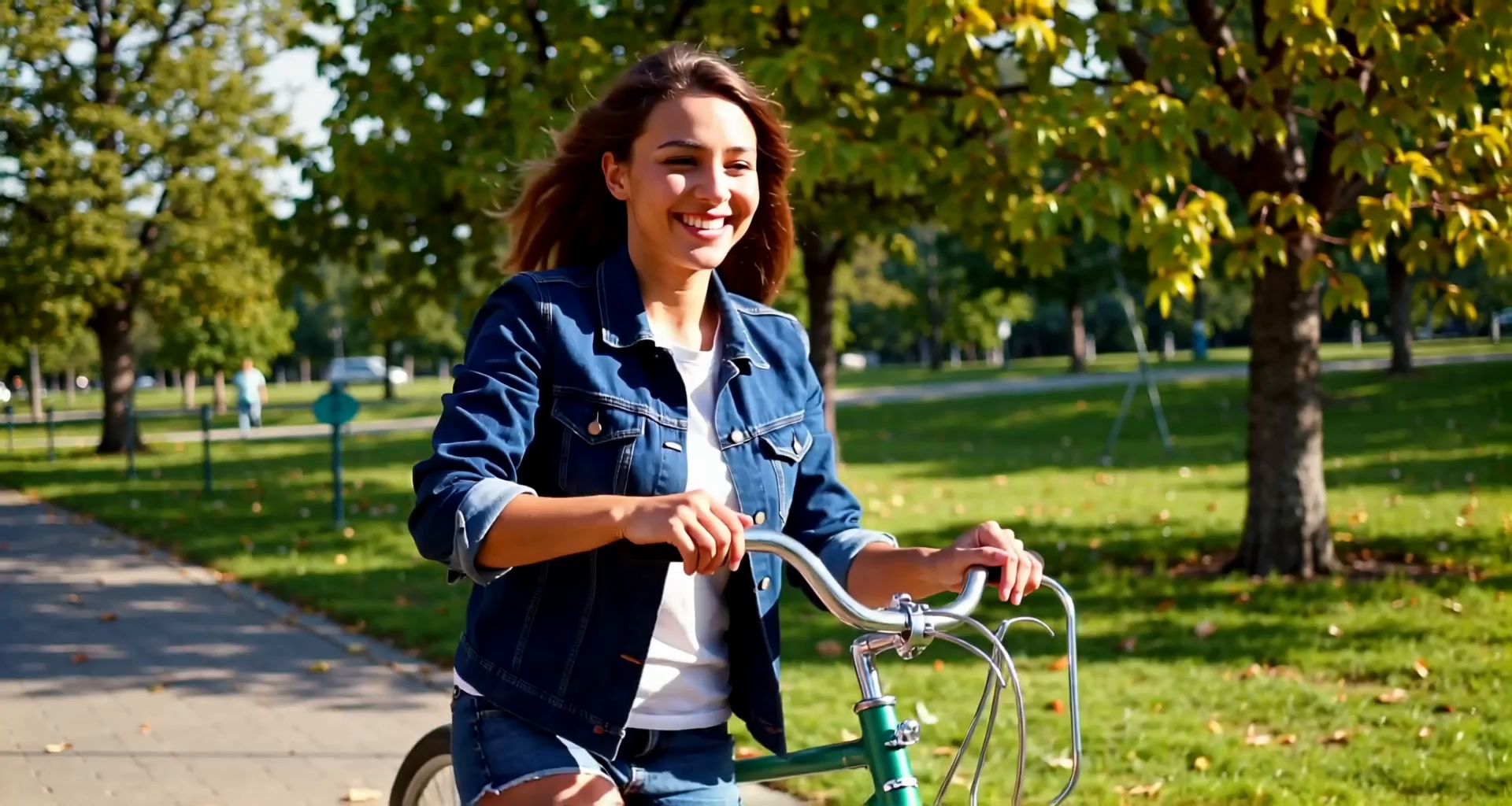} & \frameimg{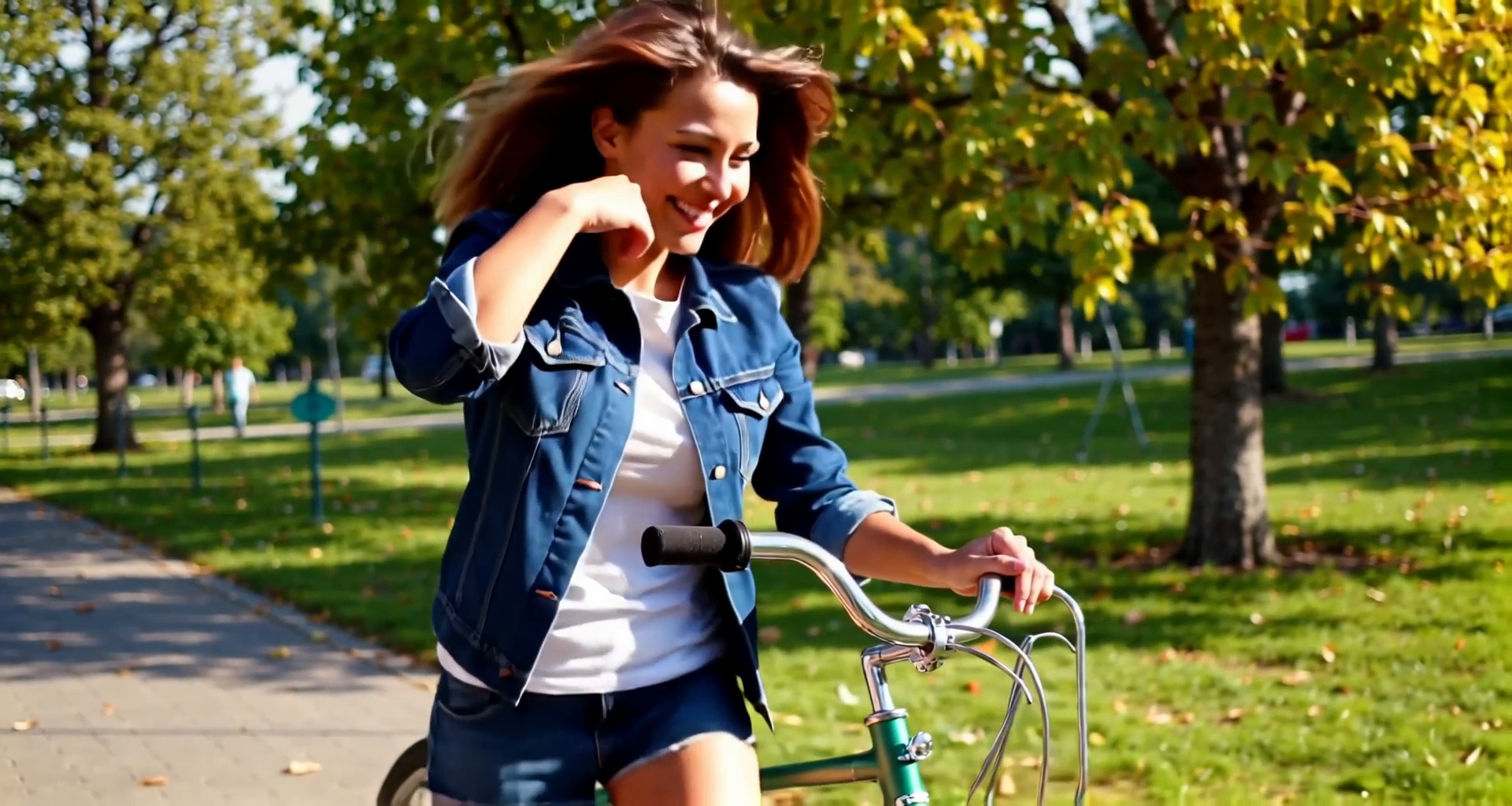} & \frameimg{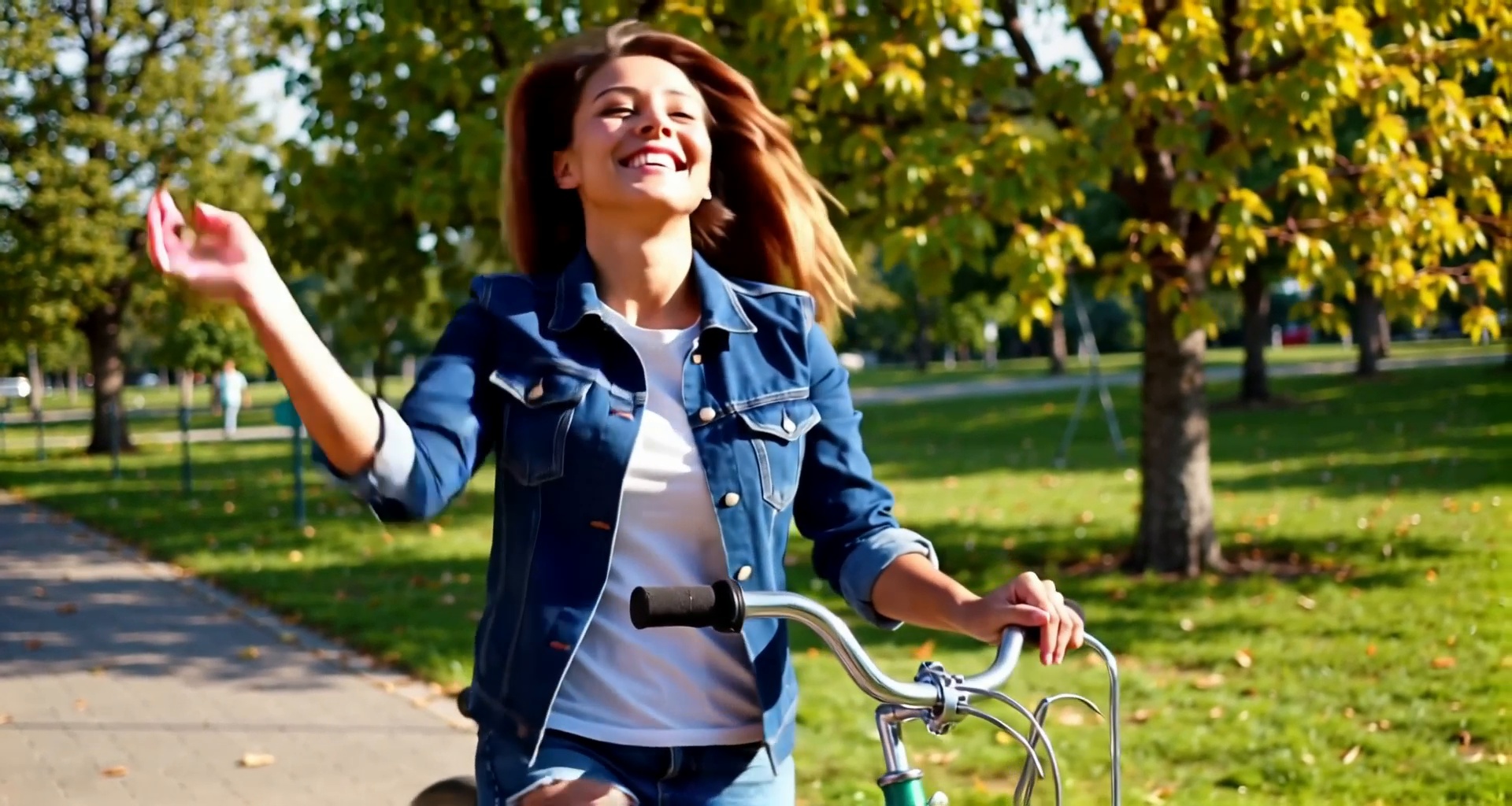} & \frameimg{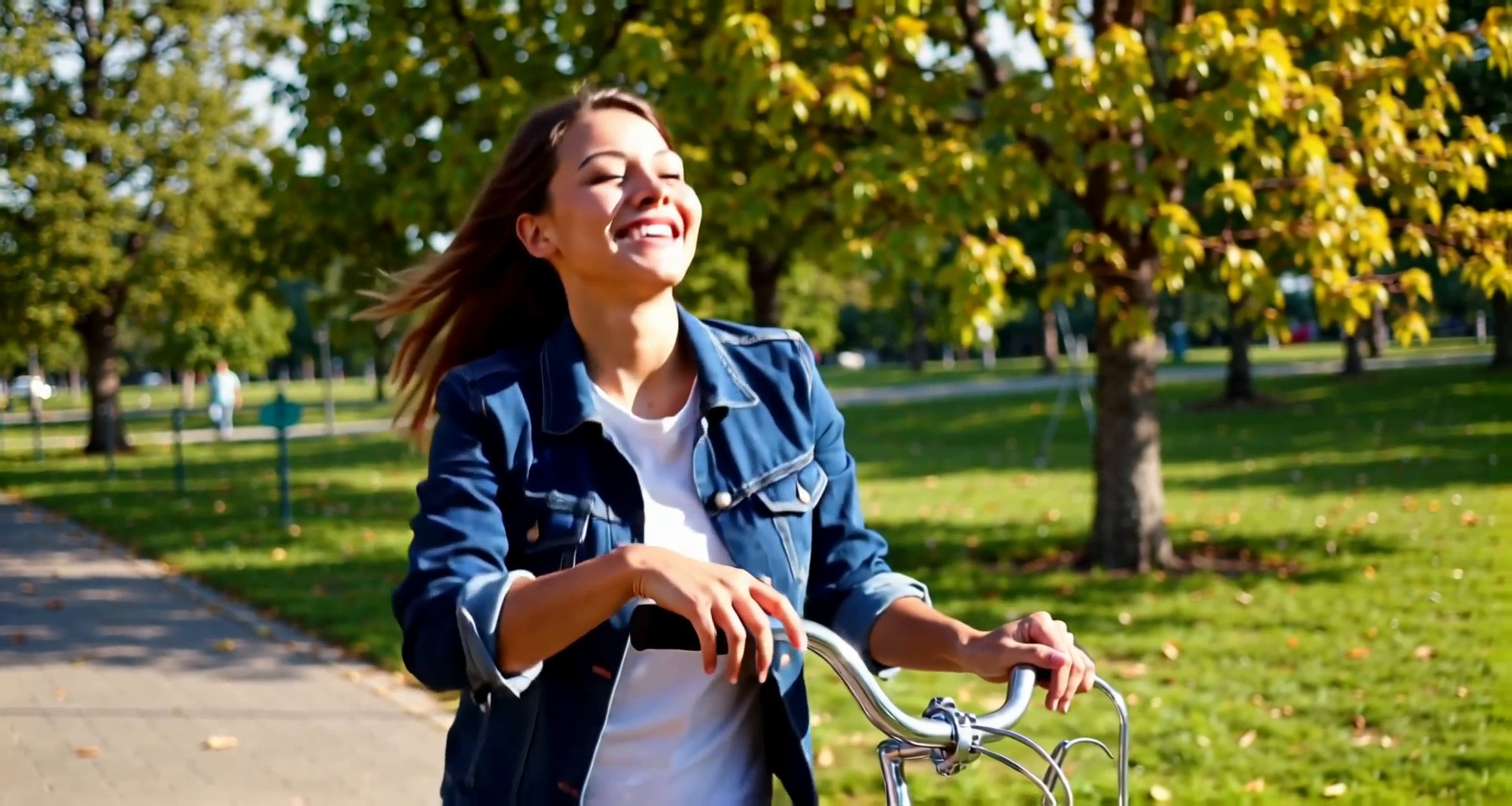} & 
\frameimg{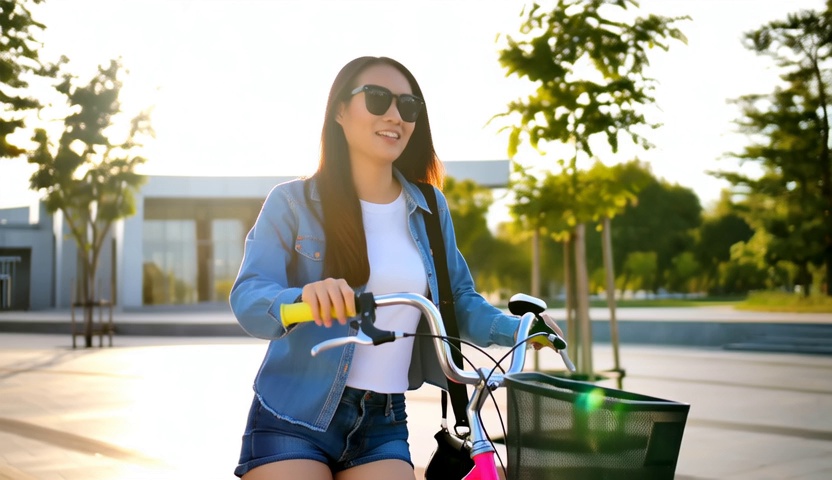} & \frameimg{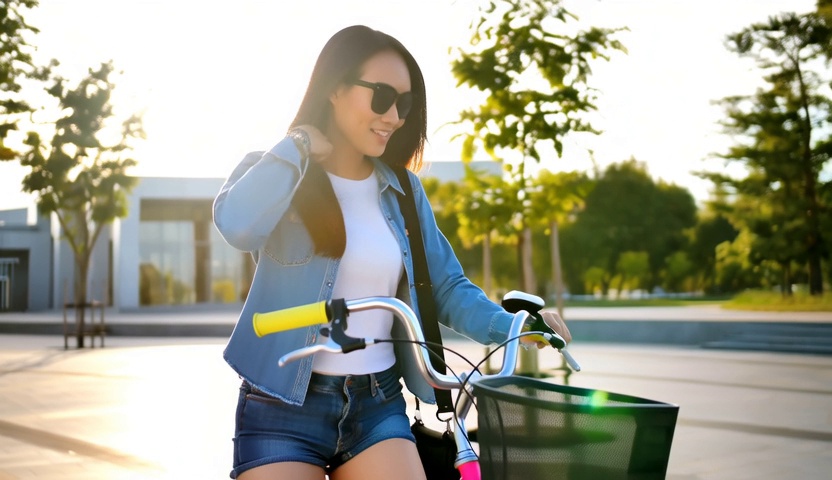} & \frameimg{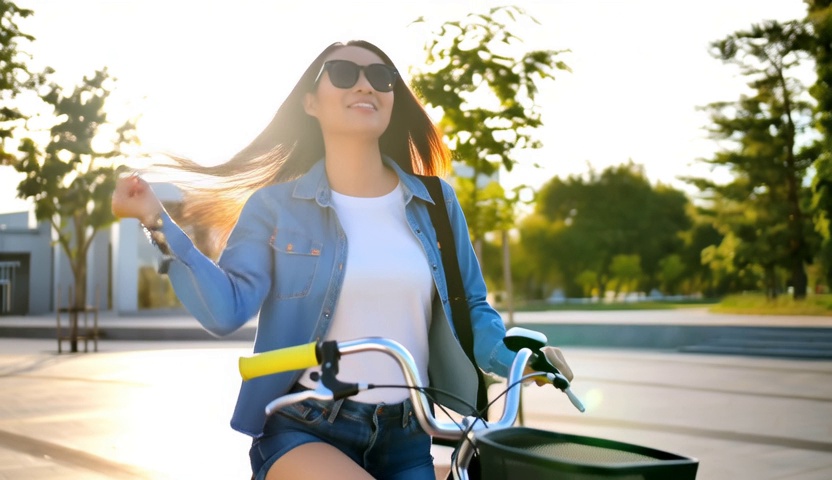} & \frameimg{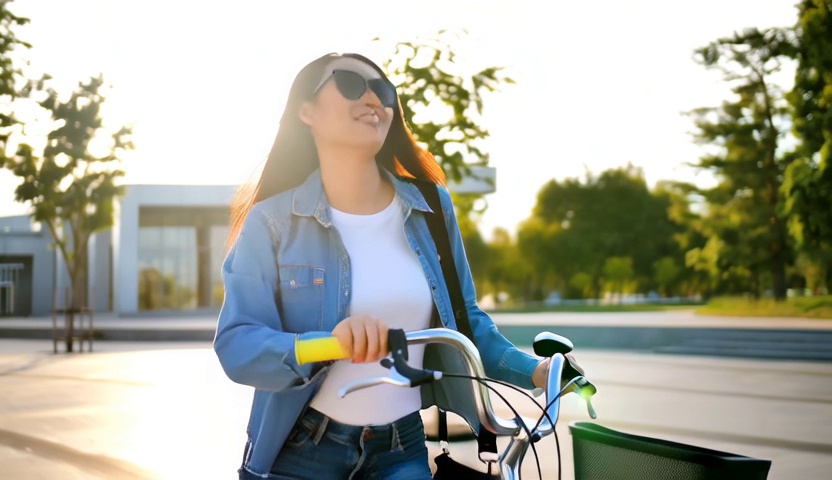} \\ 
\frameimg{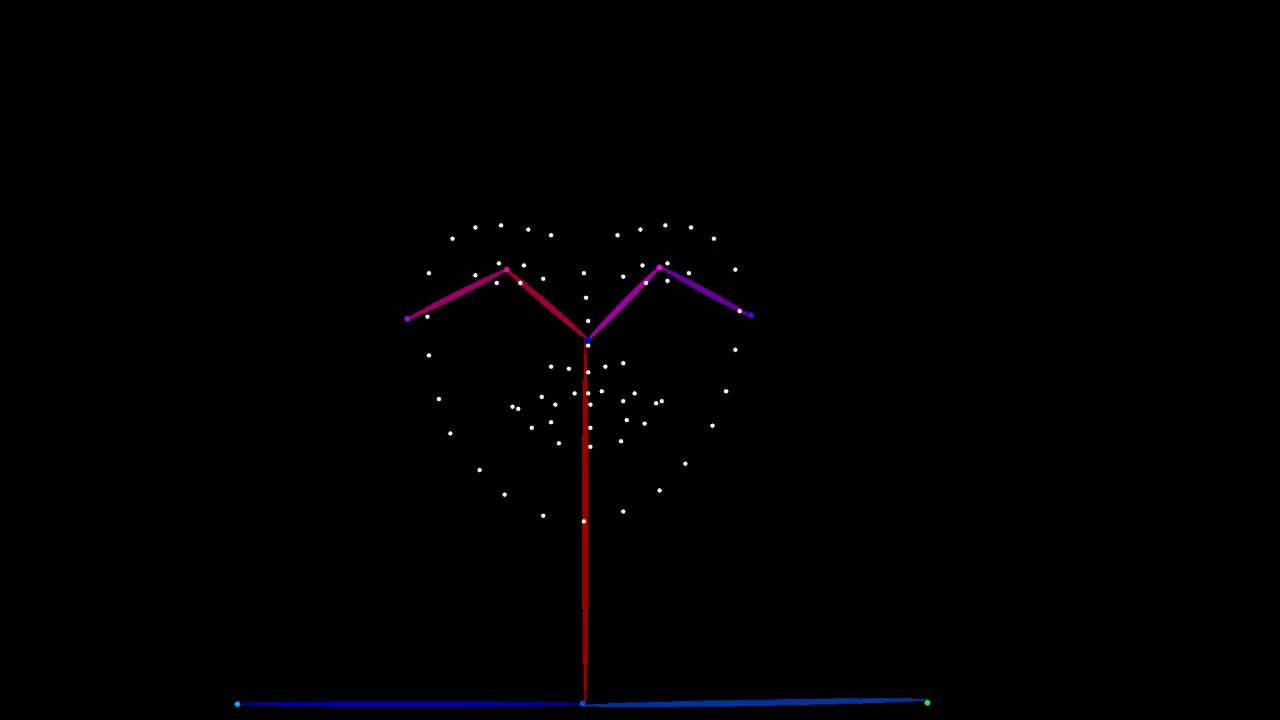} & \frameimg{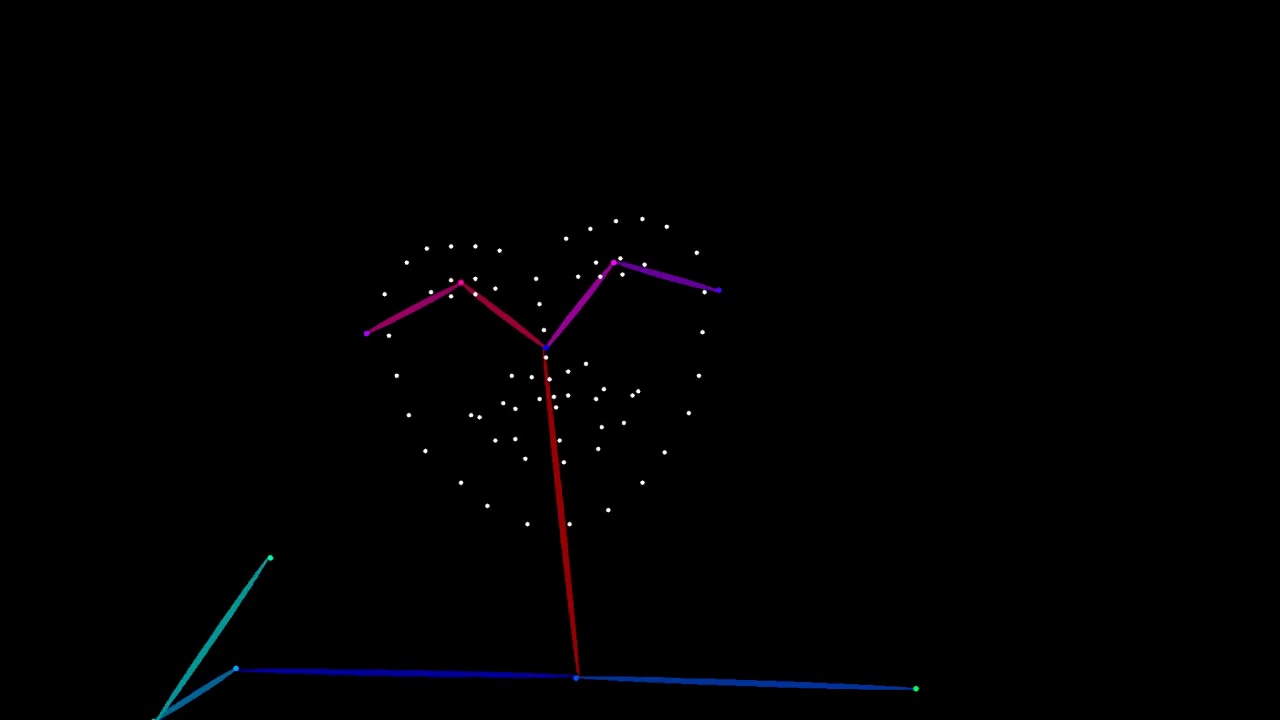} & \frameimg{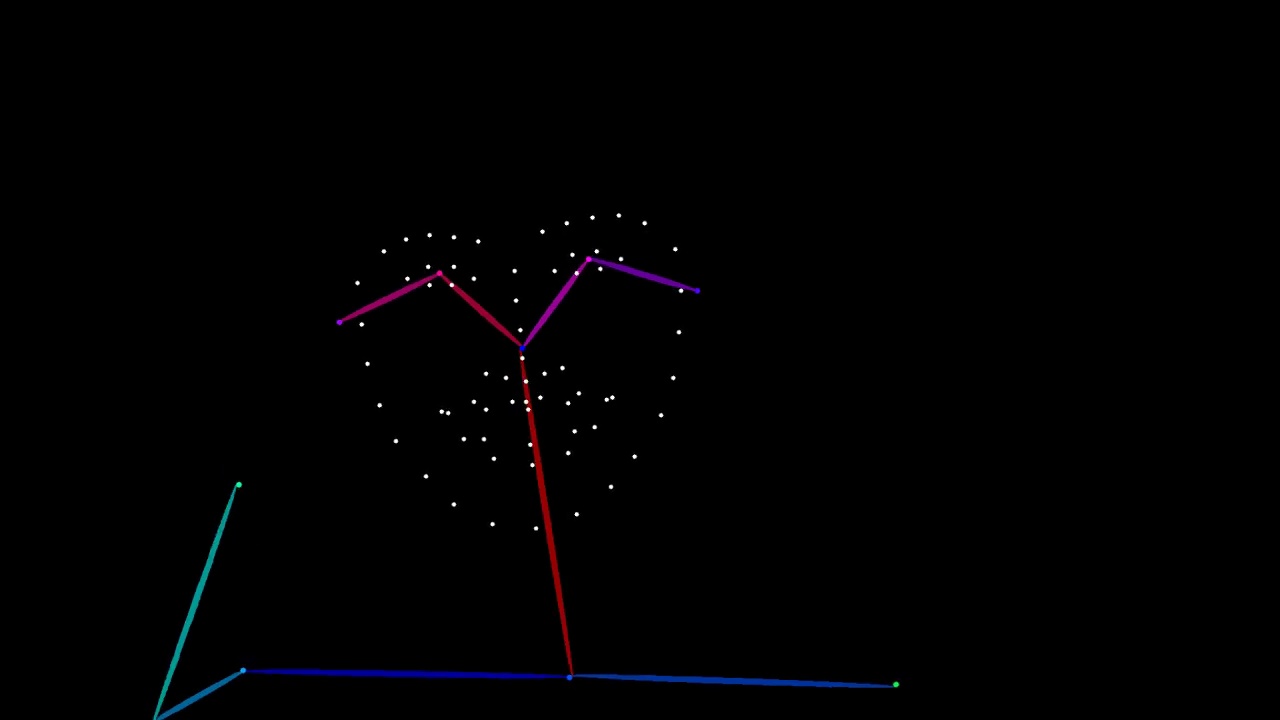} & \frameimg{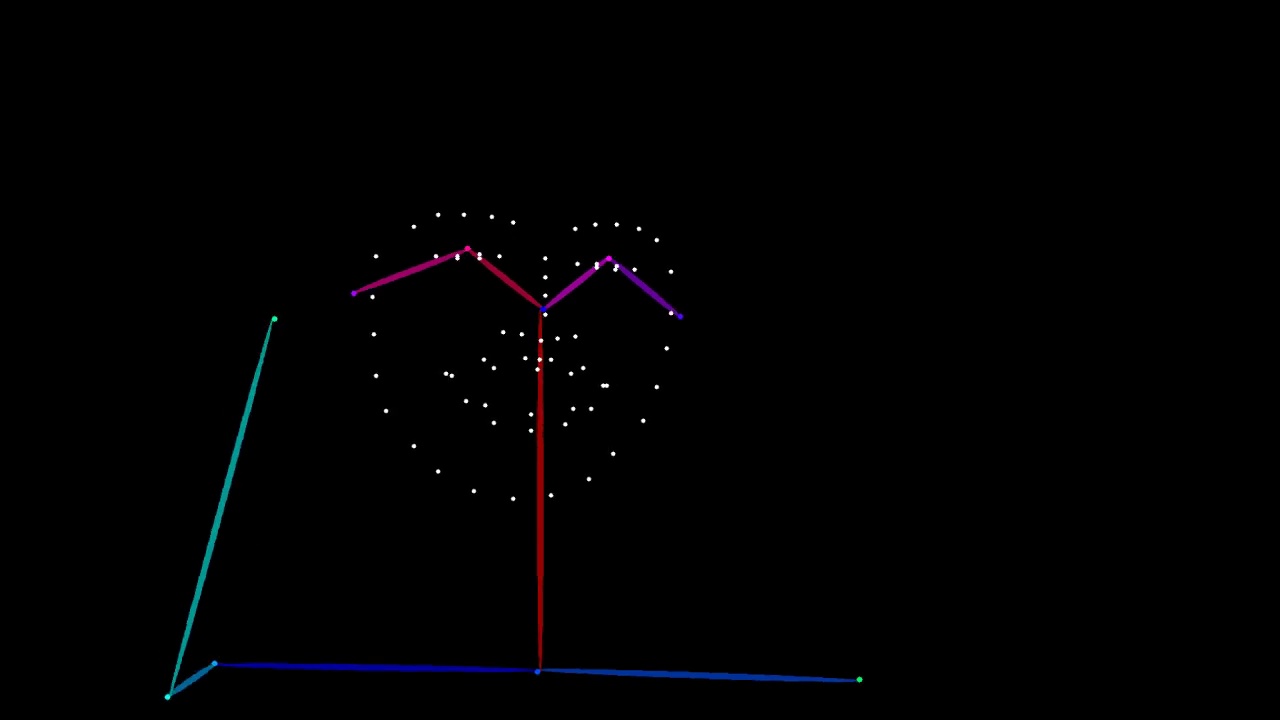} & 
\frameimg{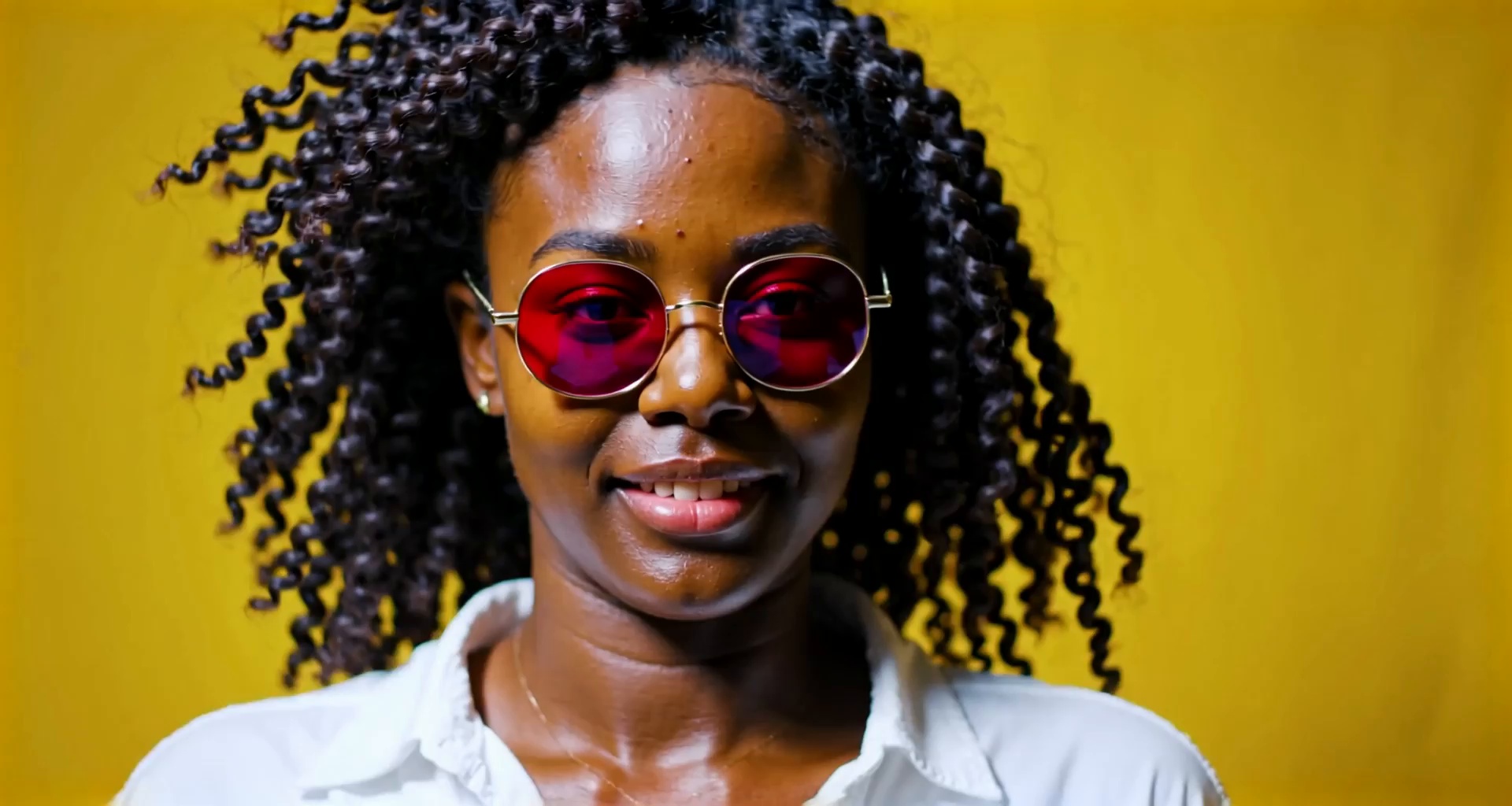} & \frameimg{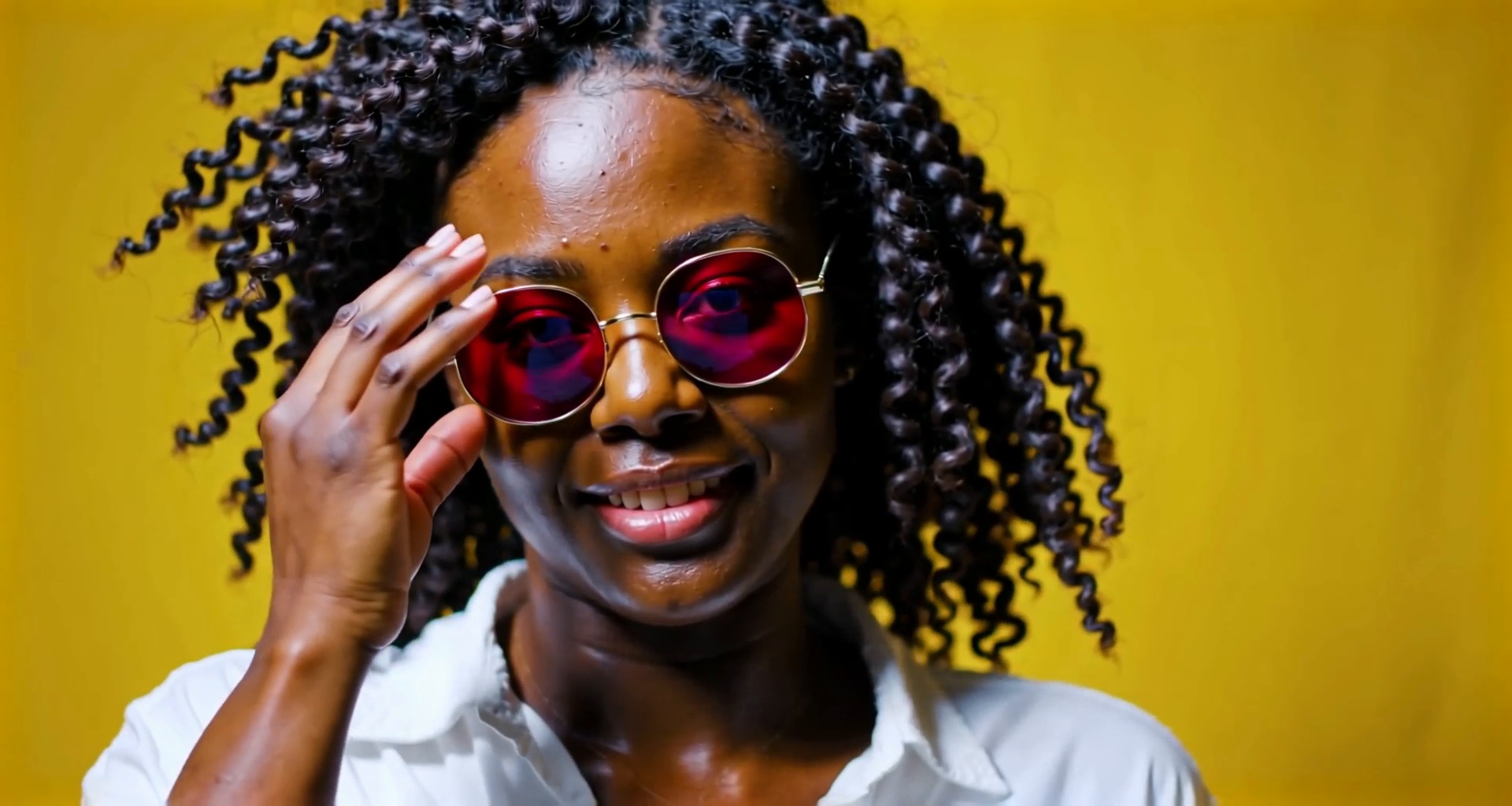} & \frameimg{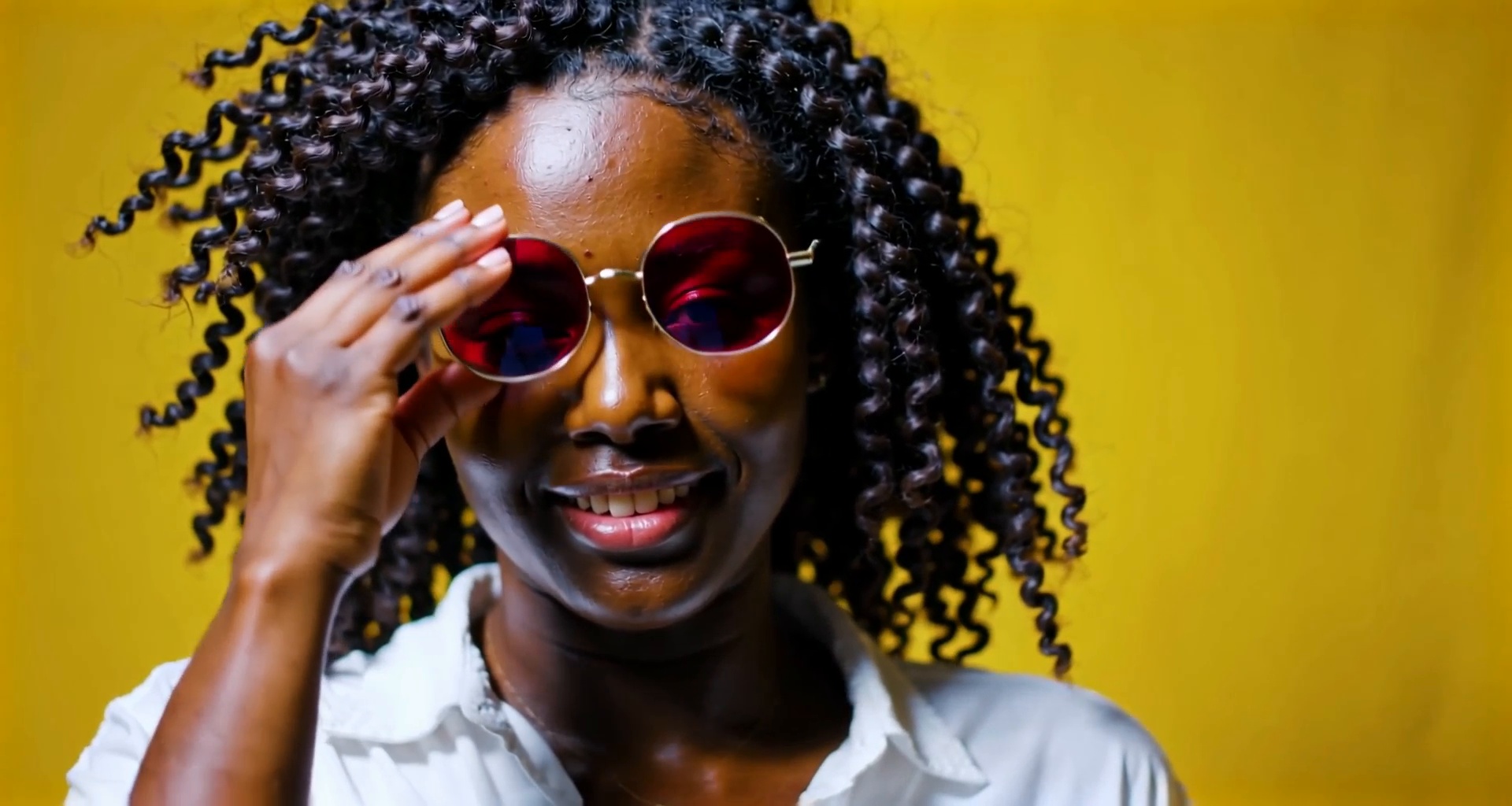} & \frameimg{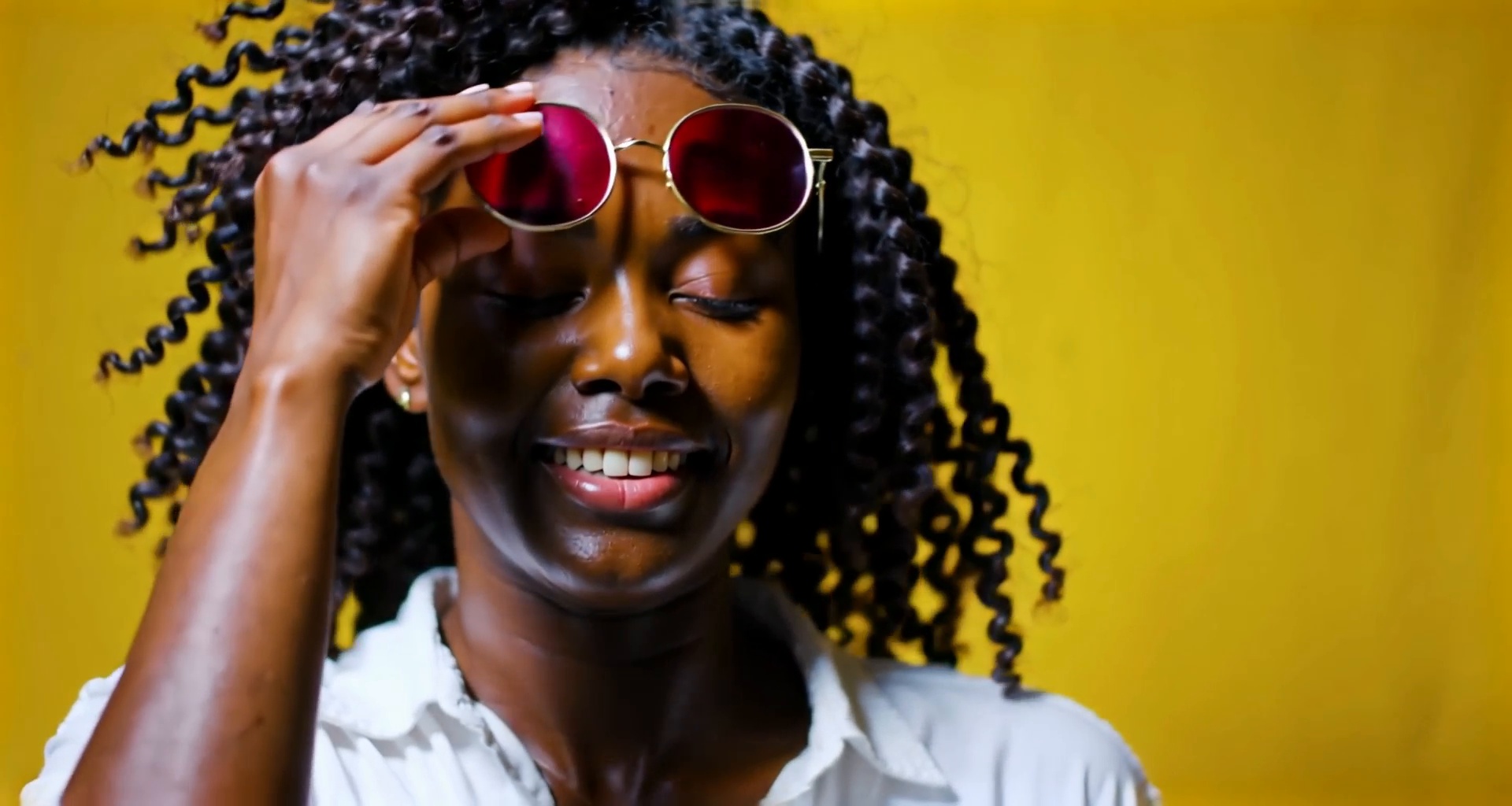} & 
\frameimg{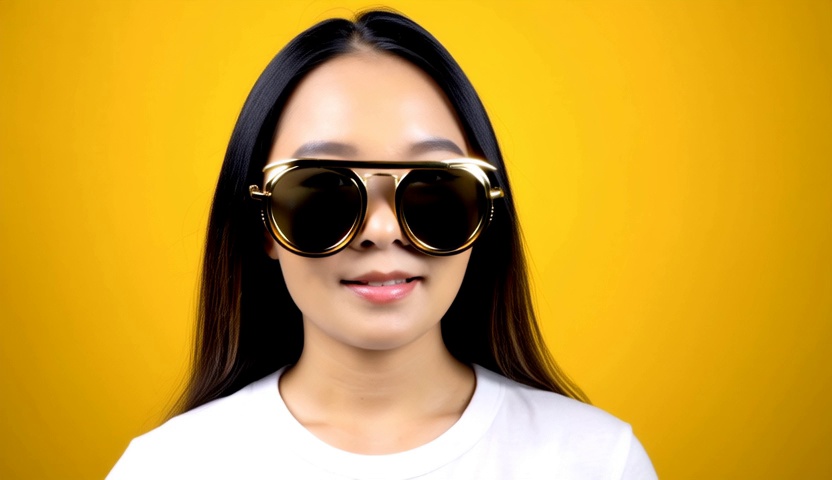} & \frameimg{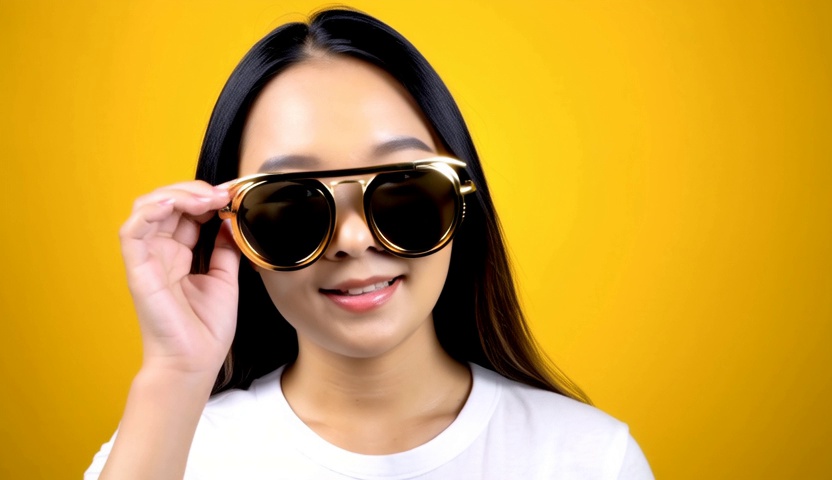} & \frameimg{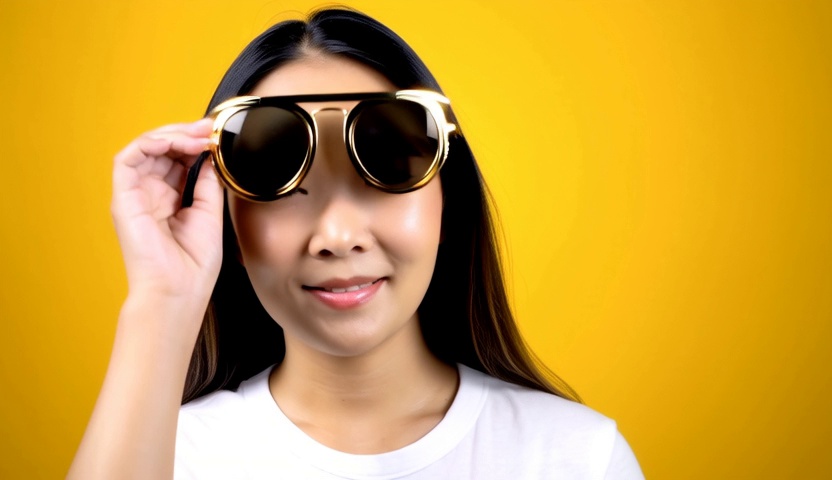} & \frameimg{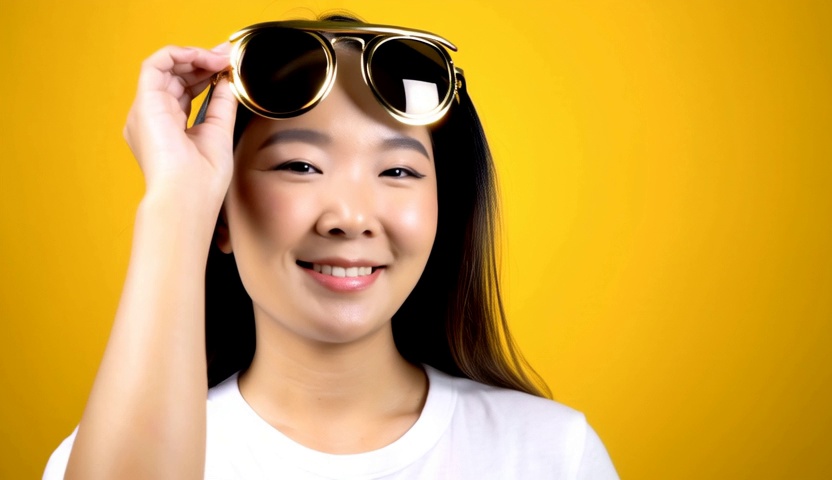} \\ 
\frameimg{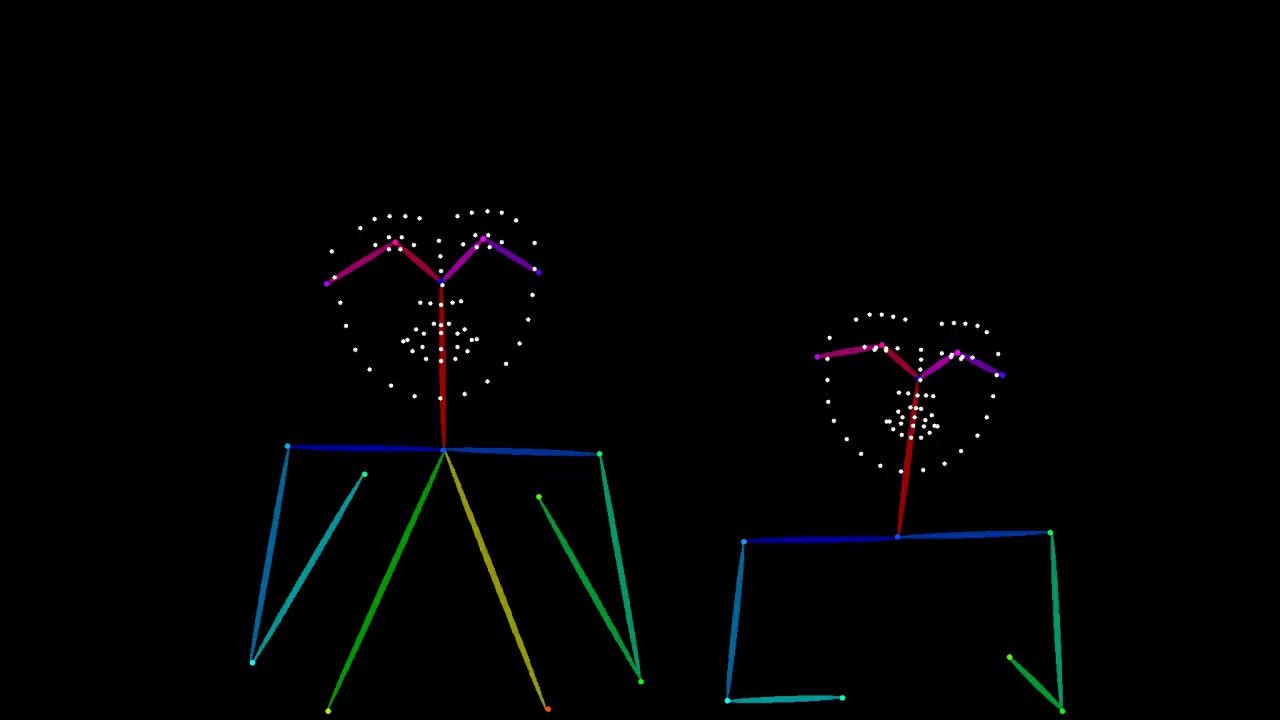} & \frameimg{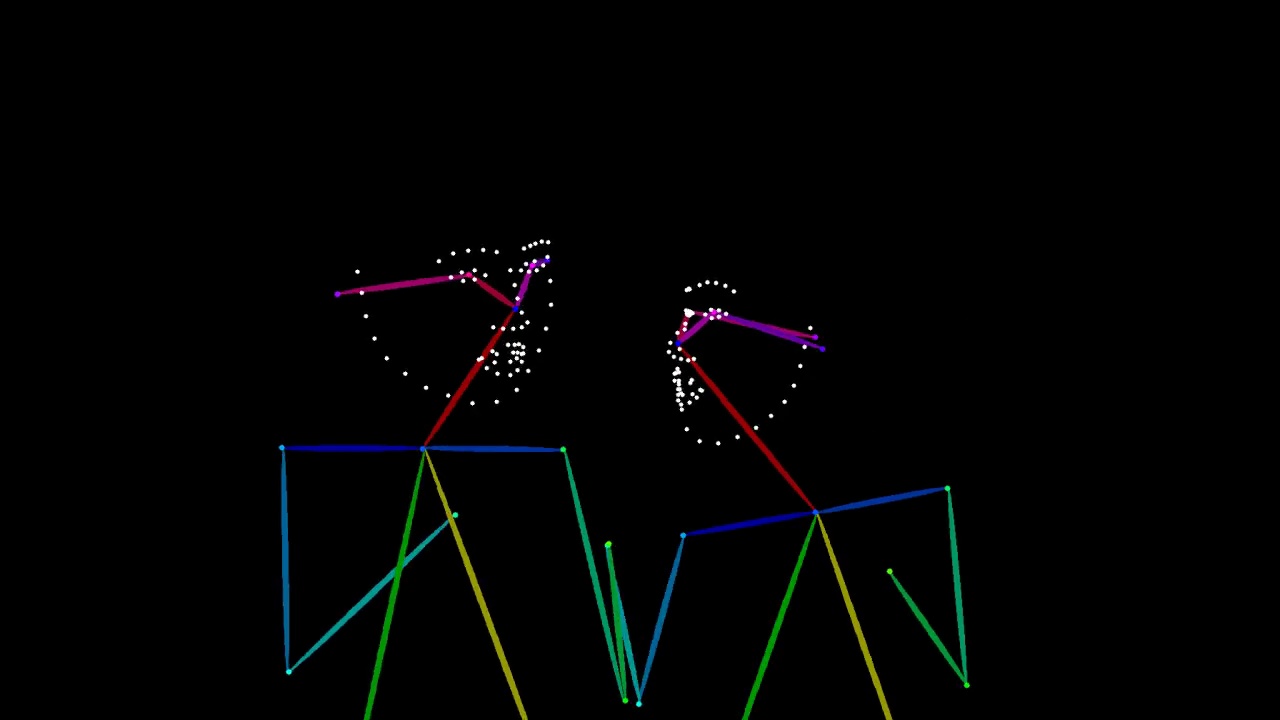} & \frameimg{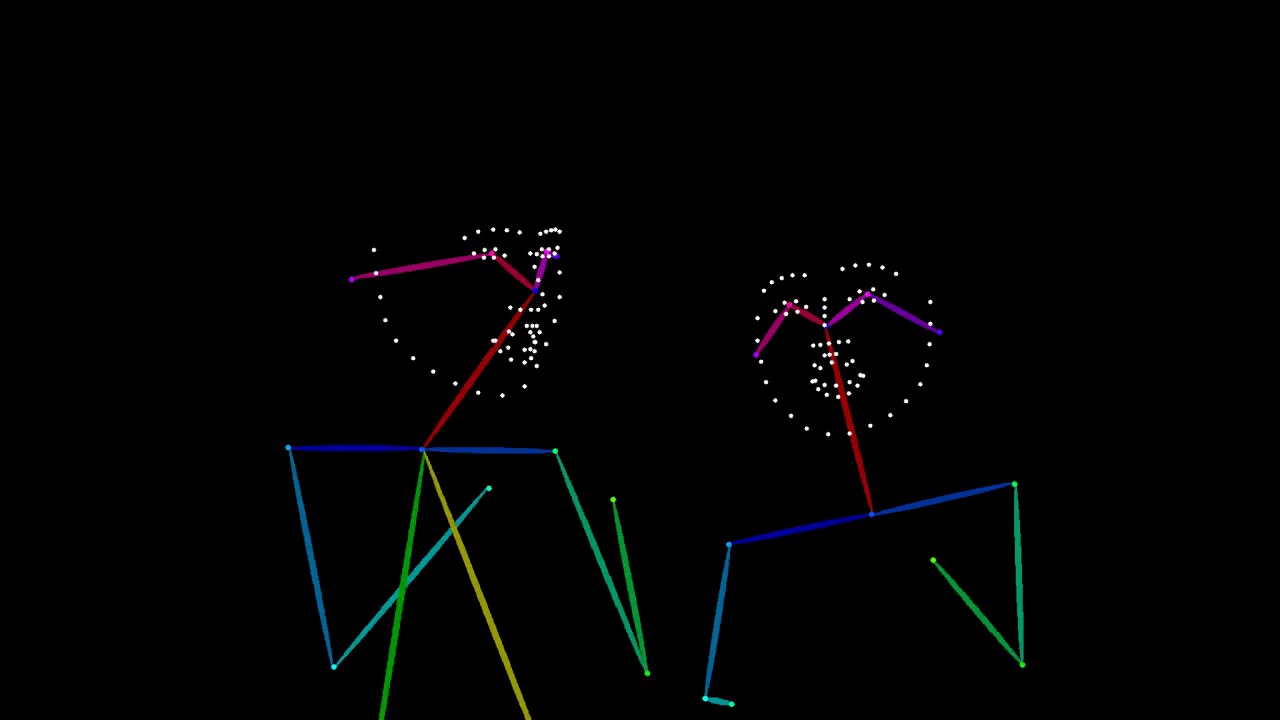} & \frameimg{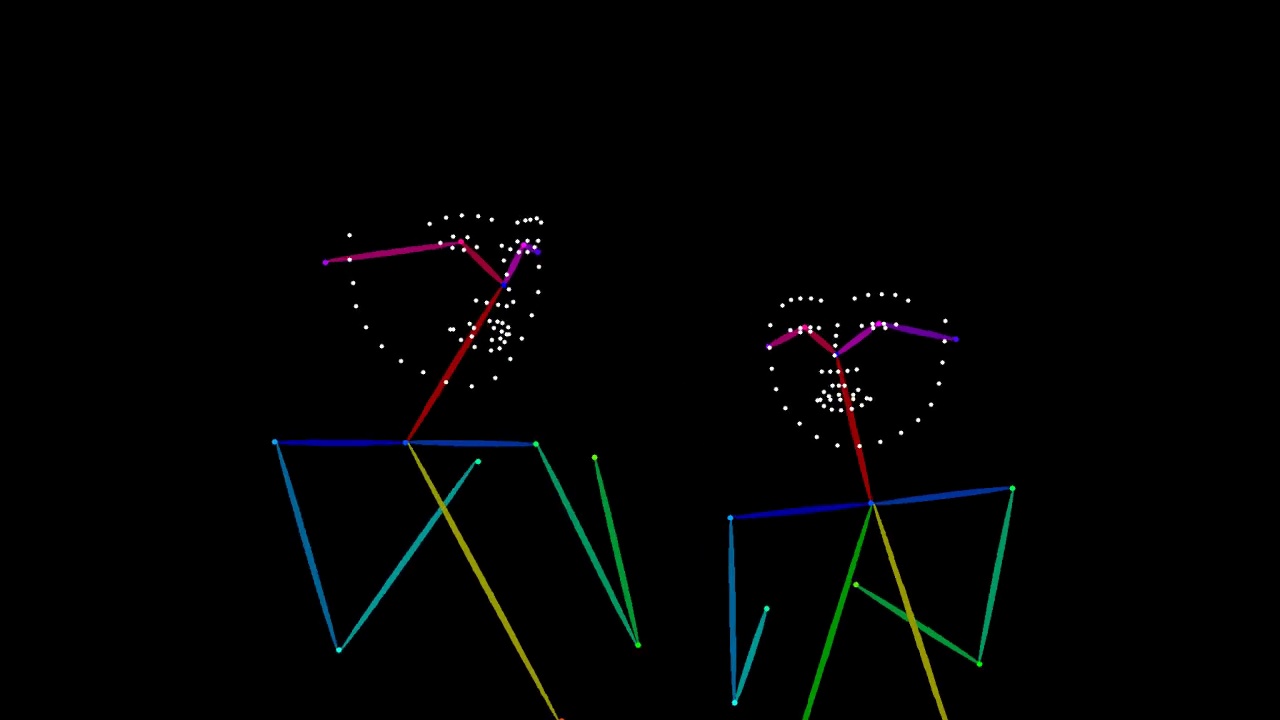} & 
\frameimg{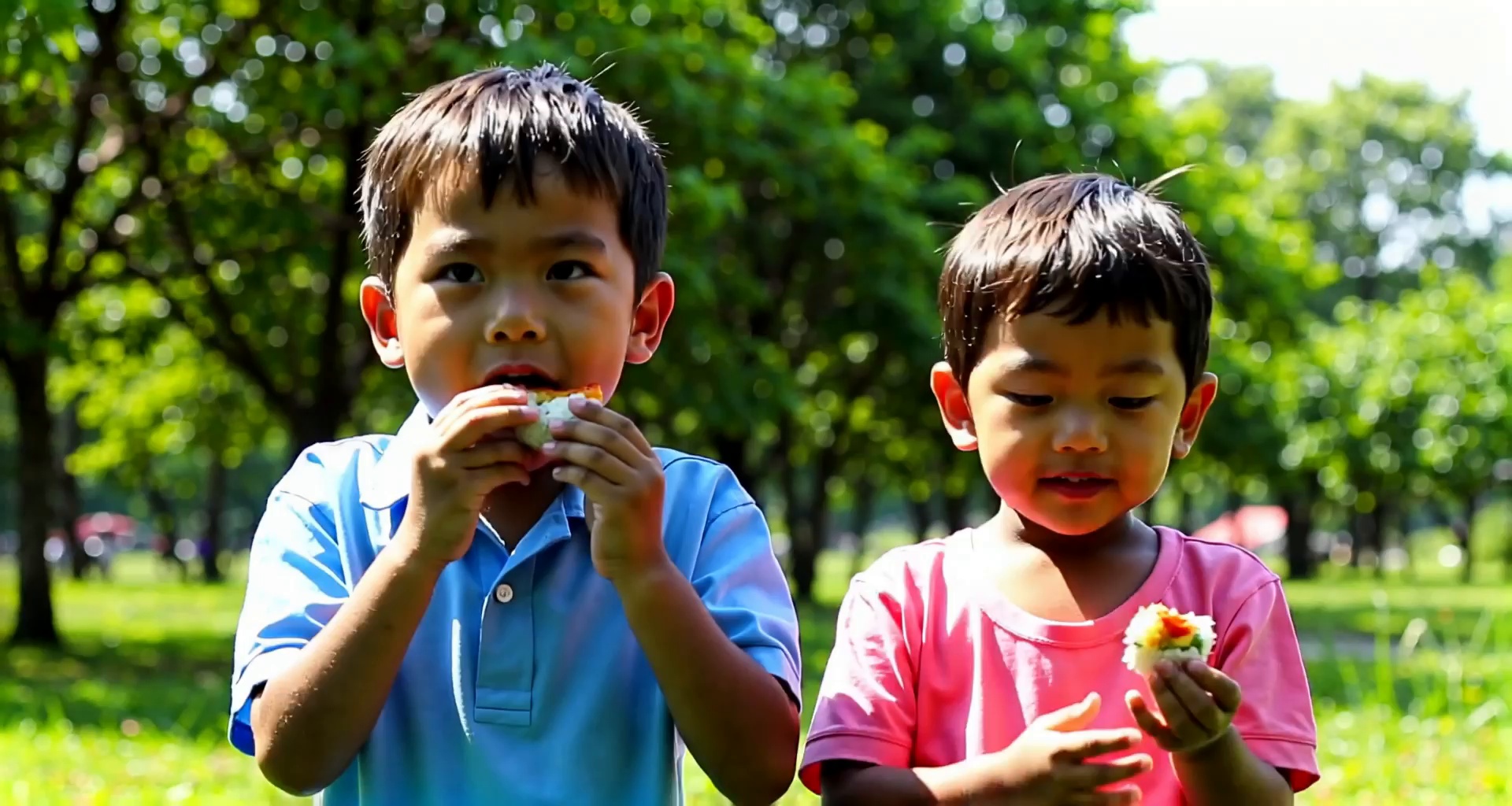} & \frameimg{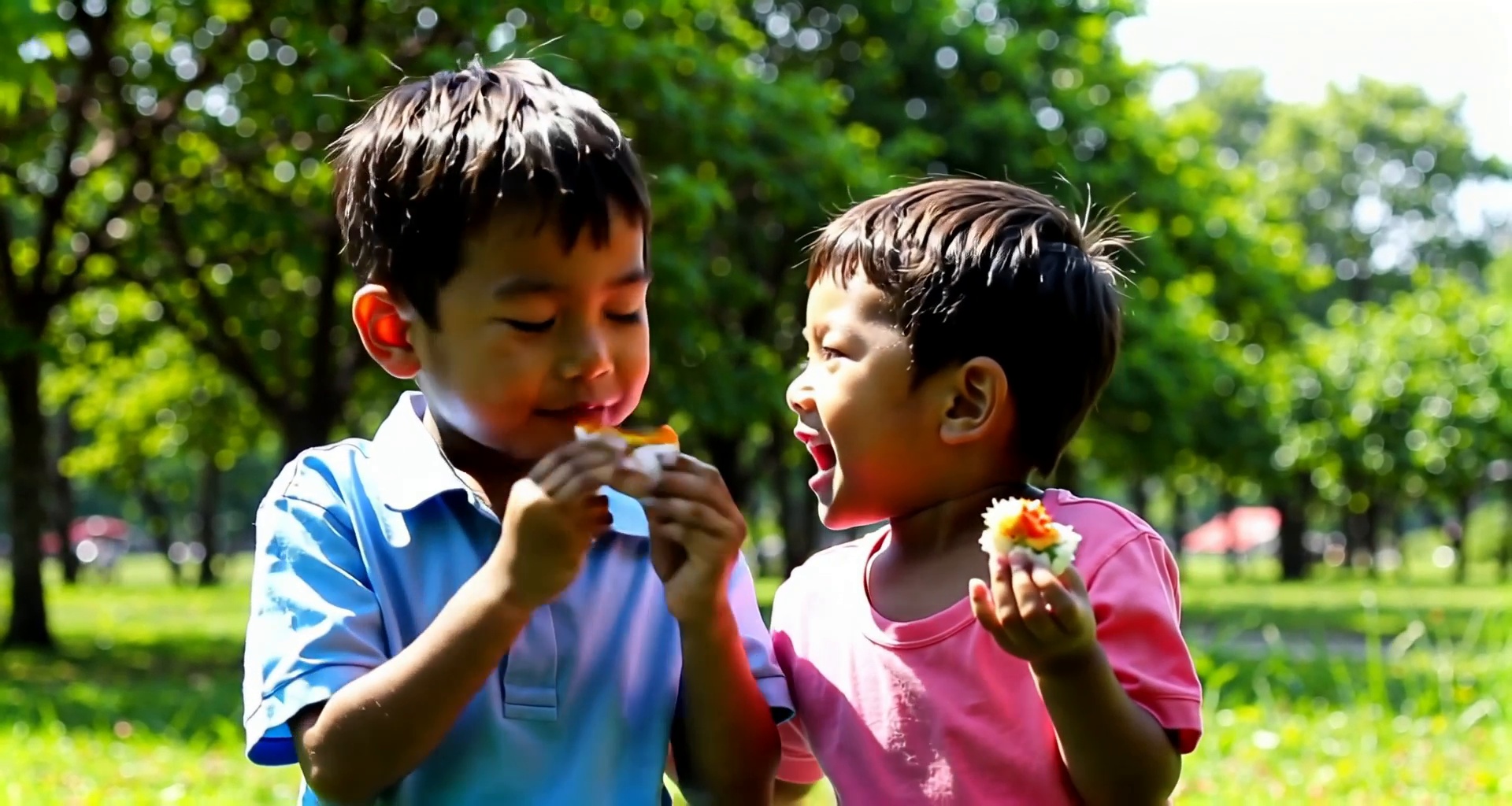} & \frameimg{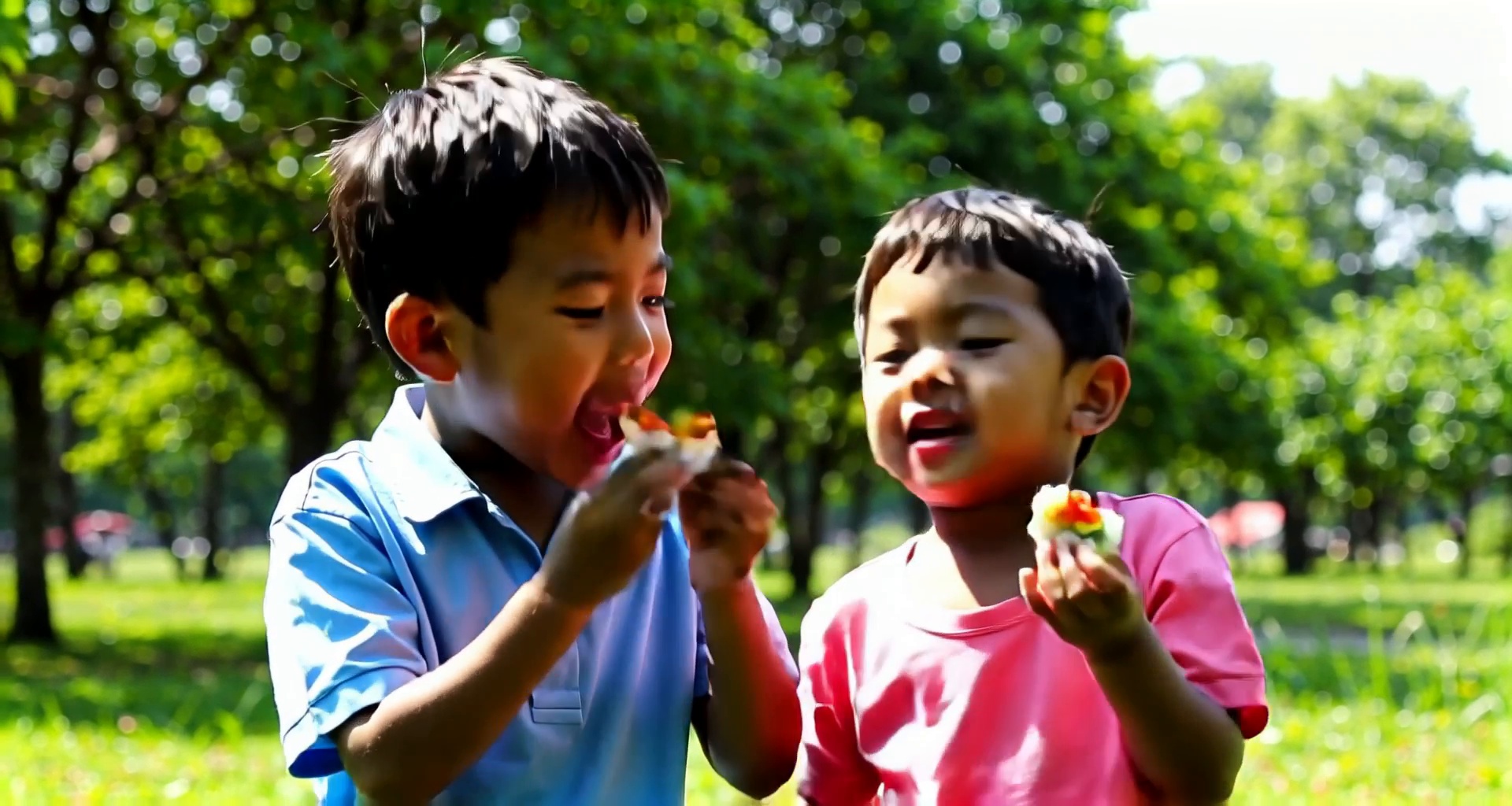} & \frameimg{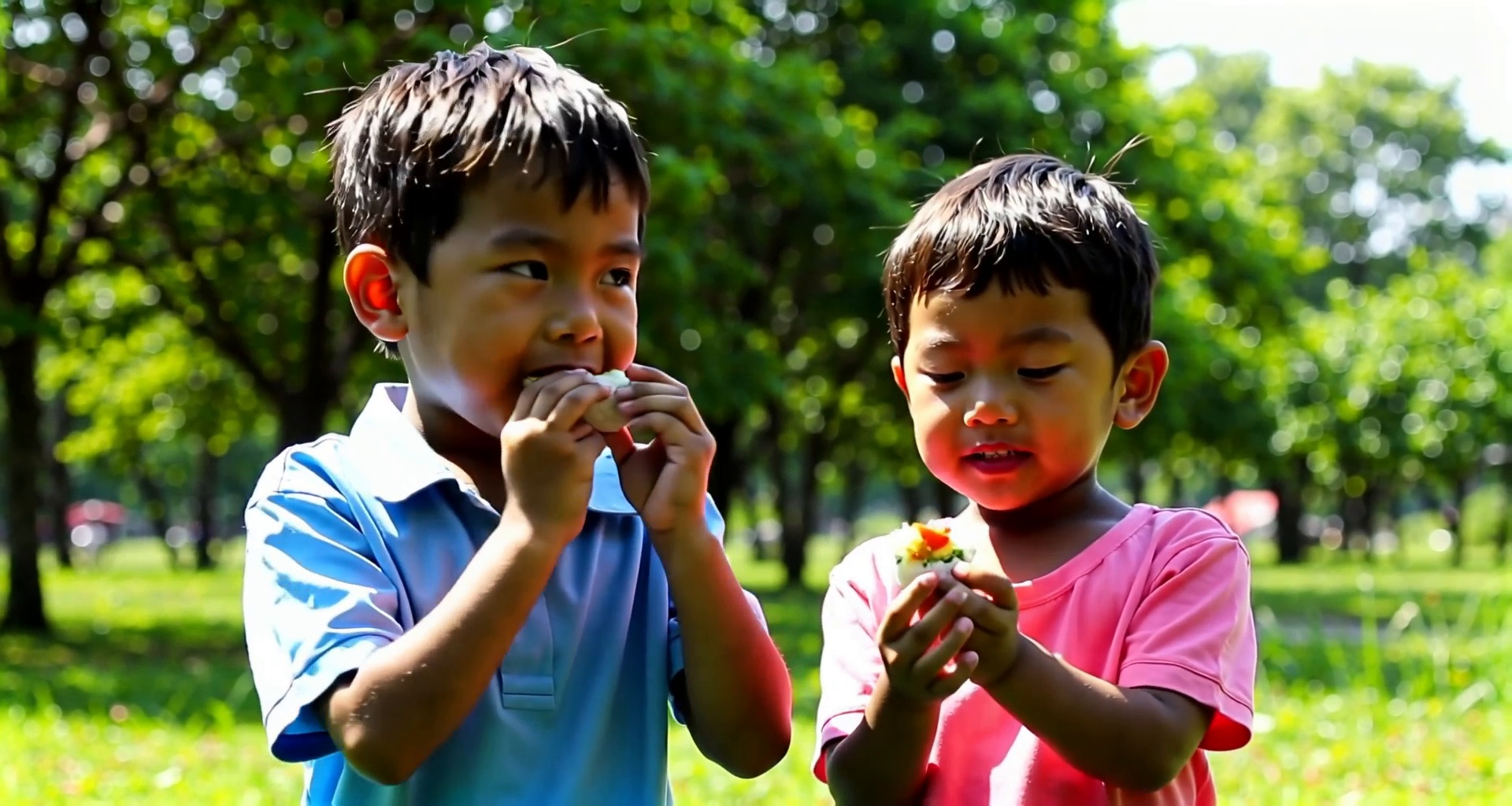} & 
\frameimg{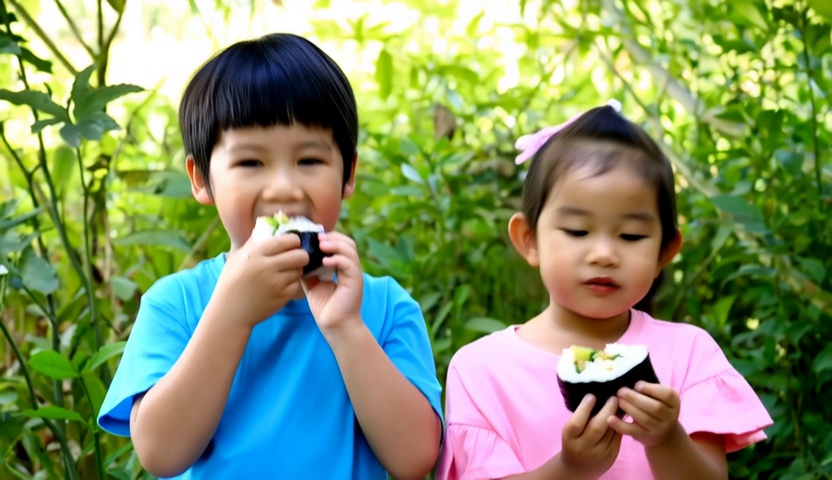} & \frameimg{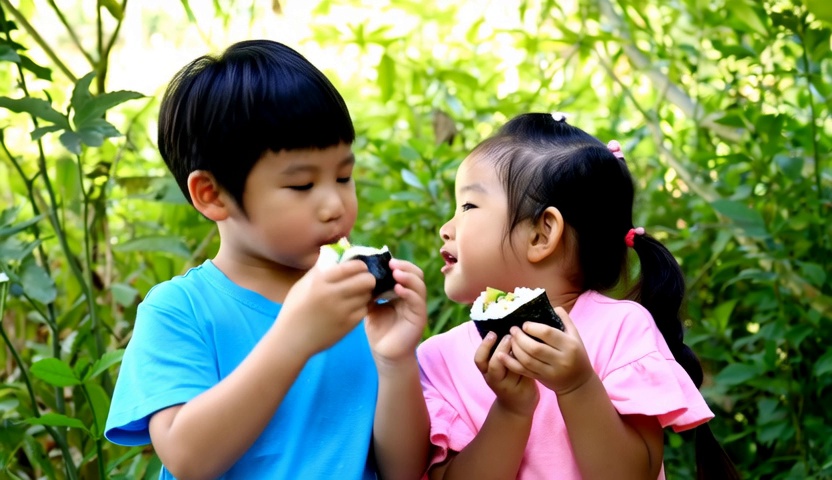} & \frameimg{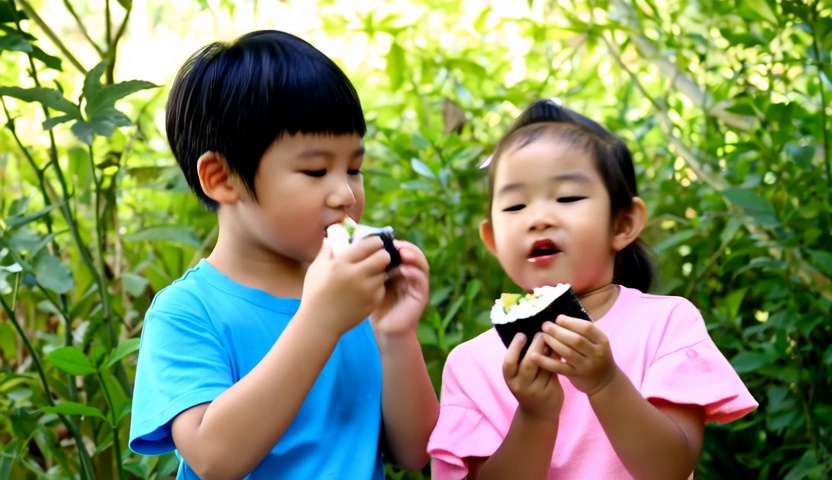} & \frameimg{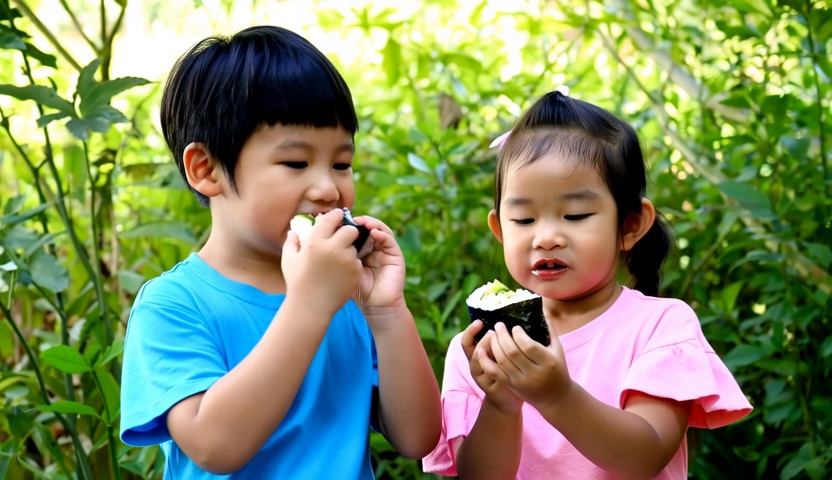} \\ 
\end{tabular}
\caption{Qualitative comparison on the VACE Benchmark (depth and pose). Each triplet: control input, \textcolor{blue}{ours}, \textcolor{red}{VACE}~\cite{jiang2025vace}. Our outputs show higher structural fidelity, consistent with Table~\ref{tab:vbench_comparison}.}
\label{fig:qualitative_comparison}
\end{figure*}

Figure~\ref{fig:qualitative_comparison} presents a qualitative comparison with VACE on the benchmark. Our outputs exhibit higher structural fidelity while maintaining natural motion and visual quality.

\subsection{Extended Modalities}
\label{sec:modalities}

\begin{figure*}[t]
\centering
\definecolor{oursblue}{HTML}{3B9DD9}%
\newlength{\mgfw}\setlength{\mgfw}{0.184\textwidth}%
\newcommand{\mgimg}[1]{\includegraphics[width=\mgfw]{figures/images/modality_gallery/#1}}%
\newcommand{\mgout}[1]{{\setlength{\fboxsep}{0pt}\setlength{\fboxrule}{0.5pt}%
  \fcolorbox{oursblue}{white}{\includegraphics[width=\dimexpr\mgfw-1pt\relax]{figures/images/modality_gallery/#1}}}}%
\newcommand{\modpair}[2]{%
  \begin{minipage}[c]{0.04\textwidth}%
    \centering\rotatebox[origin=c]{90}{\scriptsize\textbf{#1}}%
  \end{minipage}%
  \hfill
  \begin{minipage}[c]{0.95\textwidth}%
    \mgimg{#2_ctrl_1}\hfill\mgimg{#2_ctrl_2}\hfill\mgimg{#2_ctrl_3}\hfill\mgimg{#2_ctrl_4}\hfill\mgimg{#2_ctrl_5}\\[0.5pt]%
    \mgout{#2_out_1}\hfill\mgout{#2_out_2}\hfill\mgout{#2_out_3}\hfill\mgout{#2_out_4}\hfill\mgout{#2_out_5}%
  \end{minipage}%
}%
\modpair{Canny}{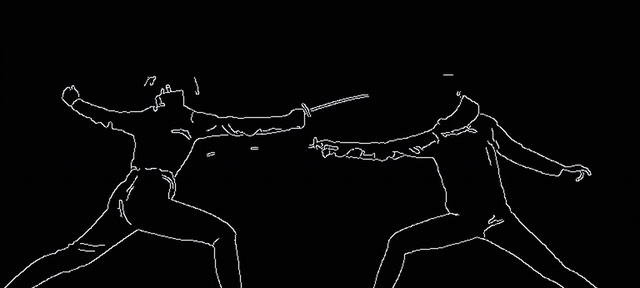}\\[3pt]
\modpair{Pose}{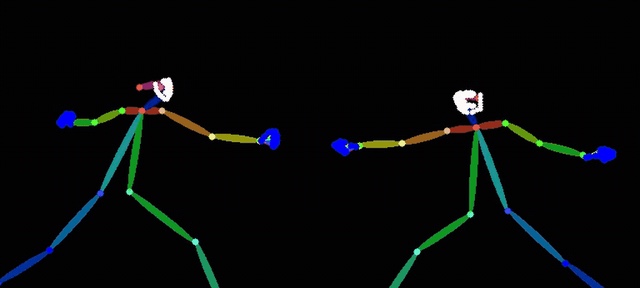}\\[3pt]
\modpair{Depth}{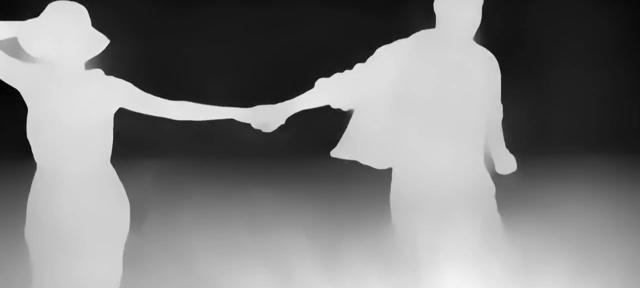}\\[3pt]
\modpair{Camera (Img)}{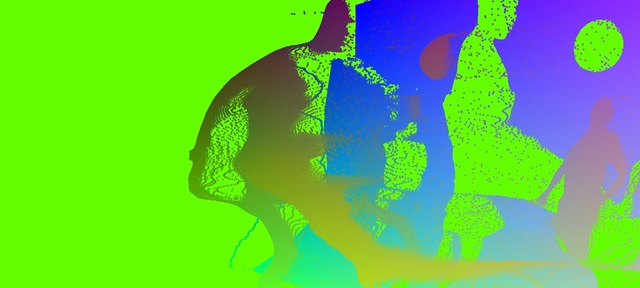}\\[3pt]
\modpair{Camera (Vid)}{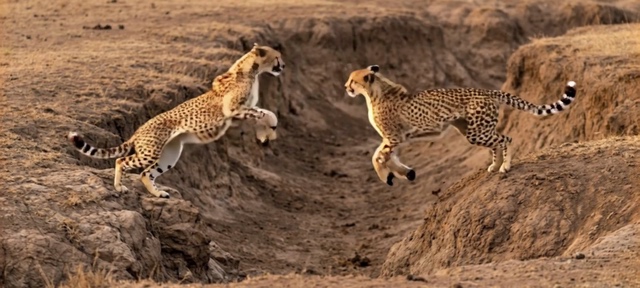}\\[3pt]
\modpair{Sparse Track.}{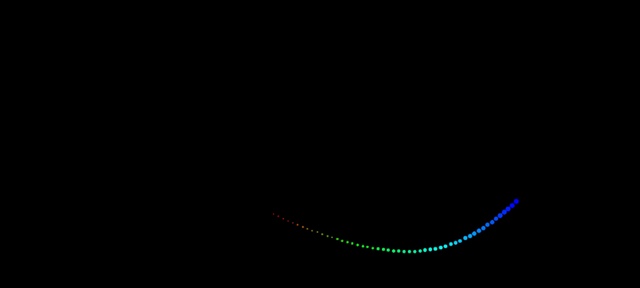}\\[3pt]
\modpair{Cut-on-Action}{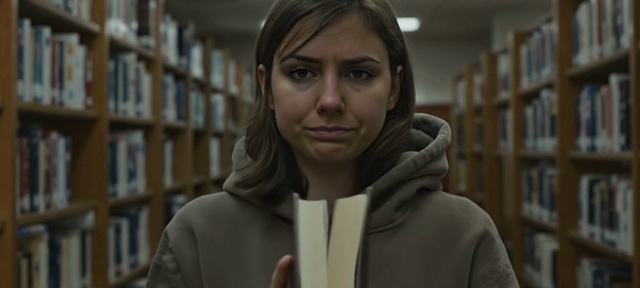}\\[3pt]
\modpair{Inpainting}{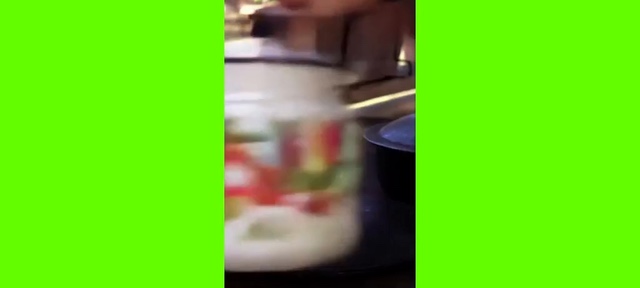}%
\caption{Partial gallery of control modalities. Each row pair shows control input (top) and generated output (bottom, \textcolor{oursblue}{blue border}) across five sampled frames. Each modality is an independent LoRA trained in 200--15{,}000 steps. We provide additional video examples in the supplementary.}
\label{fig:modality_gallery}
\end{figure*}

Beyond the benchmark controls, we demonstrate the breadth of modalities the framework supports (Figure~\ref{fig:modality_gallery}). Each modality uses its own LoRA, and adding a new one requires no retraining of existing ones. We support ControlNet-style modalities (depth, pose, canny), video editing (inpainting/outpainting, detailing), and composited controls (e.g., masked depth with pose for Blender rendering). We also train a \emph{sparse tracks} LoRA for point-trajectory-based motion control, similar to ATI~\cite{wang2025ati}, with tracks extracted via AllTracker~\cite{harley2025alltracker}. More results are demonstrated in the supplementary video.

\paragraph{Camera trajectory control.}
\label{sec:camera}
We support two modes of camera control: (1)~generating diverse camera motions from a single input image, and (2)~re-rendering an existing video at a new camera trajectory while preserving the original scene motion. For the latter, we estimate full camera parameters (extrinsics and intrinsics, including per-frame FOV) from the source video using SpatialTrackerV2~\cite{xiao2025spatialtracker}, then re-render each frame at the desired camera configuration. Optionally, rendering from a different timestamp retimes the output. Training uses a synthetic dataset of synchronized moving cameras, similar to ReCamMaster~\cite{bai2025recammaster}.
Unlike ReCamMaster~\cite{bai2025recammaster}, which controls only camera extrinsics, our camera LoRAs also control intrinsics, specifically field of view (FOV). This enables simulating focal length changes and effects such as the dolly zoom (``vertigo effect''), which is impossible with extrinsics-only methods.

\begin{table}[t]
\centering
\caption{Camera control on the ReCamMaster Benchmark (200 videos). Baselines from~\protect\cite{bai2025recammaster}. $^\dagger$COLMAP Structure-from-Motion (SfM) fails on 27\% of our videos; SpatialTrackerV2~\protect\cite{xiao2025spatialtracker} (all videos) yields 3.55$^\circ$.}
\label{tab:camera_comparison}
\setlength{\tabcolsep}{6pt}
\begin{tabular}{lcc}
\toprule
Method & CLIP-F (\%) $\uparrow$ & RotErr ($^\circ$) $\downarrow$ \\
\midrule
GCD~\cite{van2024gcd} & 95.66 & 2.27 \\
Traj-Attn~\cite{xiao2025trajattn} & 96.52 & 2.18 \\
DaS~\cite{cheng2025das} & 98.32 & 1.45 \\
ReCamMaster~\cite{bai2025recammaster} & 98.74 & \textbf{1.22} \\
\midrule
\textbf{Ours} & \textbf{99.13} & 6.00$^\dagger$ \\
\bottomrule
\end{tabular}
\end{table}

Table~\ref{tab:camera_comparison} compares our camera control against dedicated methods on the ReCamMaster Benchmark~\cite{bai2025recammaster}. We evaluate on 200~randomly sampled videos across 10~trajectory types. Our method achieves the highest CLIP-F score (99.13\%), surpassing ReCamMaster (98.74\%). Our COLMAP-based RotErr is 6.00$^\circ$, though SfM fails on 27\% of our videos, likely underestimating the true average error. SpatialTrackerV2 tracking on all 200~videos yields 3.55$^\circ$ (not directly comparable to baselines). While dedicated camera control architectures achieve lower rotation error, our camera LoRA is a lightweight adapter on a general-purpose audio-visual model and additionally controls camera intrinsics (FOV), which extrinsics-only methods cannot.

\paragraph{Audio-visual applications.}
\label{sec:av}
Audio-visual modalities are qualitatively different from the video-only controls above: training uses audio pairs, and the generated output spans both modalities (see supplementary video for qualitative results). We demonstrate two audio modalities: \emph{audio intensity control} and \emph{speech-to-ambient}, plus the cross-modal \emph{who-is-talking} control. Each audio LoRA is trained on audio-only pairs, yet at inference time generates both audio and video in a single joint pass: training on a single modality, deploying on both.

\emph{Audio intensity control} generates audio whose temporal dynamics follow visual content, addressing the same task as ReWaS~\cite{jeong2024rewas} and CAFA~\cite{benita2025cafa}. We train a single LoRA on the joint model rather than a dedicated video encoder and adapter on a unimodal backbone. The reference canvas carries the original video with its audio replaced by an energy-envelope sonification.

\begin{table}[t]
\centering
\caption{Audio intensity evaluation on VGGSound. Baselines are dedicated V2A methods; ours generates both audio and video jointly via a single LoRA. Baseline numbers from~\protect\cite{benita2025cafa}. \textbf{Bold}: best per metric.}
\label{tab:audio_intensity}
\setlength{\tabcolsep}{3.5pt}
\begin{tabular}{lccccc}
\toprule
Method & FAD$\downarrow$ & KL$\downarrow$ & IS$\uparrow$ & IB$\uparrow$ & Train Data \\
\midrule
ReWaS~\cite{jeong2024rewas}      & 14.71 & 2.69 & 8.45  & 0.15 & 160K \\
CAFA~\cite{benita2025cafa}       & 12.60 & 2.02 & 13.45 & 0.21 & 200K \\
MMAudio~\cite{cheng2025mmaudio}  & \textbf{5.32}  & \textbf{1.64} & 17.18 & \textbf{0.33} & 180K \\
\midrule
Ours & 57.25 & 7.74 & \textbf{34.51} & 0.13 & 7.8K \\
\bottomrule
\end{tabular}
\vspace{1mm}

\small
FAD: Fr\'{e}chet Audio Distance (PANNS), KL: KL divergence (PANNS), IS: Inception Score (PANNS), IB: ImageBind audio-visual similarity. ReWaS and CAFA numbers use the TV2A configuration from~\cite{benita2025cafa} Table~2; MMAudio uses the V2A configuration (video input only, no text).
\end{table}

Table~\ref{tab:audio_intensity} compares our intensity LoRA against dedicated video-to-audio methods on 254 VGGSound~\cite{chen2020vggsound} test samples. We report FAD, KL, IS (via PANNS~\cite{kong2020panns}) and audio-visual alignment (IB via ImageBind~\cite{girdhar2023imagebind}), following~\cite{benita2025cafa}. Our method achieves the highest IS (34.51), indicating diverse and class-distinctive audio generation. As expected, FAD and KL are higher because our model generates both audio and video jointly in a single pass rather than specializing in audio-only synthesis, and trains on 20--25$\times$ less data (7.8K vs.\ 160--200K samples). Notably, our IB score (0.13) is within 0.02 of ReWaS (0.15), a dedicated V2A model, despite our fundamentally different architecture: a single lightweight LoRA on a joint audio-visual backbone with no dedicated video encoder. Additionally, our audio quality (FAD) is capped by the quality of the LTX-2 vocoder.

\emph{Speech-to-ambient} embeds clean speech within ambient sounds matching a text-described scene. It operates in two modes: \emph{audio replacement} (AV$\to$A), where the video's audio is replaced while preserving the video, and \emph{joint generation} (A$\to$AV), where both audio and video are generated from the speech signal and a text prompt. The LoRA is trained on 2{,}600 paired samples in 5{,}000 steps.

\emph{Who-is-talking control} demonstrates an inherently cross-modal control: given a reference encoding spatial layout (bounding boxes) and temporal activity (which speaker is active when), the model generates multi-person talking video with synchronized lip motion and audio. The reference is abstract (colored rectangles on a black background, colored when speaking and gray when silent), yet the parallel canvas maps this to realistic video with joint audio.

To compare with MultiTalk~\cite{kong2025multitalk}, we start from the \emph{same raw inputs}: per-speaker identity references and separated audio tracks. We generate a first frame conditioned on the reference images, derive a bounding-box activity map via voice activity detection, and mix all tracks into a single audio signal. Our method supports an arbitrary number of speakers, unlike MultiTalk which is limited to two, and can generate audio jointly when none is provided.

\begin{table}[t]
\centering
\caption{Talking-head comparison on HDTF test set. Sync-C (confidence, $\uparrow$) and Sync-D (distance, $\downarrow$) measure lip-audio synchronization via SyncNet. E-FID measures expression quality. FID measures visual quality.
Baseline numbers are from~\cite{kong2025multitalk}.}
\label{tab:whoistalking}
\setlength{\tabcolsep}{5pt}
\begin{tabular}{lcccc}
\toprule
Method & Sync-C $\uparrow$ & Sync-D $\downarrow$ & E-FID $\downarrow$ & FID $\downarrow$ \\
\midrule
AniPortrait~\cite{wei2024aniportrait} & 3.09 & 10.94 & 1.32 & 32.83 \\
Hallo3~\cite{cui2025hallo3} & 6.55 & 8.49 & 1.12 & 33.98 \\
Sonic~\cite{ji2025sonic} & 8.35 & 6.43 & 1.22 & 29.53 \\
MultiTalk~\cite{kong2025multitalk} & \textbf{8.54} & \textbf{6.69} & 1.00 & 24.01 \\
\midrule
\textbf{Ours} & 4.50 & 10.31 & \textbf{0.18} & \textbf{12.31} \\
\bottomrule
\end{tabular}
\end{table}

Table~\ref{tab:whoistalking} compares on the HDTF benchmark. Our method achieves strong expression quality (E-FID~0.18) and visual fidelity (FID~12.31), outperforming all baselines on both. Lip-sync scores lag behind dedicated methods, as expected for a single general-purpose LoRA without purpose-built audio cross-attention.

\subsection{Ablations and Analysis}
\label{sec:efficiency}

Individual LoRAs range from 200~steps to 15{,}000~steps, with an aggregate budget of ${\sim}$55{,}000 steps across all trained modalities, less than one third of VACE's~\cite{jiang2025vace} 200{,}000 and comparable to a single specialized method like BulletTime~\cite{wang2025bullettime} (40K iterations). See Figure~\ref{fig:training_efficiency} in the supplementary for a visual comparison.

\paragraph{Training convergence.}
Training loss plateaus early for spatially-aligned controls. For depth, VBench scores reach 81.1 at 1{,}000 steps and 81.6 at 3{,}000 steps. The rapid convergence is consistent with LoRA fine-tuning existing weights rather than training new projections from scratch.

\paragraph{LoRA rank.}
LoRA ranks 32, 64, and 128 for depth yield VBench averages of 80.9, 81.3, and 81.6, a spread of less than 1~point. We use rank~128 as the default for a small quality margin.

\paragraph{Parallel canvas vs.\ spatial concatenation.}
A concatenation-based baseline (analogous to IC-LoRA~\cite{huang2024iclora}) trained with the same data, schedule, and rank fails to follow the depth signal (Figure~\ref{fig:canvas_vs_concat}), confirming the conditioning mechanism is the critical factor.

\paragraph{Generalization from synthetic data.}
Several LoRAs (camera-from-video, cut-on-action, local edit) are trained exclusively on synthetic or generated data yet generalize to real-world videos without fine-tuning. See the supplementary for details.

\section{Limitations}
\label{sec:limitations}
Despite these results, our framework has notable limitations.
It inherits the capabilities and limitations of the underlying base model; improvements in character motion, high-frequency details, and audio quality will directly translate to better results from our method as well.

Beyond base-model limitations, our design introduces specific constraints and failure modes.
\textbf{Mask representation.}
We encode masks within the reference video rather than providing them as a separate input. This works reliably in practice, with a rare failure mode when the video contains colors similar to the designated inpainting color.

\textbf{Complex character motion.}
When the reference depth or pose signal contains rapid, intricate character movements, the generated video may exhibit temporal jitter or implausible limb configurations.

\textbf{Camera control with fast scene dynamics.}
Camera trajectory control from video re-renders each frame as a point cloud from the target viewpoint. In scenes with rapid, non-rigid motion, the per-frame reprojection can produce stretching or ghosting artifacts in the reference canvas, and the LoRA may faithfully reproduce these artifacts rather than correcting them.

\textbf{Reference image conditioning.}
The most notable capability gap relative to VACE is reference image conditioning. Identity preservation is fundamentally different from the spatial and temporal controls our framework targets and is better handled by a separate, complementary mechanism.

\section{Conclusion}
We have presented AVControl, a modular framework for training controls for audio-visual foundation models. The parallel canvas, where per-token timestep disambiguates reference from generation tokens, enables faithful structural control where alternative conditioning approaches fail, without any positional encoding changes. The same mechanism allows fine-grained strength control at inference time. Combined with per-modality LoRA adapters on a frozen backbone, the framework outperforms all evaluated baselines on the VACE Benchmark while keeping the total training budget across all modalities to ${\sim}$55K steps. Adding a new control requires only a small dataset and a short training run, making the framework practical for the constantly growing space of useful controls.

The extendability of our framework is further validated by concurrent work that has already adopted it to train new control modalities: JUST-DUB-IT~\cite{chen2026justdubit} uses our framework for video dubbing via joint audio-visual diffusion, ID-LoRA~\cite{dahan2026idlora} extends it to identity-driven audio-video personalization, and In-Context Sync-LoRA~\cite{polaczek2025synclora} applies it to portrait video editing. These independent efforts demonstrate that the framework generalizes beyond our own set of trained modalities.

Several directions remain open: quantitative evaluation of audio-visual modalities via perceptual metrics and user studies, lightweight mechanisms for combining LoRAs at inference time, and user-specific controls trained on a handful of personal examples.

\clearpage

\bibliographystyle{splncs04}
\bibliography{egbib}

\clearpage
\section*{Supplementary Material}

This supplementary material provides comprehensive per-modality training details (Section~\ref{sec:supp_implementation}), the small-to-large control grid design (Section~\ref{sec:supp_control_grid}), extended ablations (Section~\ref{sec:supp_ablations}), and an extended qualitative gallery (Section~\ref{sec:supp_gallery}).
We strongly encourage viewing the \textbf{supplementary video}, which demonstrates all control modalities in motion, including audio-visual results that cannot be conveyed in static figures.

\section{Comprehensive Training Details}
\label{sec:supp_implementation}

Each per-modality LoRA is trained independently on a single NVIDIA H100 GPU with the AdamW optimizer and a linearly decaying learning rate schedule from $1 \times 10^{-4}$ to $1 \times 10^{-5}$. We select checkpoints by generating outputs on a held-out validation set and manually ranking checkpoints by overall preference across those samples. Table~\ref{tab:training_details} reports the exact configuration used for each modality.

\begin{table}[!ht]
\centering
\caption{Per-modality training details. \emph{Steps}: optimization steps at the selected checkpoint. \emph{Size}: number of training samples. \emph{Rank}: LoRA rank (alpha $=$ rank in all cases). \emph{Modules}: which transformer blocks receive LoRA adapters (SA = self-attention, CA = cross-attention, FF = feed-forward; prefixed by V = video, A = audio). Video-only LoRAs skip the audio loss; audio-only LoRAs skip the video loss.}
\label{tab:training_details}
\setlength{\tabcolsep}{2.5pt}
\footnotesize
\begin{tabular}{@{}llcccc@{}}
\toprule
Cat. & Modality & Steps & Size & Rank & Modules \\
\midrule
\multirow{3}{*}{Spatial}
& Depth          & 3K  & 2K   & 128 & V: SA \\
& Pose           & 3K  & 2K   & 128 & V: SA \\
& Canny / edges  & 3K  & 2K   &  32 & V: SA \\
\midrule
\multirow{3}{*}{Editing}
& Inpaint / outpaint     & 1K  & 1K  & 128 & V: SA \\
& Local edit             & 1K  & 1K  & 128 & V: SA, FF \\
& Video detailing        & 200 & 200 & 128 & V: SA, CA, FF \\
\midrule
\multirow{3}{*}{Camera}
& Camera (image)  & 3K  & 3K  &  32 & V: SA, CA, FF \\
& Camera (video)  & 10K & 10K & 128 & V: SA \\
& Cut-on-action   & 15K & 15K & 128 & V: SA, FF \\
\midrule
Motion & Sparse tracks & 5K & 5K & 32 & V: SA, CA, FF \\
\midrule
\multirow{2}{*}{Audio}
& Audio intensity    & 2K   & 7.8K & 128 & A: SA, FF, V\!$\to$\!A CA \\
& Speech-to-ambient  & 5K   & 2.6K & 128 & A: SA, FF, V\!$\to$\!A CA \\
\midrule
AV & Who-is-talking & 3.5K & 500 & 128 & All \\
\bottomrule
\end{tabular}
\end{table}

\begin{figure}[h]
\centering
\includegraphics[width=0.85\linewidth]{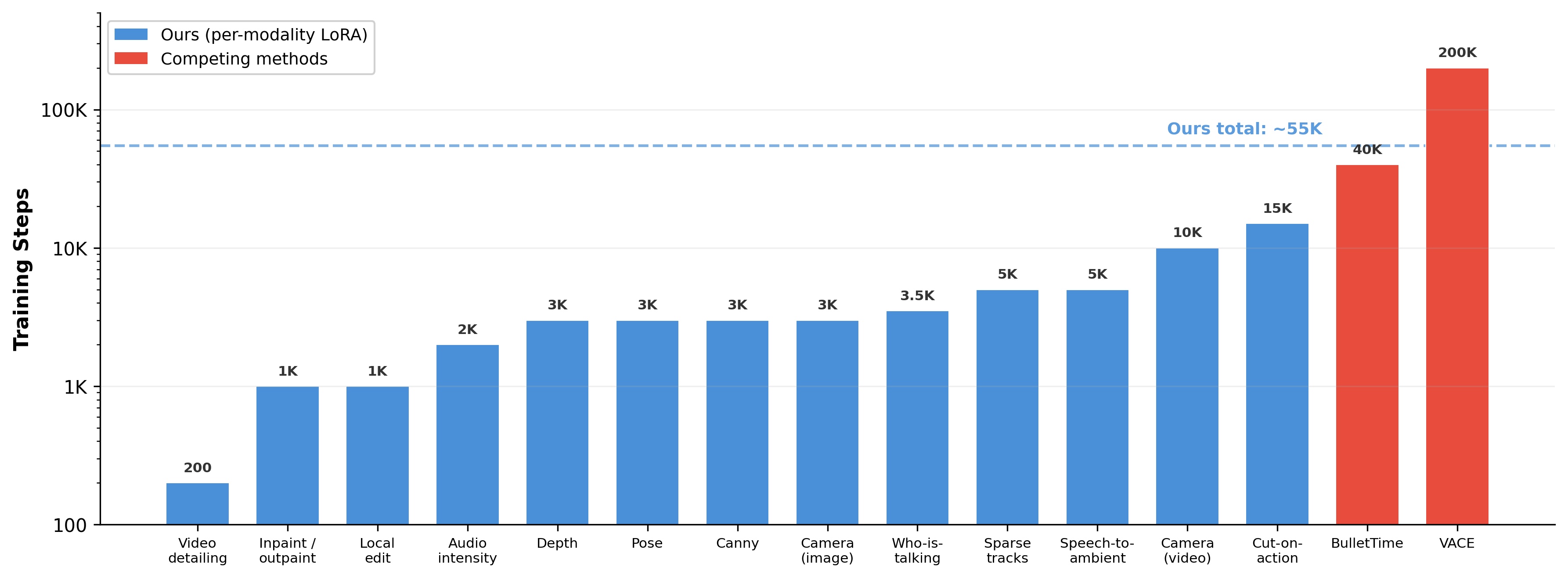}
\caption{Training efficiency comparison. Our per-modality LoRAs range from 200 steps (video detailing) to 15{,}000 steps (cut-on-action), with most spatially-aligned controls converging in approximately 3{,}000 steps. Even summing across all 13 trained LoRAs (${\sim}$55K steps total), the training budget is less than one-third of a single VACE training run (200K steps) and comparable to camera-specific methods like BulletTime (40K iterations). This efficiency stems from fine-tuning only existing weights via LoRA adapters on a frozen backbone, with no new layers to train from scratch.}
\label{fig:training_efficiency}
\end{figure}

\paragraph{Per-modality data construction.}
We describe how each modality's training dataset is constructed.

\begin{itemize}
    \item \textbf{Depth.} We estimate per-frame depth maps from real videos using Video Depth Anything~\cite{chen2025videodepthanything}. The depth video is placed on the reference canvas at half resolution ($2\times$ downscale factor); the target is the original video.

    \item \textbf{Pose.} Pose skeletons are extracted from real videos using DWPose~\cite{yang2023dwpose}. The skeleton video is placed on the reference canvas at half resolution ($2\times$ downscale factor).

    \item \textbf{Canny / edges.} Edge maps are extracted using OpenCV's Canny detector with default thresholds (100/200). The edge video is placed on the reference canvas at half resolution ($2\times$ downscale factor).

    \item \textbf{Inpainting / outpainting.} Training pairs are constructed from real videos by applying random rectangular and free-form masks of varying size (10--60\% of the frame area). Masked regions in the reference canvas are filled with a fixed green color (\texttt{\#66FF00}). The same LoRA handles both inpainting (interior masks) and outpainting (border masks), unified by mask representation.

    \item \textbf{Local edit.} We train on the ROSE dataset~\cite{miao2025rose}, which provides synthetically rendered paired videos of scenes with and without target objects, including their side effects (shadows, reflections). The reference canvas contains the original (unedited) video; the target is the edited version. At inference, we typically provide an initially edited first frame produced by an image editing model, and the LoRA propagates the edit consistently across the full video without requiring an explicit mask.

    \item \textbf{Camera trajectory (from image).} We use real videos and estimate camera extrinsics and per-frame field of view (FOV) using SpatialTrackerV2~\cite{xiao2025spatialtracker}. The reference canvas encodes the target camera trajectory as a canonical grid: a point cloud constructed by unprojecting the input image using a monocular depth estimate, then rendering it along the desired camera trajectory. Disoccluded regions (holes) are filled with the same green color (\texttt{\#66FF00}) used for inpainting masks, so the model learns to inpaint them naturally. The canonical grid is placed at $4\times$ reduced resolution alongside the input image in an image-to-video (im2vid) setup.

    \item \textbf{Camera trajectory (from video).} Training uses the SynCamVideo dataset~\cite{bai2024syncammaster}, which provides synchronized multi-camera videos of 3{,}400 dynamic scenes rendered in Unreal Engine~5, each captured by 10~cameras. For each training pair, the video from one camera is rendered from the trajectory of another camera as a dynamic point cloud; this re-rendered video serves as the LoRA reference. Disoccluded regions are filled with the same green inpainting color (\texttt{\#66FF00}). The target is the ground-truth video from the second camera, and the generated output is aligned to it.

    \item \textbf{Cut-on-action.} Training data is sourced from the MultiCamVideo dataset~\cite{bai2025recammaster}, a synthetic multi-camera dataset rendered in Unreal Engine~5. We use a small subset of this dataset to bootstrap initial training. Because the original dataset contains relatively small inter-camera angles, we employ a multi-angle bootstrapping strategy: after initial training, we use the LoRA itself to generate videos with progressively larger viewpoint differences, and then continue training on these generated pairs to extend coverage to substantially different viewpoints (yaw difference up to 135$^\circ$). This iterative process yields 15{,}000 training samples in total.

    \item \textbf{Video detailing (upscaling).} Training pairs are constructed by downscaling real videos to create degraded inputs. The low-resolution video is upscaled with bilinear interpolation and placed on the reference canvas at full resolution; the target is the original high-resolution video. Convergence is rapid (200~steps compared to 1{,}000--3{,}000 for other modalities), consistent with the simplicity of the super-resolution mapping.

    \item \textbf{Audio intensity.} Training uses 7{,}800 samples from VGGSound~\cite{chen2020vggsound}. The reference audio is a \emph{sonification of the energy envelope}: the amplitude-over-time curve of the original audio is converted into an audio signal. The LoRA generates audio whose temporal dynamics follow the visual content, conditioned on this energy envelope reference. Only the audio branch is trained; the video serves as clean conditioning context.

    \item \textbf{Speech-to-ambient.} We isolate speech from real videos using Demucs (htdemucs\_ft model, 5-pass)~\cite{rouard2023demucs} and Ultimate Vocal Remover (UVR)\footnote{\url{https://github.com/Anjok07/ultimatevocalremovergui}}, followed by spectral gating noise reduction. The reference canvas carries the speech-only audio track; the target is the full audio including ambient sounds. Training uses 2{,}600~samples in 5{,}000~steps.

    \item \textbf{Sparse tracks.} Point tracks are extracted from real videos using AllTracker~\cite{harley2025alltracker} and rendered as colored dots on a black canvas. The reference canvas uses a $2\times$ downscale factor relative to the target resolution.

    \item \textbf{Who-is-talking.} Training data is generated by detecting faces with RetinaFace~\cite{deng2020retinaface}, running voice activity detection (VAD) per speaker using Silero VAD~\cite{silero2024vad}, and constructing an abstract reference layout: colored bounding boxes on a black background (colored when the speaker is active, gray when silent), paired with the mixed audio of all speakers. This is the only LoRA that trains all modules (video, audio, and cross-modal attention), as the task is inherently cross-modal.
\end{itemize}

\section{Small-to-Large Control Grid}
\label{sec:supp_control_grid}

\begin{figure}[t]
\centering
\includegraphics[width=\linewidth]{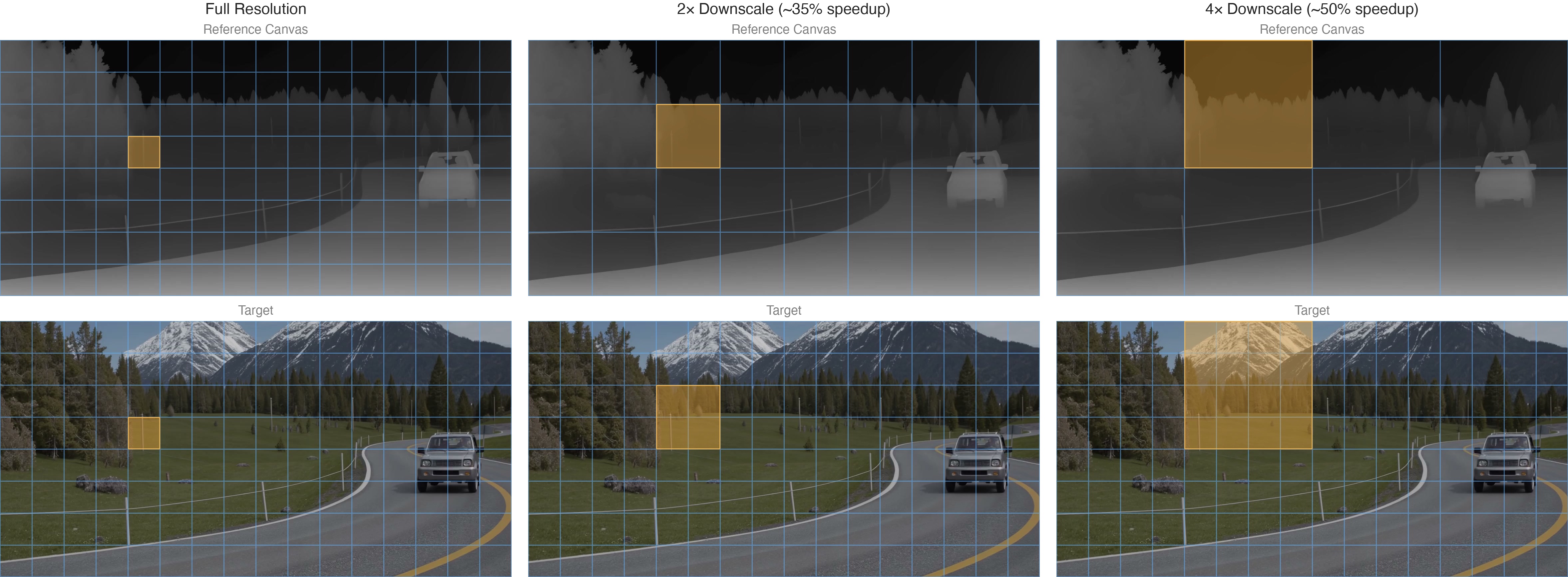}
\caption{Small-to-large control grid, illustrated on a depth-guided example. The reference canvas resolution scales with information density: each highlighted cell in the reference (top row) maps to the corresponding region in the full-resolution target (bottom row). At $2\times$ downscale, one reference token covers $2{\times}2$ target tokens; at $4\times$, one covers $4{\times}4$. Sparser controls thus require fewer reference tokens, reducing self-attention cost and yielding 35--50\% inference speedup at $4\times$ downscale.}
\label{fig:control_grid}
\end{figure}

Not all control modalities carry the same information density (Figure~\ref{fig:control_grid}). Sparser controls such as camera trajectory parameters can be expressed with far fewer tokens than dense, spatially-aligned signals, while modalities that must transfer pixel content require full detail.

By default, every video modality uses a full-resolution ($1\times$) reference canvas. We selectively reduce the canvas resolution for modalities whose information density permits it, controlled by a \emph{reference downscale factor}. Camera-from-image uses a $4\times$ reduction, as the canonical grid encoding is inherently sparse. Dense spatially-aligned controls (depth, pose, canny) and sparse tracks use a $2\times$ reduction; notably, this halved grid is sufficient for faithful spatial control despite the pixel-level correspondence these modalities require. All remaining video modalities (inpainting, video detailing, local edit, camera-from-video, and cut-on-action) use the full-resolution canvas, either because they must reproduce pixel-level content or because reduced-resolution variants have not yet been validated.

The reduction in reference tokens translates directly to reduced self-attention computation, since attention cost scales quadratically with the total token count (target + reference). In our measurements on an H100 GPU, a $2\times$ canvas reduction yields a 25--35\% inference speedup and a $4\times$ reduction yields 35--50\%, with the exact gain depending on the output resolution; see ablation in Section~\ref{sec:supp_ablations}.

\section{Extended Ablations}
\label{sec:supp_ablations}

\begin{figure}[t]
\centering
\includegraphics[width=0.65\linewidth]{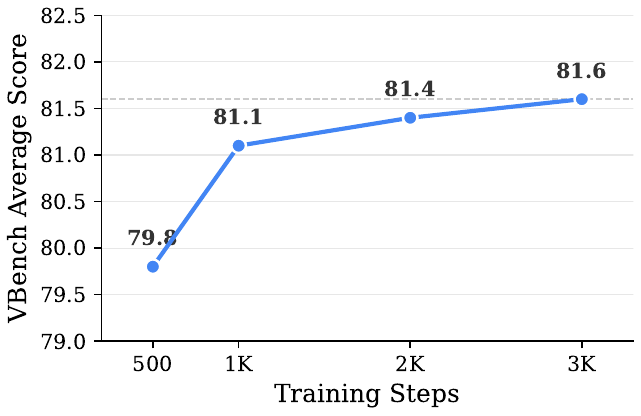}
\caption{VBench average score vs.\ training steps for the depth LoRA. Performance rises steeply from 500 to 1{,}000 steps and plateaus beyond 2{,}000 steps.}
\label{fig:steps_ablation}
\end{figure}

\paragraph{Training steps vs.\ quality.}
Expanding on the main paper's ablation, we evaluate depth checkpoints at finer granularity: 500, 1{,}000, 2{,}000, and 3{,}000 steps (Figure~\ref{fig:steps_ablation}). VBench average scores are 79.8, 81.1, 81.4, and 81.6 respectively, confirming that performance plateaus early. We use 3{,}000 steps as the default for spatially-aligned modalities, as later checkpoints yield diminishing returns (less than 0.2~points beyond 3{,}000 steps).

\begin{figure}[t]
\centering
\includegraphics[width=\linewidth]{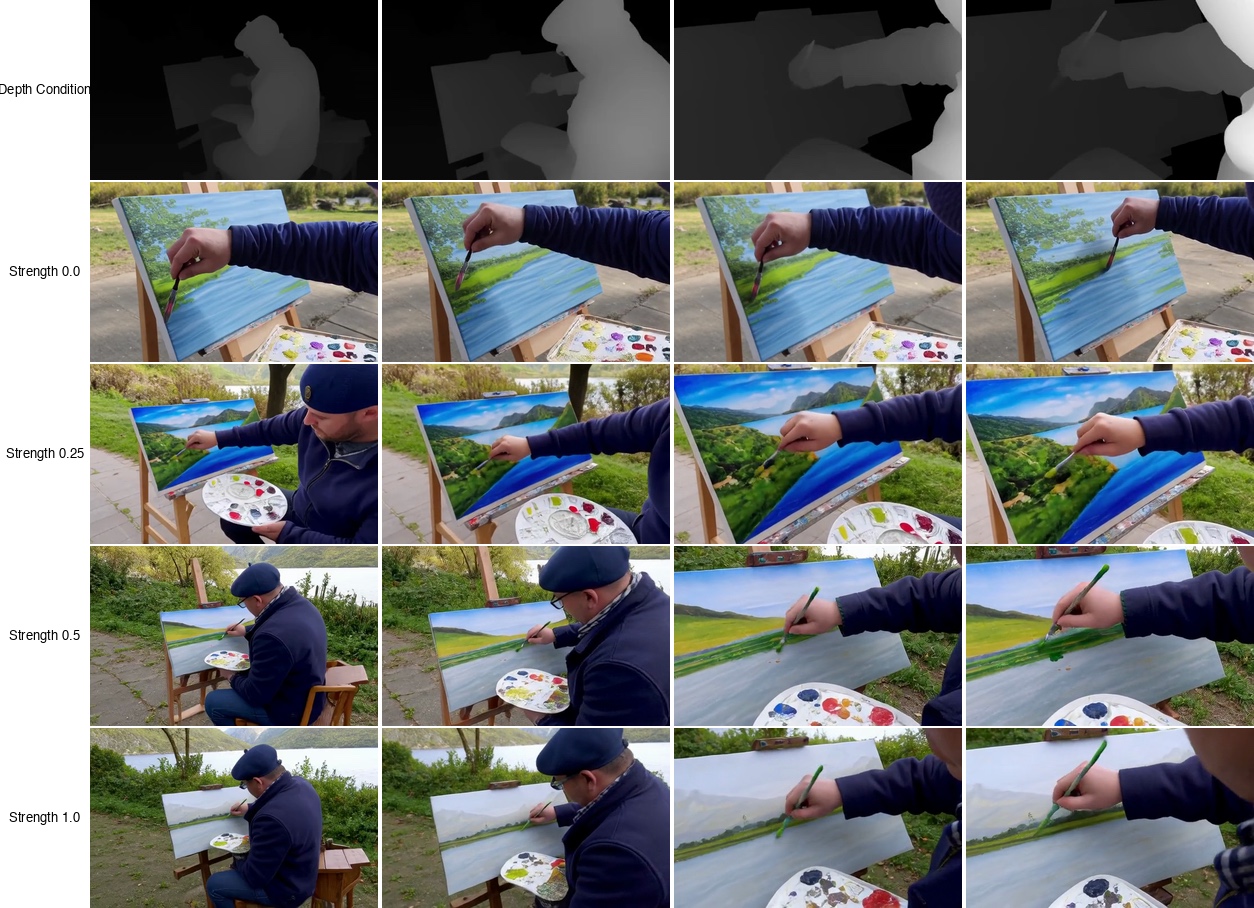}
\caption{Inference-time strength modulation for depth-guided generation. Each row shows four evenly-spaced frames. \textbf{Top row:} depth condition. \textbf{Subsequent rows:} outputs at global strength 0.0, 0.25, 0.5, and 1.0. At lower strengths the model respects the coarse scene layout while exercising creative freedom; at strength 1.0 the output closely follows the depth map.\protect\footnotemark}
\label{fig:strength_modulation}
\end{figure}

\paragraph{Inference-time strength modulation.}
We sweep the global strength parameter from 0 (no reference influence) to 1 (full reference) for depth-guided generation. At intermediate strengths, the model respects the coarse structure of the depth map while exercising more creative freedom on fine details, enabling a continuous trade-off between structural fidelity and generative diversity. The same mechanism naturally extends to temporal modulation (fading the reference influence over time) and spatial modulation (varying strength across regions), though we do not evaluate these variants here. All modulations are purely inference-time operations with no retraining required (Fig.~\ref{fig:strength_modulation}).

\footnotetext{Prompt: ``A person is painting on a canvas outdoors, using a palette with various colors of paint. The person is wearing a dark blue jacket and a matching beret, and is seated on a wooden chair. The canvas depicts a landscape with a body of water and mountains in the background.'' (VACE Benchmark)}

\paragraph{Control grid resolution.}
We compare reference canvases at $1\times$ (full resolution), $2\times$ downscale, and $4\times$ downscale for the camera-from-image modality. Qualitative results are comparable at all resolutions, with a $4\times$ downscale reducing inference latency by 35--50\% relative to the full-resolution canvas depending on output resolution. For dense spatially-aligned modalities (depth, pose), reducing the canvas resolution beyond $2\times$ downscale causes some loss of structural fidelity, as expected given the pixel-level correspondence these modalities require.

\paragraph{Generalization from synthetic data.}
Several LoRAs are trained entirely on synthetic or generated data yet generalize to real-world videos without fine-tuning. Cut-on-action and camera trajectory from video train on multi-camera scenes rendered in Unreal Engine~5 (MultiCamVideo~\cite{bai2025recammaster} and SynCamVideo~\cite{bai2024syncammaster}, respectively). Local edit trains on the ROSE dataset~\cite{miao2025rose}, which consists of synthetically rendered scene pairs. In all cases the LoRAs transfer to diverse real-world content. Who-is-talking is a related but distinct case: its training data is generated by the base model itself rather than by an external renderer, yet the learned control still generalizes to held-out speakers and scenes. These results suggest that the parallel canvas formulation is robust to domain shift between training and inference, likely because the LoRA adapts a small number of weights while the frozen backbone retains its broad visual prior.

\section{Extended Qualitative Gallery}
\label{sec:supp_gallery}

We present additional qualitative results for all control modalities. Each figure shows the control input and generated output across sampled frames. All results use the default generation parameters described in Section~\ref{sec:setup} of the main paper. We group results by category: spatially-aligned controls, video editing, camera trajectory control, and audio-visual modalities.

\begin{figure}[!ht]
\centering
\includegraphics[width=\linewidth]{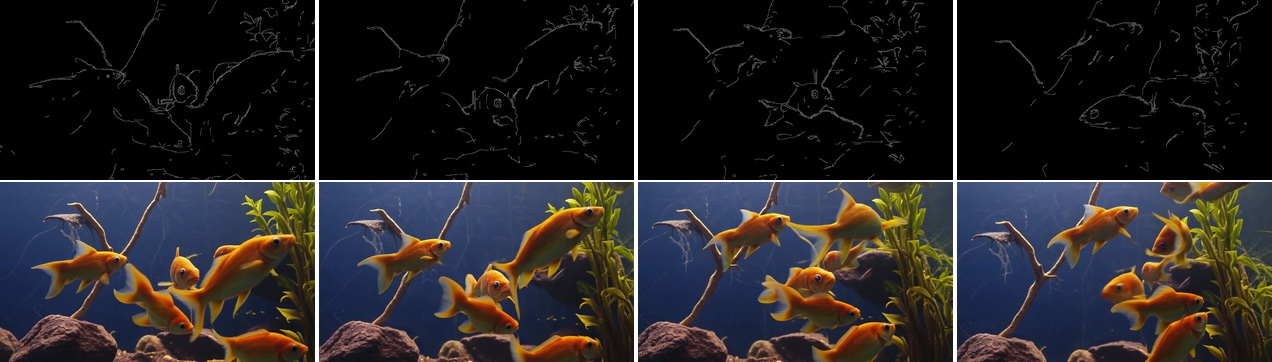}
\caption{Canny edge-guided generation. Control input (top) and generated output (bottom).}
\label{fig:supp_canny}
\end{figure}
\begin{figure}[!ht]
\centering
\includegraphics[width=\linewidth]{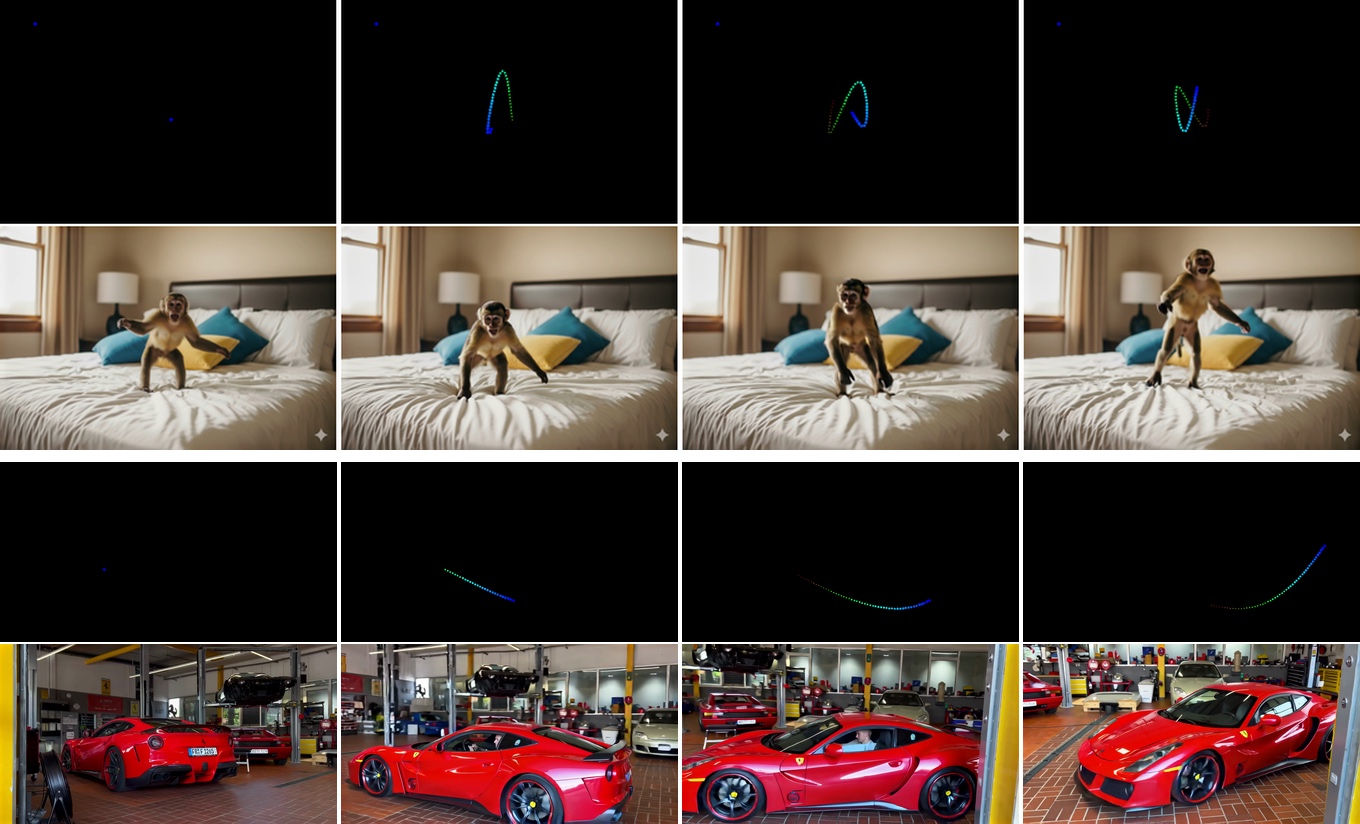}
\caption{Sparse track-guided generation. Point trajectories rendered as colored dots on a black canvas (top) and generated output (bottom).}
\label{fig:supp_sparse_tracks}
\end{figure}

\paragraph{Spatially-aligned controls.}
Figures~\ref{fig:supp_canny}--\ref{fig:supp_sparse_tracks} show results for canny edge and sparse track control. 

\clearpage

\begin{figure}[!ht]
\centering
\includegraphics[width=\linewidth]{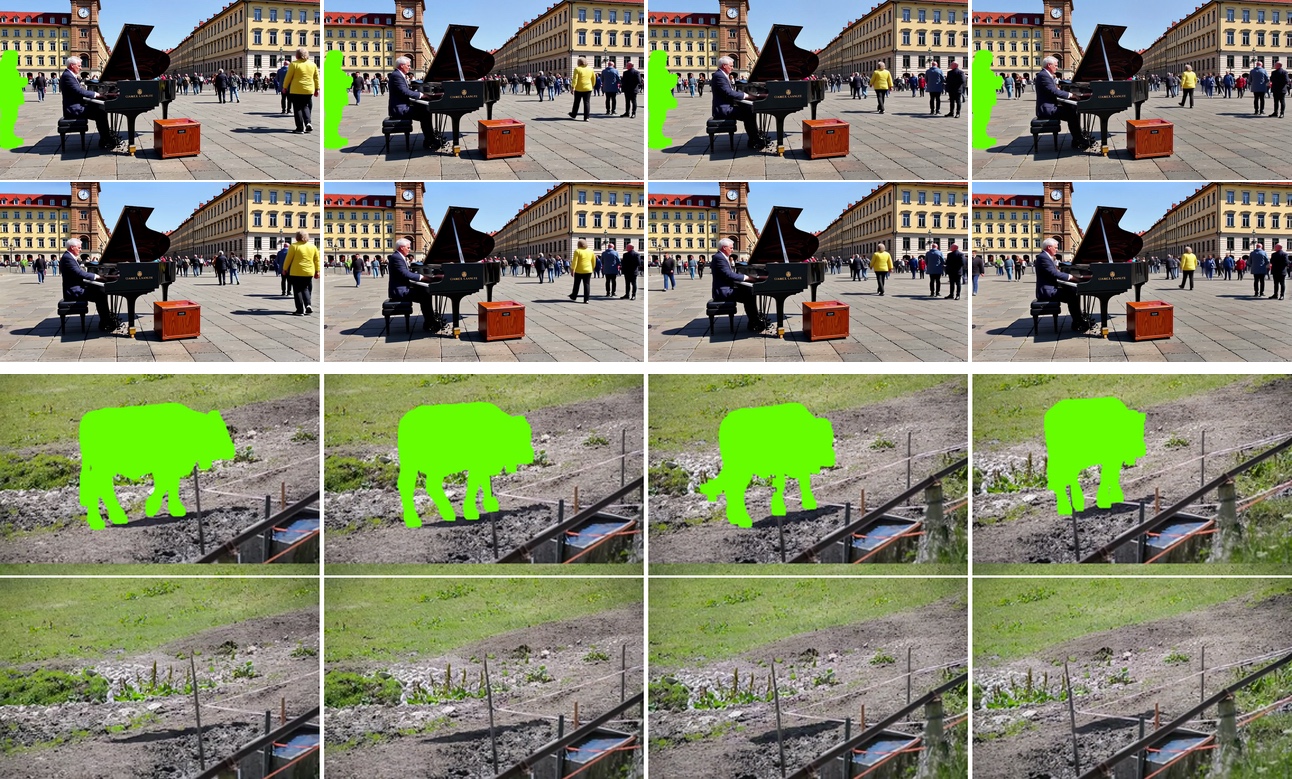}
\caption{Video inpainting. Masked input (top) and completed output (bottom).}
\label{fig:supp_inpainting}
\end{figure}

\begin{figure}[!ht]
\centering
\includegraphics[width=\linewidth]{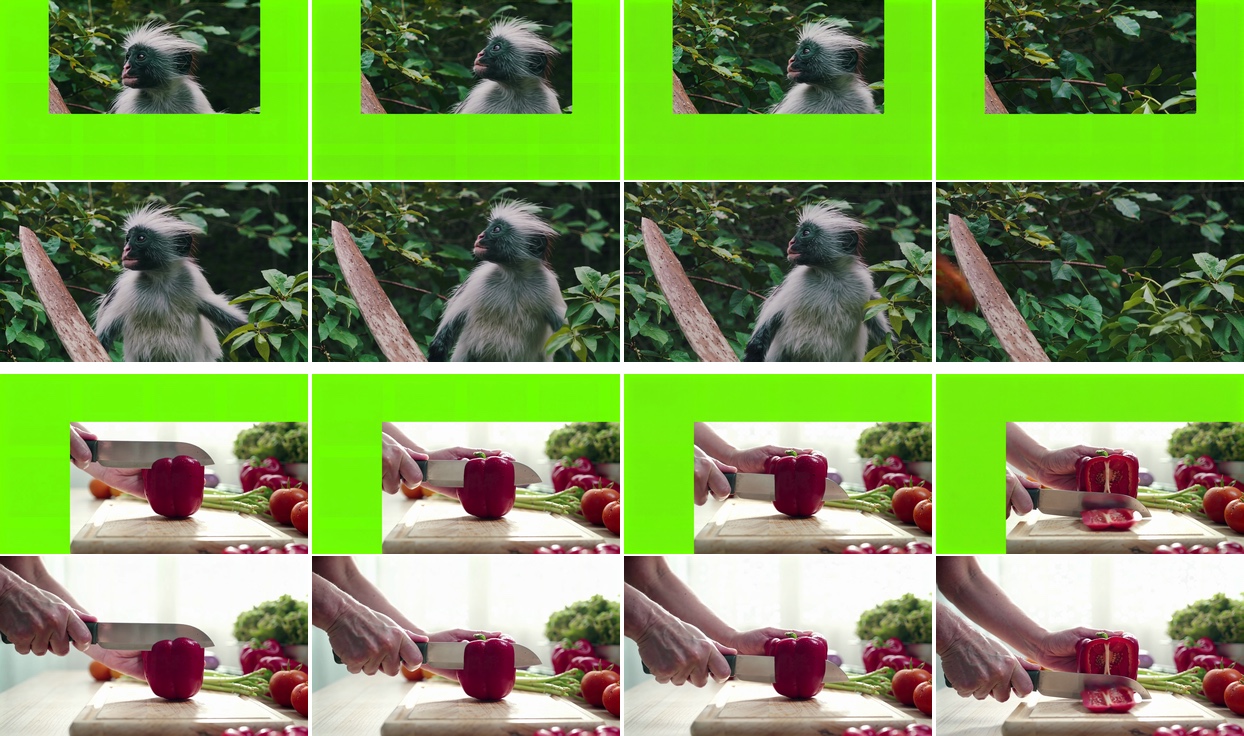}
\caption{Video outpainting. Masked input with border regions (top) and extended output (bottom).}
\label{fig:supp_outpainting}
\end{figure}

\begin{figure}[!ht]
\centering
\includegraphics[width=\linewidth]{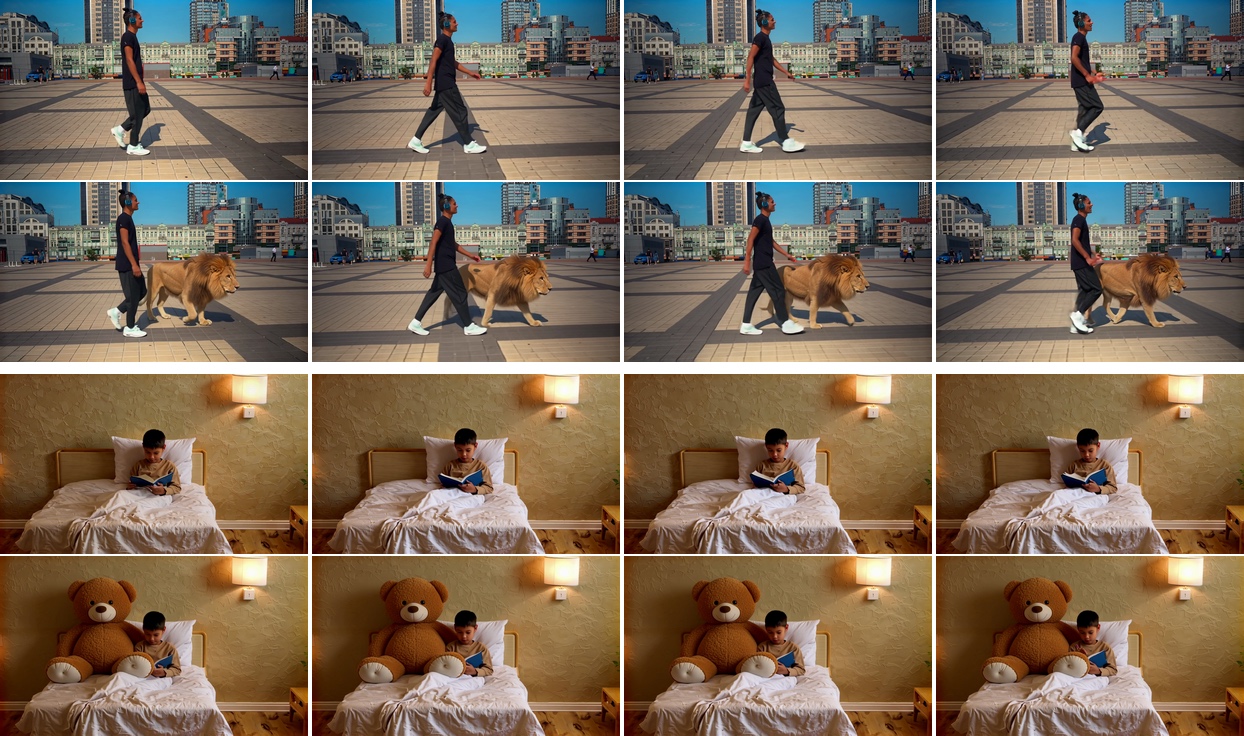}
\caption{Local video editing. Reference video (top) and edited output (bottom). The LoRA propagates a first-frame edit consistently across the video.}
\label{fig:supp_local_edit}
\end{figure}

\begin{figure}[!ht]
\centering
\includegraphics[width=\linewidth]{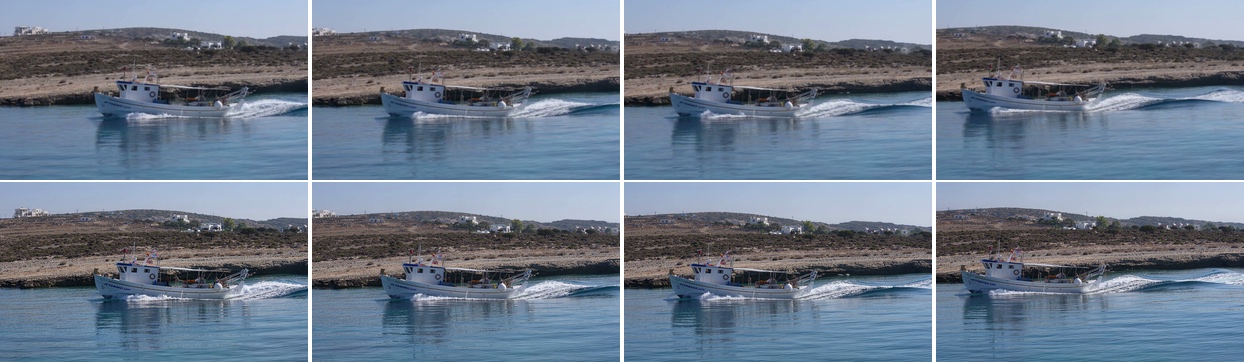}
\caption{Video detailing (upscaling). Low-resolution input (top) and upscaled output (bottom).}
\label{fig:supp_detailing}
\end{figure}

\paragraph{Video editing.}
Figures~\ref{fig:supp_inpainting}--\ref{fig:supp_detailing} show video editing results: inpainting, outpainting, local edit, and video detailing (upscaling).

\clearpage

\begin{figure}[!ht]
\centering
\includegraphics[width=\linewidth]{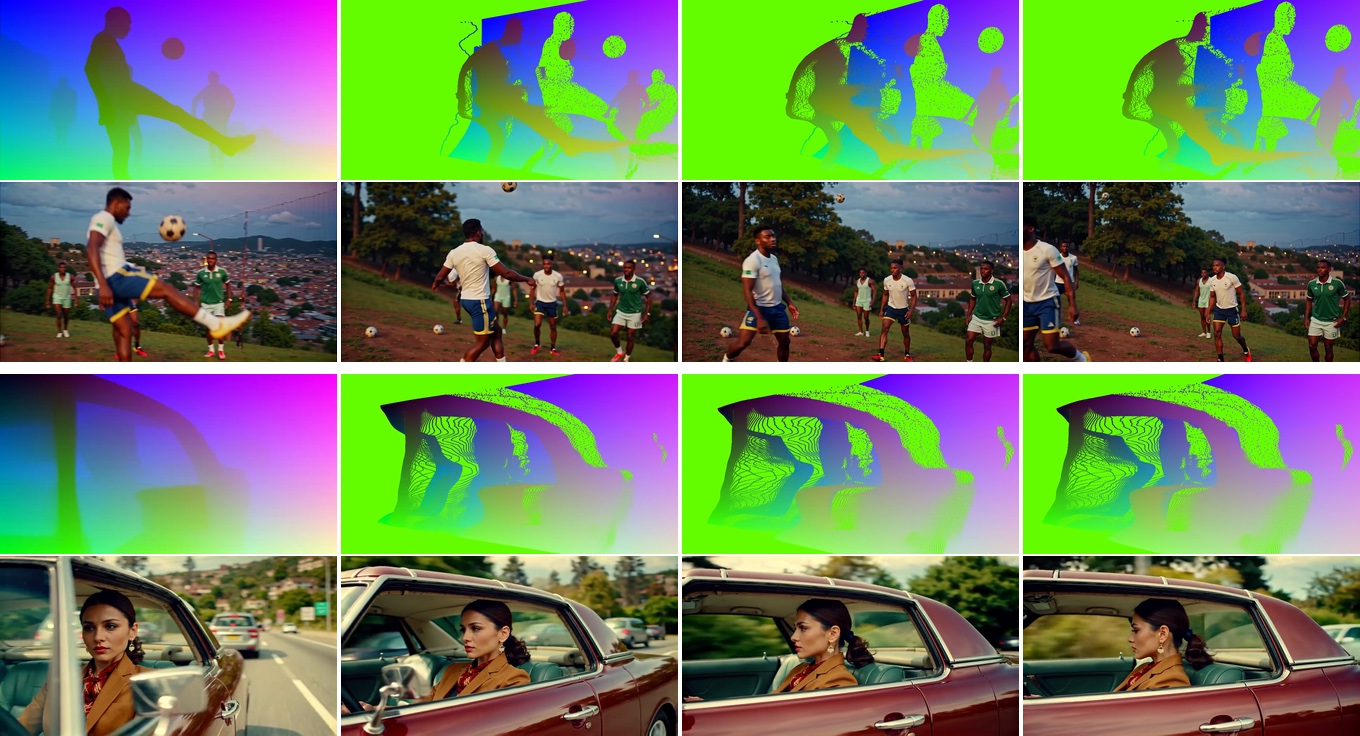}
\caption{Camera trajectory from a single image. Input image and canonical grid reference (top) and generated video with the target camera motion (bottom).}
\label{fig:supp_camera_image}
\end{figure}

\begin{figure}[!ht]
\centering
\includegraphics[width=\linewidth]{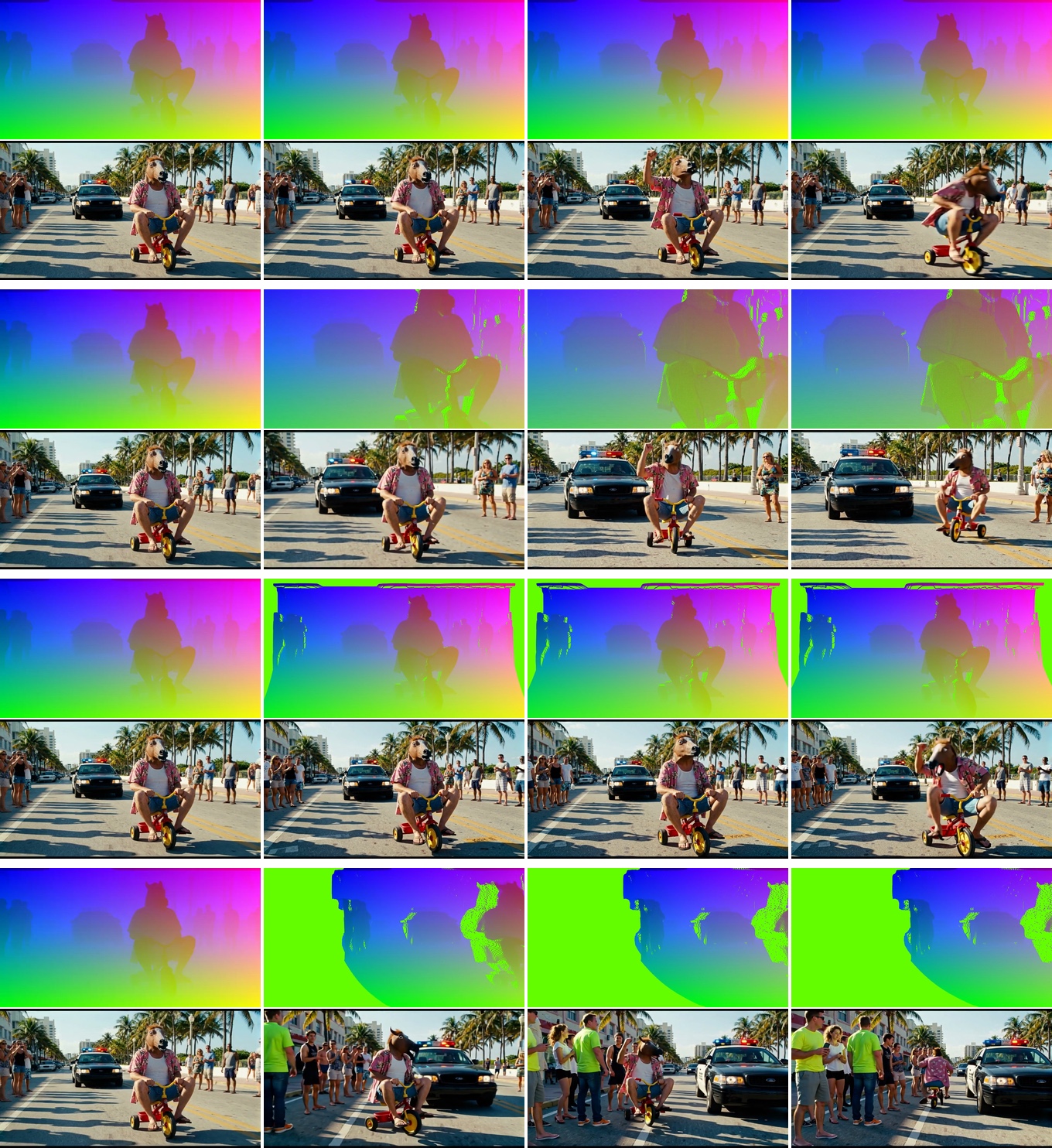}
\caption{Diverse camera trajectories from the same input image, demonstrating controllable camera motion.}
\label{fig:supp_camera_image_diverse}
\end{figure}

\begin{figure}[!ht]
\centering
\includegraphics[width=\linewidth]{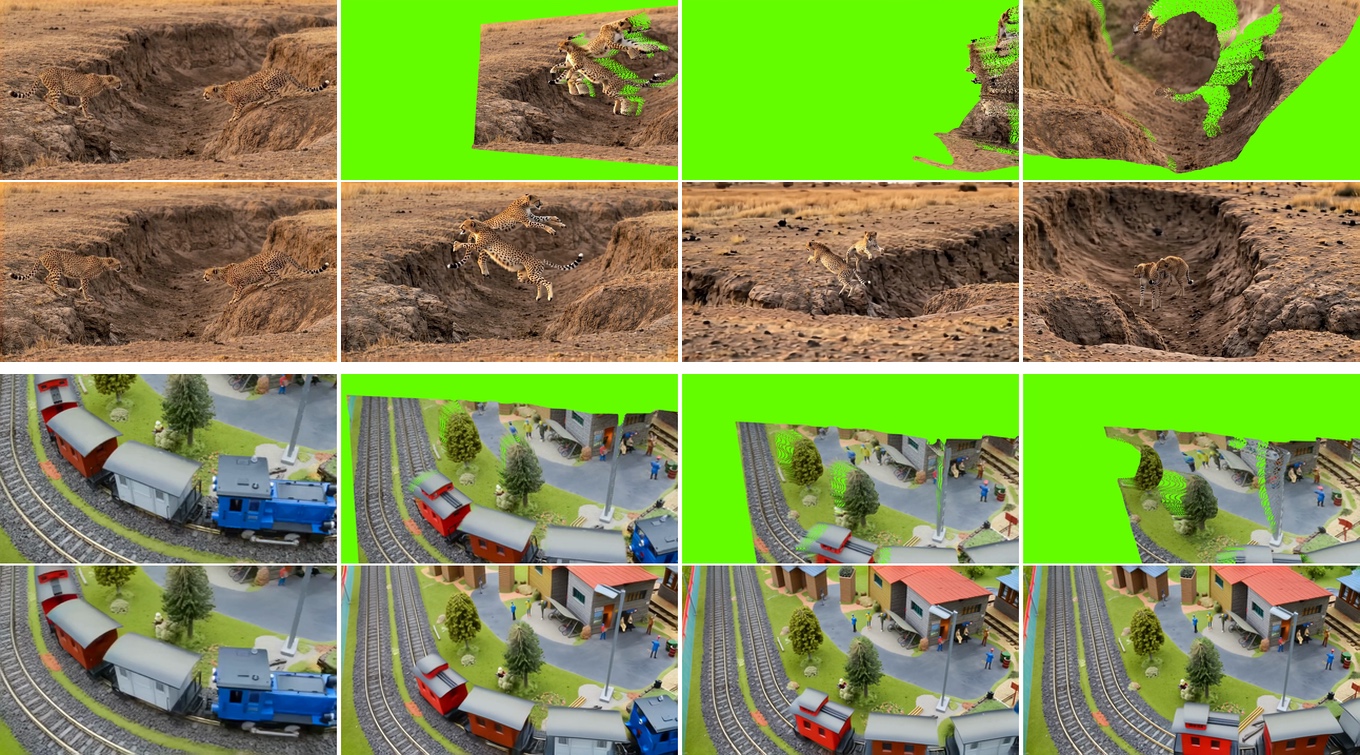}
\caption{Camera trajectory from video. Source video re-rendered at a new camera trajectory while preserving scene motion.}
\label{fig:supp_camera_video}
\end{figure}

\begin{figure}[!ht]
\centering
\includegraphics[width=\linewidth]{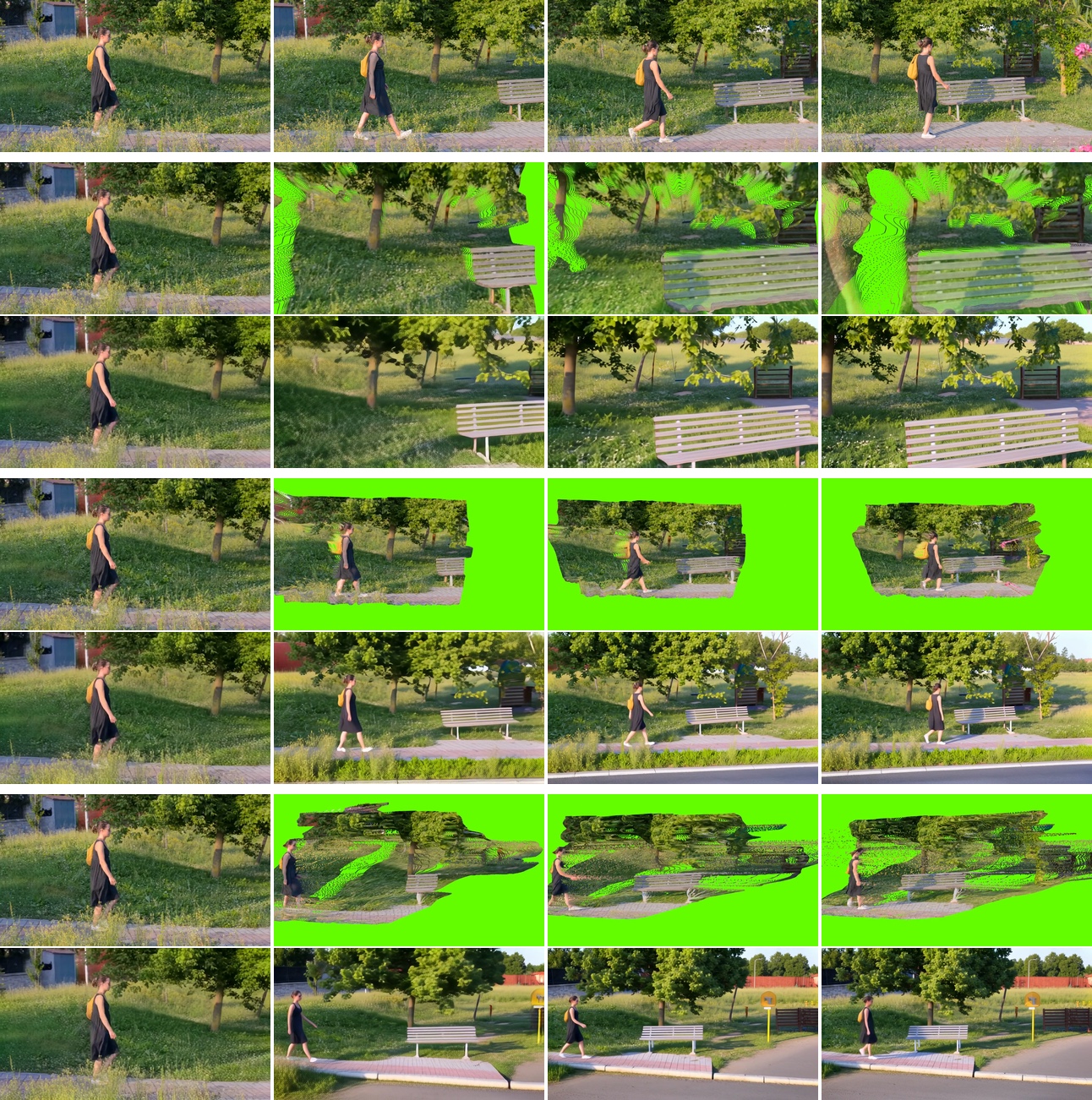}
\caption{Diverse re-rendered camera trajectories from the same source video.}
\label{fig:supp_camera_video_diverse}
\end{figure}

\begin{figure}[!ht]
\centering
\includegraphics[width=\linewidth]{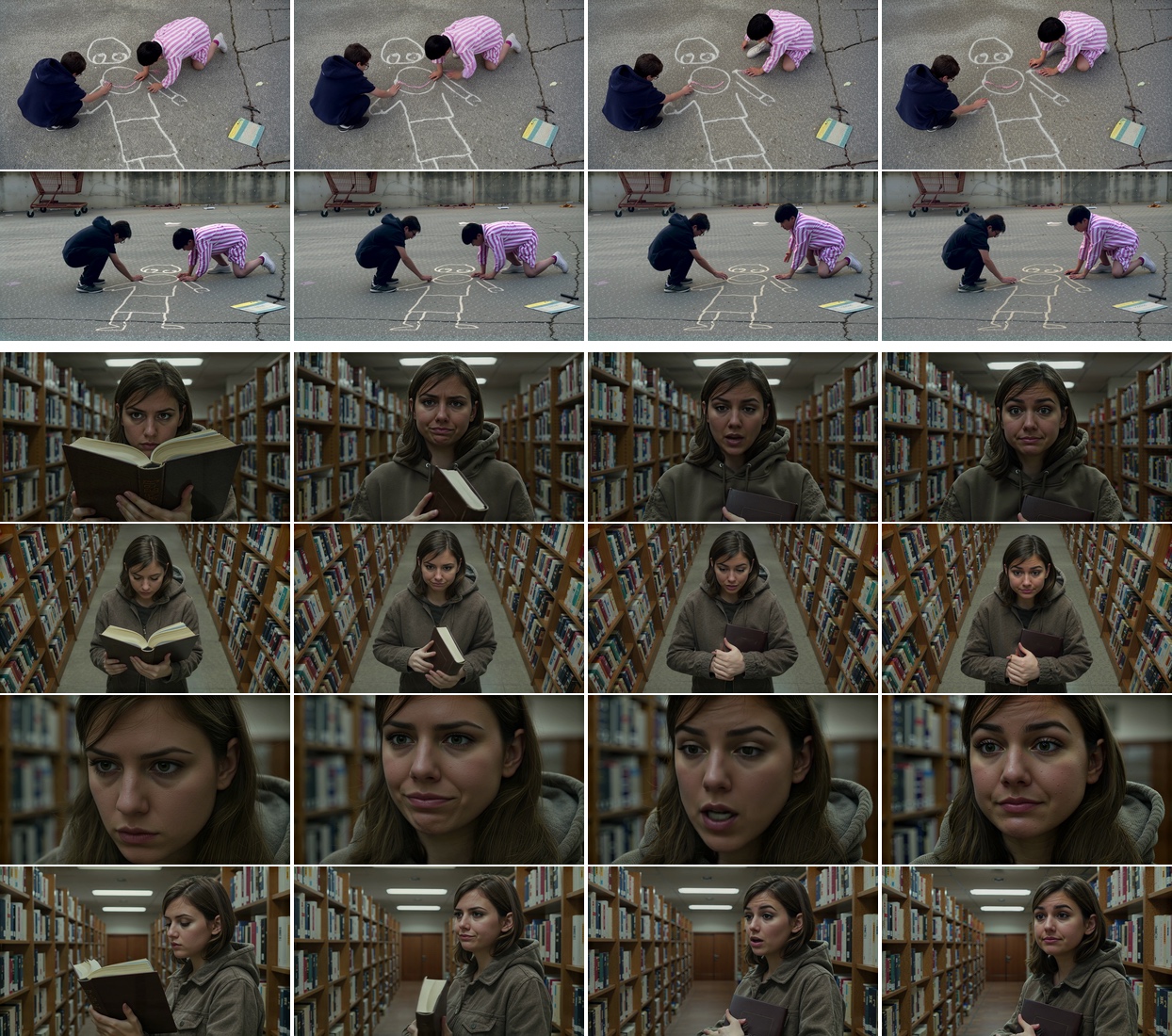}
\caption{Cut-on-action. The source video (top) is re-rendered from a substantially different camera angle (bottom).}
\label{fig:supp_cut_on_action}
\end{figure}

\paragraph{Camera trajectory control.}
Figures~\ref{fig:supp_camera_image}--\ref{fig:supp_cut_on_action} show camera control results, including image-to-video, video-to-video, and cut-on-action with diverse viewpoints.

\clearpage

\begin{figure}[!ht]
\centering
\includegraphics[width=\linewidth]{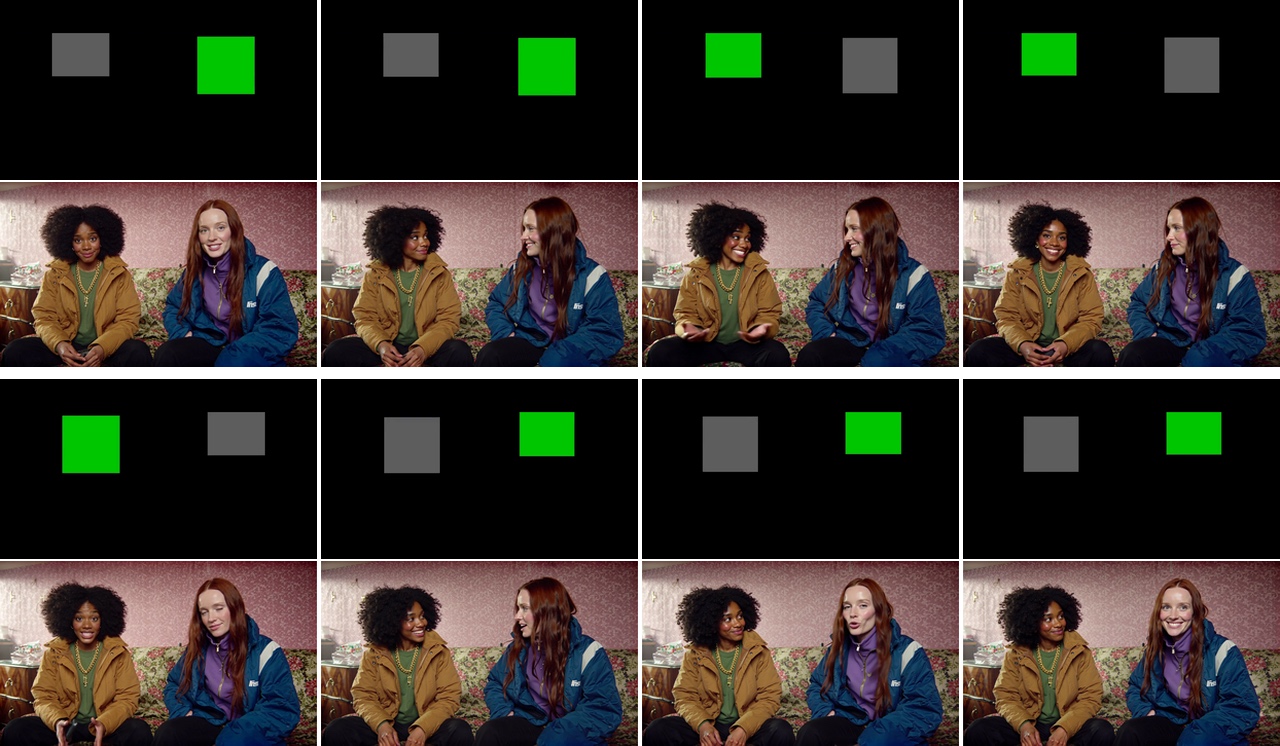}
\caption{Who-is-talking control. An abstract reference layout of colored bounding boxes with voice activity (top) produces multi-person talking video with synchronized lip motion and audio (bottom).}
\label{fig:supp_whoistalking}
\end{figure}

\paragraph{Audio-visual modalities.}
Figure~\ref{fig:supp_whoistalking} shows a who-is-talking result. Audio-only modalities (audio intensity, speech-to-ambient) are best evaluated in the supplementary video, as static frames cannot convey audio quality.

\end{document}